\documentclass[12pt,english]{article}

\usepackage[T1]{fontenc}
\usepackage{textcomp}
\usepackage[latin9]{inputenc}
\usepackage{url}
\usepackage{amstext}
\usepackage{multirow}
\usepackage{amssymb}
\usepackage{float}
\usepackage{graphicx}

\usepackage{geometry}

\usepackage{setspace}
\usepackage{subcaption}
\makeatletter 

\newtheorem{theorem}{Theorem}

\newtheorem{condition}{Assumption}

\newtheorem{corollary}{Corollary}

\newtheorem{example}{Example}

\newtheorem{lemma}{Lemma}

\floatstyle{ruled}
\newfloat{algorithm}{tbp}{loa}
\providecommand{\algorithmname}{Algorithm}
\floatname{algorithm}{\protect\algorithmname}

\newenvironment{proof}[1][Proof]{\textbf{#1.} }{\ \rule{0.5em}{0.5em}}

\makeatother

\usepackage{caption}
\usepackage[labelfont={bf,sf}]{caption}

\usepackage[font=small]{caption}

\usepackage{url}
\usepackage{xcolor}

\usepackage{babel}
\usepackage{comment}
\usepackage{adjustbox}
\usepackage{booktabs}
\usepackage{rotating}
\usepackage{import}

\usepackage{amsmath}
\usepackage{graphicx}

\usepackage{tkz-graph}  
\usetikzlibrary{arrows.meta}

\makeatother

\usepackage{babel}
\begin{document} 

\title{Dynamic Strategy Selection\thanks{An earlier version of this paper, titled Action-State Dependent Dynamic Model Selection, was presented at the Markov Decision Process and Reinforcement Learning Workshop at Trinity College, Cambridge, UK, in 2023. We thank the participants for their comments and suggestions. The first two authors acknowledge financial support from the Leverhulme Trust Grant Award RPG-2021-359. }}
\author{Francesco Cordoni \thanks{Francesco Cordoni contributed to this research while affiliated with Royal Holloway, University of London. Email: fc.cordoni@gmail.com} 
\and Alessio Sancetta\thanks{Corresponding Author. Department of Economics, Royal Holloway University of London, Egham TW20 0EX, UK. Email: asancetta@gmail.com} \and Yuqian Zhao\thanks{University of Sussex Business School, University of Sussex, Brighton BN1 9RH, UK. Email: yuqian.zhao@sussex.ac.uk} }
\date{}  
\maketitle

\vspace*{-1cm}

\begin{abstract}
We consider the problem of dynamically selecting among competing strategies using observable predictors. 
To mitigate estimation noise and excessive switching, we regularise the estimated policy with a switching cost, which may exceed actual transaction costs when these are too small to provide sufficient regularisation.
Since switching costs make current choices affect future decisions, we formulate the problem using dynamic programming. The optimal policy is estimated using a reinforcement learning algorithm based on iterated regressions, making the procedure simple to implement. We apply the method to monthly rotation among momentum, value, profitability, and defensive U.S. equity factors over a sample of about 60 years. Using a parsimonious set of state variables, the estimated policy outperforms both the myopic dynamic policy and the best fixed factor in hindsight in out-of-sample evaluation. Under weak conditions, we show theoretically that the decision rule estimated from empirical data is asymptotically consistent for the unknown optimal decision rule. Simulation results illustrate the finite-sample relevance of these asymptotic results.



\end{abstract}


\section{Introduction}\label{sec:introduction}
This paper considers the problem of dynamically selecting among competing strategies. The relative performance of these strategies may vary over time and depend on observable state variables. At each point in time, a decision maker observes a set of state variables and selects one strategy from a finite collection. Because the strategy selection rule must be estimated, small estimated differences in the performance of competing strategies may induce excessive switching and reduce out-of-sample performance. We therefore introduce a switching cost to discourage such changes. The switching cost can represent transaction costs, for example. However, actual transaction costs may be too small to prevent excessive switching induced by estimation error. Hence, we allow the switching cost to exceed actual transaction costs and act as a regularisation parameter. The regularising role of transaction costs in portfolio choice has also been studied by Hautsch and Voigt (2019). Introducing the switching cost makes the problem dynamic, since the strategy chosen today affects future switching decisions and discounted rewards. The state therefore includes the strategy selected in the previous period. The reward can be based on a Bernoulli utility function to incorporate risk preferences. Our objective is to derive a state-dependent decision rule that maximises expected discounted future rewards.

The distinction between our objective and a myopic decision is important. The discount factor determines the importance given to future rewards. When it is set to zero, the problem reduces to a myopic decision rule, which we refer to as the greedy policy (Greedy). A myopic rule chooses the strategy with the highest expected one-period reward, conditional on the information currently available. It accounts for the immediate cost of switching but ignores how the current choice affects future decisions. Our optimal dynamic rule instead uses a nonzero discount factor and compares the immediate gain from changing strategy with the discounted future value associated with each choice. Consequently, the strategy with the highest expected payoff over the next period need not be selected by our dynamic rule. This feature distinguishes our setting from static selection or ranking exercises and from model-combination methods, which compare or combine competing models based on predictive performance (Pesaran, 1974, 1982; Hansen et al., 2011; Yang, 2004; Timmermann, 2006; Sancetta, 2007, 2010). In our setting, the current choice affects future decisions through the switching cost. 

We formulate the problem as a dynamic programming problem in which the state includes both the observable predictors and the currently held strategy. The optimal policy selects the action that maximises the corresponding action-value function, which satisfies the Bellman equation.
Directly solving such a problem can be difficult when the state contains several predictors and its dynamics are not modelled parametrically. We instead use a reinforcement learning (RL) procedure based on fitted action-value functions. The algorithm repeatedly constructs regression targets by combining current rewards with the estimated discounted value of future rewards, and uses them to estimate the action-specific value functions. The resulting procedure is simple to implement and its estimation reduces to iterating through a sequence of regressions.

RL provides a natural way of estimating dynamic decision rules directly from observed data. Traditional economics approaches to dynamic programming often specify a model for the dynamics of the state variables (Rust, 1996). Our approach instead builds on the least-squares RL literature (Lagoudakis and Parr, 2003; Lazaric et al., 2012). Related RL methods have also been used in recent econometric work. Adusumilli and Eckardt (2026) use temporal-difference methods in the estimation of dynamic discrete-choice models, while Chen et al. (2023) consider optimal policy learning from batch data. Our setting differs in an important respect. The raw state evolves independently of the chosen action. The action remains dynamically relevant because it becomes part of the next period's state and determines whether a switching cost is incurred. We exploit this structure to augment a single observed time series with simulated actions. We estimate the optimal action-value functions through successive least-squares regressions, without estimating the state-transition kernel or requiring access to a generative model. The estimated policy is obtained directly by maximising these functions.

We apply the methodology to factor rotation in the U.S. equity market. The application is related to the recent literature on factor timing, including Dichtl et al. (2019) and Haddad et al. (2020). In contrast to predictive factor-timing approaches, our objective is to learn a dynamic decision rule in which current factor choices affect future decisions through the switching cost. We compare this rule with the Greedy policy. The investor chooses each month among momentum, value, profitability, and a defensive low-volatility factor. The available information contains the recent trend in each factor and, in richer specifications, market-cycle indicators (Goulding et al., 2023) and macroeconomic variables. The latter include the interest rate, the term spread, the credit spread, and inflation. Factor returns are available from 1963 to 2025. The policy is evaluated recursively out of sample beginning in 1973. We consider different levels of the switching cost to vary the strength of regularisation.

The empirical results show that relatively simple predictors contain useful information for dynamic factor selection. The RL and Greedy policies generally outperform the fixed factor strategies, but the advantage of RL becomes more pronounced when switching is more heavily penalised. 
Importantly, this improvement does not appear to be driven simply by less frequent switching. With a switching cost of 120 bps, our best-performing set of state variables, and quadratic utility, the RL and Greedy policies switch a similar number of times: switching frequencies of 7.15\% and 7.95\%, respectively. Nevertheless, their annualised Sharpe ratios are 0.715 (RL) and 0.511 (Greedy), while the Sharpe ratio of the best fixed strategy in hindsight, momentum, is 0.445. 
The results also illustrate a trade-off between information and estimation error. Adding macroeconomic predictors does not improve out-of-sample performance, suggesting that their additional information is more than offset by the estimation error associated with a richer state space.

We provide an asymptotic analysis of the procedure. Under weak conditions on the state process, the rewards, and the approximating class, the value associated with the estimated policy converges to the unknown optimal value as the sample size and the number of fitted-value iterations increase. Under stronger conditions, the convergence rate is close to the usual parametric rate, apart from logarithmic terms. More general results in the Electronic Supplement allow for approximation error and for quantities entering the reward function to be estimated using the same sample. Thus, the procedure can consistently recover an asymptotically optimal decision rule from observed data.

The paper makes three main contributions. First, it formulates a general problem of dynamic strategy selection in which decisions depend on observable state variables, the strategy currently held, risk preferences, and a switching cost that regularises strategy changes. Second, it provides an RL procedure that is straightforward to implement and establishes conditions under which the policy estimated from the data is asymptotically optimal. Third, the application to U.S. equity factors shows that the RL policy can outperform both a myopic decision rule and fixed factor strategies out of sample. The framework is not specific to factor selection and can be used more generally whenever a decision maker must repeatedly choose among competing strategies and current choices affect the value of future decisions.

The paper is structured as follows. The problem setup is discussed in Section \ref{Section_problemSetup}. Section \ref{Section_algorithmMainText} introduces the estimation algorithm for optimal strategy selection. We apply the algorithm to the factor rotation problem in Section \ref{Section_empirical}. Section \ref{Section_asymptoticsSimple} presents the asymptotic analysis under simplified conditions, while the more general results and proofs are given in Section \ref{Section_asymptotics} of the Electronic Supplement. Final remarks are in Section \ref{Section_conclusion}. Appendix \ref{Section_remarksConditions} discusses the assumptions and provides examples. The Electronic Supplement also contains additional empirical results and simulation evidence. The latter confirms the importance of a high switching cost as a regulariser and illustrates the asymptotic results in finite samples.

\section{Problem Setup\protect\label{Section_problemSetup}}

The objective is to select a strategy according to the state of the world. The state may contain economic or market predictors. Because changing strategy may be costly, the relevant state also includes the strategy selected in the previous period. The switching penalty can represent actual costs and frictions, or can be used as a form of regularisation. Let $\tilde{S}_{t}$ denote the vector of economic and market variables observed at time $t$, which we call the raw state. The raw state excludes information about the strategy previously selected. Let $a\in\mathcal{A}$ denote a strategy, equivalently an action selecting that strategy; $\mathcal{A}$ is the set of possible strategies and is finite. We write $S_{t}^{a}=\left(\tilde{S}_{t},a\right)$, where $a$ is the strategy selected in the previous period. We assume that the choice of strategy does not affect the evolution of $\tilde{S}_{t}$, i.e., the raw state is exogenous to the strategy-selection process. We are interested in a map $\pi$ that, given $S_{t}^{a}$, will return the strategy to be selected at time $t$: $\pi\left(S_{t}^{a}\right)\in\mathcal{A}$. This is called a (stationary) policy. We then write $S_{t}^{\pi}$ for $S_{t}^{a}$ with $a=\pi\left(S_{t-1}^{\pi}\right)$. The policy then implies the recursion $S_{t}^{\pi}=S_{t}^{\pi\left(S_{t-1}^{\pi}\right)}$, $t\geq2$, with initial state $S_{1}^{a_{0}}$ for a fixed, but arbitrary $a_{0}\in\mathcal{A}$. The reward at time $t+1$ of following the policy $\pi$ is $R_{t+1}^{\pi}$. Assuming that we are in state $S_{t}^{a}$ for arbitrary $a$, the value of the policy $\pi$ is 
\begin{equation}
V^{\pi}\left(S_{t}^{a}\right)=\mathbb{E}^{\pi}\left[\sum_{i=1}^{\infty}\gamma^{i-1}R_{t+i}^{\pi}\mid S_{t}^{a}\right].\label{EQ_valueFunctionMainText}
\end{equation}
The objective is to maximise this value over the class of admissible policies. Given that the rewards depend on the policy $\pi$, their expectation $\mathbb{E}^{\pi}$ is taken under the policy. The parameter $\gamma\in\left(0,1\right)$ is a discount factor. Larger values place greater weight on future rewards. The optimal value function is
\begin{equation}
V^{*}\left(S_{t}^{a}\right)=\max_{\pi}V^{\pi}\left(S_{t}^{a}\right)\label{EQ_optimalValueFunctionMainText}
\end{equation}
where the maximum is taken over all admissible policies. Under stationarity, \textbf{$V^{\pi}$ }satisfies the Bellman equation 
\begin{equation}
V^{\pi}\left(S_{t}^{a}\right)=\mathbb{E}^{\pi}\left[R_{t+1}^{\pi}+\gamma V^{\pi}\left(S_{t+1}^{\pi}\right)\mid S_{t}^{a}\right].\label{EQ_BellmanEqMainText}
\end{equation}
Because the one-period reward depends on both the inherited and current strategies, let $R_{t+1}^{\left(a,a'\right)}$ be the reward at time $t+1$ from selecting strategy $a'$ at time $t$ when the strategy at time $t-1$ was $a$. An action value function is defined to be 
\[
Q^{\pi}\left(S_{t}^{a},a'\right)=\mathbb{E}\left[R_{t+1}^{\left(a,a'\right)}+\gamma V^{\pi}\left(S_{t+1}^{a'}\right)\mid S_{t}^{a}\right],
\]
where the first term inside the expectation is the immediate reward, while we refer to the second term as the (discounted) continuation value. Here, the expectation operator in the definition of the action value function does not depend on $\pi$. The optimal action value function is 
\[
Q^{*}\left(S_{t}^{a},a'\right)=\mathbb{E}\left[R_{t+1}^{\left(a,a'\right)}+\gamma V^{*}\left(S_{t+1}^{a'}\right)\mid S_{t}^{a}\right].
\]
In this case, we can relate it to the optimal value function: 
\[
V^{*}\left(S_{t}^{a}\right)=\max_{a'\in\mathcal{A}}Q^{*}\left(S_{t}^{a},a'\right).
\]
Moreover, given the current state $S_{t}^{a}$, the optimal policy selects the strategy that maximises the optimal action-value function:
\[
\pi^{*}\left(S_{t}^{a}\right)=\arg\max_{a'\in\mathcal{A}}Q^{*}\left(S_{t}^{a},a'\right)
\]
assuming the maximiser is unique; if not, in the above display we replace the equal sign with $\in$. The examples below show how rewards and costs can be specified in data applications.

\begin{example}\label{Example_reward}(Rewards) Let $\mathrm{Ret}_{t+1}^{\left(a'\right)}$ be the return at time $t+1$ from having selected strategy $a'\in\mathcal{A}$ at time $t$. The reward is given by 
\begin{equation}
R_{t+1}^{\left(a,a'\right)}=U\left(\mathrm{Ret}_{t+1}^{\left(a'\right)}-\mathrm{Cost}_{t}^{\left(a,a'\right)}\right)\label{EQ_rewardUtility}
\end{equation}
where $U\left(x\right)$ is a Bernoulli utility function and $\mathrm{Cost}_{t}^{\left(a,a'\right)}$ is a cost function. In the empirical application (Section \ref{Section_empirical}) we use $U\left(x\right)=x-\left(\eta/2\right)x^{2}$, where $\eta$ is a risk-aversion parameter. 
\end{example}

\begin{example}\label{Example_costRegularisation}(Fixed cost) There is a fixed cost to switch so that $\mathrm{Cost}_{t}^{\left(a,a'\right)}=\mathrm{cost}\times1_{\left\{ a\neq a'\right\} }$ for some $\mathrm{cost}\geq0$. This is particularly plausible when the role of the cost is to reduce switching, so that the policy becomes smoother.
\end{example}

More generally, the switching cost may depend on the pair of strategies, rather than being constant across all switches. For example, switching between two similar strategies may incur a smaller penalty than switching between substantially different strategies.

\section{Estimation Algorithm\protect\label{Section_algorithmMainText}}

The estimation algorithm performs a number of regressions whose objective is to iteratively estimate the optimal value function. 
For the purpose of estimation we augment the data with a sequence of randomly generated actions $\left(A_{t}\right)_{t\geq0}$ with values in $\mathcal{A}$, independent of the observed raw states. In our empirical application, we take them to be independent identically distributed (i.i.d.) and uniformly distributed over $\mathcal{A}$. We motivate more general choices in Section \ref{EQ_conditionStateActionPair} of the Electronic Supplement. Using the notation of Section \ref{Section_problemSetup} we write $S_{t}:=S_{t}^{A_{t-1}}$ and $R_{t+1}^{a}=R_{t+1}^{\left(A_{t-1},a\right)}$. The purpose of the sequence $\left(A_{t}\right)$ is to construct randomly augmented states and the corresponding rewards used for estimation. 
Define the state-space $\mathcal{S}=\tilde{\mathcal{S}}\times\mathcal{A}$, i.e. $S_{t}=\left(\tilde{S}_{t},A_{t-1}\right)\in\mathcal{S}$. To approximate the action value function we use basis functions $\varphi=\left(\varphi_{1},\varphi_{2},...,\varphi_{K}\right)$ where $\varphi_{j}:\mathcal{S}\rightarrow\mathbb{R}$. Then, the estimation algorithm can be summarised as follows. 

For $j=1,2,\dots,J$, define $Y_{t+1}^{\left(a,j\right)}=R_{t+1}^{a}+\gamma\hat{V}^{\left(j-1\right)}\left(S_{t+1}^{a}\right)$ with $\hat{V}^{\left(0\right)}:=0$ as initial estimated value of the value function. Here $J$ is the total number of iterations in the algorithm. 

For each $a\in\mathcal{A}$, regress $Y_{t+1}^{\left(a,j\right)}$ on $\varphi\left(S_{t}\right)$, $t=1,2,\dots,n$, to obtain the $K\times1$ vector of regression coefficients $\hat{b}^{\left(a,j\right)}$; here $n$ is the sample size. In particular, we include a ridge penalty. Formally, $\hat{b}^{\left(a,j\right)}$ is the minimiser of 
\[
\frac{1}{n}\sum_{t=1}^{n}\left(Y_{t+1}^{\left(a,j\right)}-\varphi\left(S_{t}\right)^{\mathrm{T}}b\right)^{2}+\rho b^{\mathrm{T}}b
\]
with respect to $b\in\mathbb{R}^{K}$, where $\rho\geq0$ is a ridge parameter. The estimated action value function is $\hat{Q}^{\left(j\right)}\left(S_{t+1}^{a},a'\right):=\left[\varphi\left(S_{t+1}^{a}\right)^{{\rm T}}\hat{b}^{\left(a',j\right)}\right]_{B}$ where $\left[x\right]_{B}={\rm sign}\left(x\right)\min\left\{ \left|x\right|,B\right\} $ is a capping operator; the action value function is bounded in a range $\left[-B,B\right]$ for some constant $B>0$. The estimated value function is 
\[
\hat{V}^{\left(j\right)}\left(S_{t+1}^{a}\right)=\max_{a'\in\mathcal{A}}\hat{Q}^{\left(j\right)}\left(S_{t+1}^{a},a'\right).
\]
Finally, the estimated policy is $\pi_{j}\left(s\right)=\arg\max_{a'\in\mathcal{A}}\hat{Q}^{\left(j\right)}\left(s,a'\right)$ for $s\in\mathcal{S}$. We suggest using a ridge penalty to have better control of the estimation error in the regression. The capping operator is used to ensure that quantities stay bounded. Under technical conditions, in Section \ref{Section_asymptotics} we show that as $J$ increases, $\hat{\pi}^{\left(J\right)}$ produces asymptotically the same optimal value function as the one obtained using the unknown optimal policy.

\section{Factor Rotation in the U.S. Equity Market}\label{Section_empirical}
In this section, we study a monthly factor allocation problem in the U.S. equity market. Our primary application considers four widely studied and economically distinct factor strategies: momentum (MOM), value (HML), profitability (RMW), and defensive low volatility (DEF). These four factors are commonly followed by academics and practitioners and provide a parsimonious setting in which to study how their relative performance and conditional dominance vary across market states. Our objective is to learn a dynamic policy for rotating among four factor strategies to maximise expected reward net of a switching penalty.
\subsection{Data and Empirical Design}\label{Section_dataDesign}

\paragraph{Data}
Returns on the momentum, value, and profitability factors are obtained from the Kenneth R. French Data Library. Because a defensive long-short low-volatility factor is not directly available, we construct a DEF factor from CRSP daily stock returns. Motivated by Novy-Marx (2014), DEF buys low-volatility stocks and shorts high-volatility stocks using a Fama-French-style 2 $\times$ 3 portfolio construction. The raw factor return data span July 1963 to December 2025.


\paragraph{State} 
To construct the conditional information, i.e. the raw state $\tilde S_t$, we consider three groups of state variables. First, we build a factor trend (Trend) state for each candidate factor strategy. The factor trend is the average monthly return on each factor over the preceding 12 months.

Second, inspired by the momentum turning-point classification of Goulding et al. (2023), we classify each month in the sample into Bull, Bear, Correction, and Rebound (Cycle) regimes according to the signs of the historical 12-month average market excess return and the current one-month market excess return. The Bull dummy is omitted to avoid multicollinearity with the intercept.

Third, we consider macroeconomic information (Macro) summarised by the interest rate, the term spread, the credit spread, and inflation variables. We use the 12-month inflation rate excluding food and energy and the difference between the 12-month headline and core inflation rates as a proxy for the contribution of food and energy.\footnote{For simplicity, the inflation variables are obtained from FRED, which is not a real-time database. We use non-seasonally adjusted CPI data because, unlike seasonally adjusted CPI, they are not subject to the routine revisions arising from the annual recalculation of seasonal adjustment factors. We compute 12-month inflation rates to remove seasonal effects and lag the resulting variables by one month to account for the delay in their release.}

All the state variables are aligned so that only information available at the time of the strategy choice is used. This avoids look-ahead bias. Details on the state variables and their transformations are provided in Section \ref{Section_empirical} of the Electronic Supplement.

\paragraph{Reward and estimation}
We use the reward function from Example \ref{Example_reward} with fixed cost from Example \ref{Example_costRegularisation}, in Section \ref{Section_problemSetup}. The switching cost is used only as a regularisation parameter rather than as a direct measure of realised transaction costs. We refer to $\mathrm{Ret}_{t+1}^{(a')}-\mathrm{Cost}_{t}^{(a,a')}$ as the net return. 

We consider risk-aversion parameters $\eta \in \{0,3\}$ and switching costs of 20, 80, and 120 basis points (bps). When $\eta=0$, the reward in Example \ref{Example_reward} coincides with the net return.  When $\eta>0$, the quadratic term penalises the second moment of net returns and thus discourages policies with high volatility of the returns, possibly resulting in higher Sharpe ratios. We use $\eta=3$ to represent a moderate level of risk aversion.

We set the discount factor to $\gamma=0.9$. Because decisions are made monthly, this value implies a discount half-life of approximately 6.6 months. This provides a reasonably long horizon to capture the continuation value of future strategy choices. Increasing $\gamma$ to higher values could place heavy weight on rewards several years in the future. Our results remain substantially the same when setting $\gamma=0.98$.

In our implementation, we approximate each action-specific $Q$-function using an additive model. The dependence on the previously selected strategy is captured by dummy variables (see the model in (\ref{EQ_linearSpecification}), in the Appendix for details). All non-binary variables enter as a linear and a quadratic term.
We estimate the fitted $Q$-regressions using ridge regularisation. The optimal ridge penalty parameter is selected using generalised cross-validation (Golub et al., 1979). For the effects of alternative ridge parameters and other computational details, we refer to Section \ref{Section_empiricalsupp} of the Electronic Supplement. 
 
In addition to the RL policy, we evaluate a myopic Greedy policy that uses the same state information and action set but omits the continuation-value component. This is achieved using one fitted $Q$-iteration, so the estimated action values depend only on the expected reward of a period. This is equivalent to setting $\gamma=0$. We also consider four fixed strategies that remain invested in momentum, value, profitability, or defensive low risk throughout the same out-of-sample evaluation period. 

We use an expanding-window estimation scheme with an initial sample of 120 months, using the complete observations available after constructing the state variables. The out-of-sample evaluation therefore begins in July 1973 and contains 629 monthly decisions, corresponding to factor returns realised from August 1973 through December 2025.


\subsection{Results}\label{sec:result}
Table \ref{tab:state_information_performance} compares the performance of RL and Greedy strategies. The results can be summarised as follows. First, RL and Greedy policies generally outperform fixed factor strategies. The relative performance is tested by computing the $t$-statistics from the out-of-sample monthly reward differences between RL and Greedy. 
The results highlight the importance of sufficiently strong regularisation. RL tends to perform best at the highest switching cost. This is despite the fact that rewards are computed net of that cost. Its advantage over Greedy also becomes particularly pronounced at 120 bps. Comparisons across costs need not be monotonic, since a higher switching cost both regularises the policy and directly reduces the reward whenever a switch occurs.
As the regularisation cost increases, the policies also switch less frequently between factors. However, at a given regularisation cost, RL does not necessarily switch less frequently than Greedy. Thus, the improvement of RL over Greedy is not driven by lower switching frequency.

Second, using factor trends as state variables contains useful information for dynamic factor selection. Adding market-cycle turning-point indicators does not systematically improve performance when $\eta = 0$. In contrast, when $\eta = 3$, the Trend + Cycle specification leads to stronger performance for the RL policy. Third, adding a broader set of macroeconomic variables does not improve out-of-sample performance. Although these variables may contain additional information about expected factor returns, any gain appears to be offset by the estimation error from a larger state space. This result reflects an information estimation trade-off, as a richer information set may increase estimation error and does not necessarily improve out-of-sample policy performance.

We also compute the annualised Sharpe ratio of net returns, as defined in Section \ref{Section_dataDesign}, i.e. returns, rather than rewards, after subtracting the switching cost. Two points are worth noting. First, for the better-performing specifications, allowing for risk aversion with $\eta=3$ can lead to substantially higher Sharpe ratios than when $\eta=0$. Second, despite returns being computed net of a switching cost that is substantially higher than expected actual transaction costs, the best-performing policies considerably outperform the fixed factors in terms of Sharpe ratio. Since the fixed-factor strategies do not switch, they incur no switching cost.

\begin{table}[htbp]
\centering
\caption{Out-of-Sample Performance across State Information Sets. Results for RL, Greedy, and fixed-factor strategies under different state information sets, risk-aversion parameters ($\eta=0,3$), and switching costs. Mean monthly rewards are in percentages. The Sharpe ratios are computed from net returns and annualised. Switching frequency is the percentage of out-of-sample months in which the selected factor differs from the previous factor position. The ``Trend'' information set contains the four factor-trend variables. ``Trend + Cycle'' additionally contains the market turning-point indicators, while ``Trend + Cycle + Macro'' further adds the macroeconomic state variables. Fixed policies maintain the same factor position throughout the out-of-sample period. Stars on the RL policy indicate statistically significant differences between the RL and Greedy rewards. The symbols $^{*}$, $^{**}$, and $^{***}$ denote significance at the 10\%, 5\%, and 1\% levels, respectively.
}
\label{tab:state_information_performance}
\begin{adjustbox}{max width=\linewidth}
\begin{tabular}{llccccccccc}
\toprule
& & \multicolumn{3}{c}{20 bps} & \multicolumn{3}{c}{80 bps} & \multicolumn{3}{c}{120 bps} \\
\cmidrule(lr){3-5} \cmidrule(lr){6-8} \cmidrule(lr){9-11}
State & Policy & \begin{tabular}[c]{@{}c@{}}Monthly\\reward (\%)\end{tabular} & Sharpe & \begin{tabular}[c]{@{}c@{}}Switch\\freq. (\%)\end{tabular} & \begin{tabular}[c]{@{}c@{}}Monthly\\reward (\%)\end{tabular} & Sharpe & \begin{tabular}[c]{@{}c@{}}Switch\\freq. (\%)\end{tabular} & \begin{tabular}[c]{@{}c@{}}Monthly\\reward (\%)\end{tabular} & Sharpe & \begin{tabular}[c]{@{}c@{}}Switch\\freq. (\%)\end{tabular} \\
\midrule
\multicolumn{11}{l}{\textit{Panel A: $\eta=0$}} \\
\addlinespace[2pt]
\multirow{2}{*}{Trend} & RL & 0.738\textsuperscript{***} & 0.586 & 8.74 & 0.746\textsuperscript{**} & 0.600 & 5.25 & 0.777\textsuperscript{**} & 0.633 & 4.29 \\
 & Greedy & 0.588 & 0.473 & 9.38 & 0.588 & 0.459 & 5.09 & 0.533 & 0.430 & 3.66 \\
\addlinespace
\multirow{2}{*}{Trend+Cycle} & RL & 0.721\textsuperscript{**} & 0.628 & 18.60 & 0.669 & 0.561 & 11.92 & 0.706\textsuperscript{*} & 0.595 & 8.74 \\
 & Greedy & 0.580 & 0.475 & 17.49 & 0.585 & 0.472 & 11.76 & 0.529 & 0.424 & 9.06 \\
\addlinespace
\multirow{2}{*}{\begin{tabular}[c]{@{}l@{}}Trend+Cycle\\+Macro\end{tabular}} & RL & 0.541 & 0.471 & 25.12 & 0.531 & 0.464 & 14.47 & 0.470\textsuperscript{*} & 0.418 & 10.02 \\
 & Greedy & 0.504 & 0.433 & 22.10 & 0.377 & 0.301 & 14.15 & 0.285 & 0.228 & 11.29 \\
\addlinespace
\multirow{4}{*}{Fixed} & MOM & 0.555 & 0.445 & 0.00 & 0.555 & 0.445 & 0.00 & 0.555 & 0.445 & 0.00 \\
 & HML & 0.282 & 0.315 & 0.00 & 0.282 & 0.315 & 0.00 & 0.282 & 0.315 & 0.00 \\
 & RMW & 0.281 & 0.417 & 0.00 & 0.281 & 0.417 & 0.00 & 0.281 & 0.417 & 0.00 \\
 & DEF & 0.016 & 0.010 & 0.00 & 0.016 & 0.010 & 0.00 & 0.016 & 0.010 & 0.00 \\
\midrule
\multicolumn{11}{l}{\textit{Panel B: $\eta=3$}} \\
\addlinespace[2pt]
\multirow{2}{*}{Trend} & RL & 0.460 & 0.603 & 8.74 & 0.417 & 0.564 & 5.72 & 0.527 & 0.670 & 4.29 \\
 & Greedy & 0.384 & 0.543 & 9.38 & 0.527 & 0.684 & 4.45 & 0.366 & 0.537 & 3.18 \\
\addlinespace
\multirow{2}{*}{Trend+Cycle} & RL & 0.477 & 0.640 & 17.49 & 0.472 & 0.629 & 10.17 & 0.561\textsuperscript{**} & 0.715 & 7.15 \\
 & Greedy & 0.441 & 0.602 & 17.01 & 0.422 & 0.589 & 10.81 & 0.342 & 0.511 & 7.95 \\
\addlinespace
\multirow{2}{*}{\begin{tabular}[c]{@{}l@{}}Trend+Cycle\\+Macro\end{tabular}} & RL & 0.247 & 0.427 & 24.48 & 0.210 & 0.391 & 13.83 & 0.295\textsuperscript{**} & 0.473 & 9.86 \\
 & Greedy & 0.189 & 0.371 & 20.51 & 0.191 & 0.373 & 12.72 & 0.064 & 0.243 & 8.74 \\
\addlinespace
\multirow{4}{*}{Fixed} & MOM & 0.271 & 0.445 & 0.00 & 0.271 & 0.445 & 0.00 & 0.271 & 0.445 & 0.00 \\
 & HML & 0.137 & 0.315 & 0.00 & 0.137 & 0.315 & 0.00 & 0.137 & 0.315 & 0.00 \\
 & RMW & 0.198 & 0.417 & 0.00 & 0.198 & 0.417 & 0.00 & 0.198 & 0.417 & 0.00 \\
 & DEF & -0.435 & 0.010 & 0.00 & -0.435 & 0.010 & 0.00 & -0.435 & 0.010 & 0.00 \\
\bottomrule
\end{tabular}
\end{adjustbox}
\end{table}

\begin{figure}[htbp]
\caption{ Cumulative Out-of-Sample Rewards. The figure plots the cumulative monthly rewards, in percent, of the RL, Greedy, and four fixed-factor policies. Results are reported when the states are Trend $+$ Cycle, with $\eta=0$ and a switching cost of 120 bps. 
}
  \label{Figure_cumulative_rewards}
  \centering 
        \includegraphics[width=\textwidth]{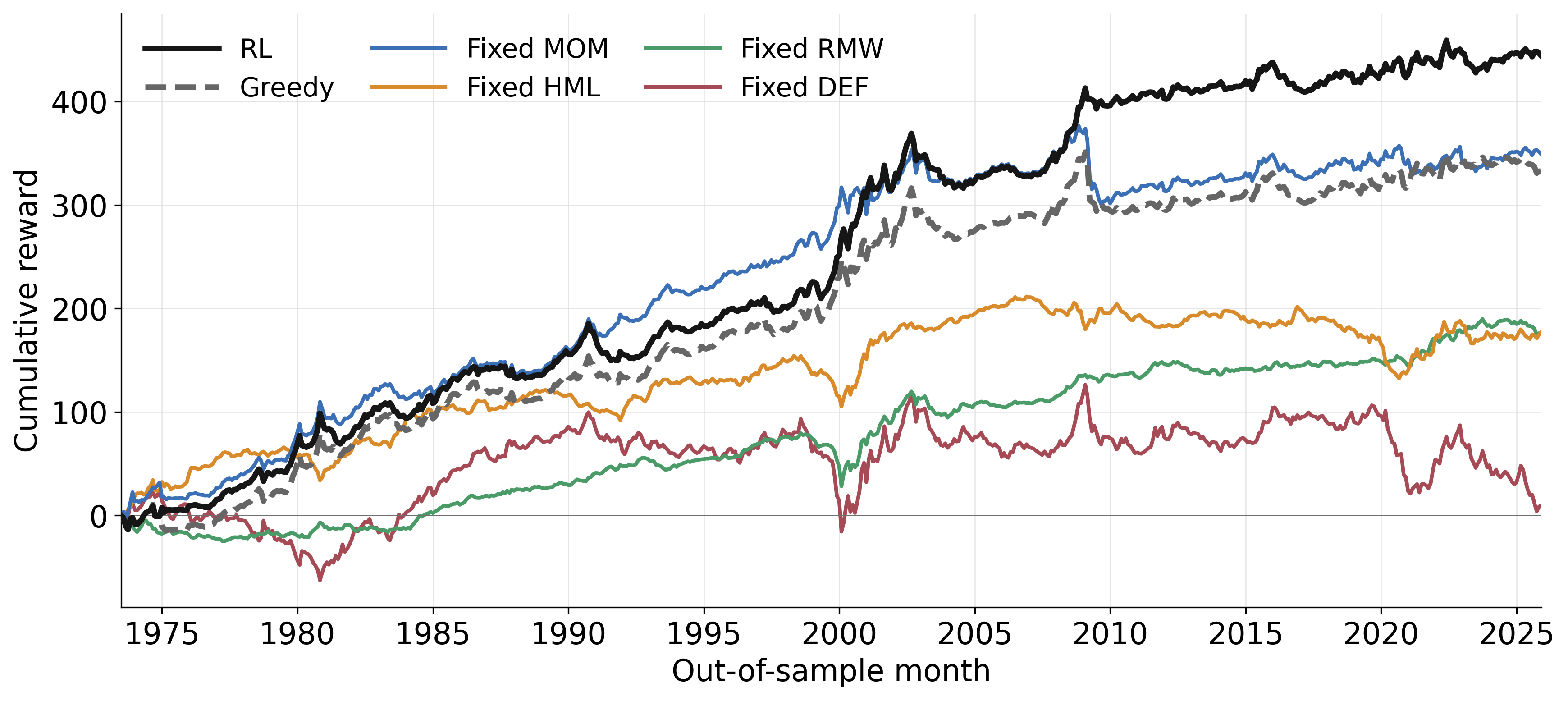} 
\end{figure}

Figure \ref{Figure_cumulative_rewards} plots cumulative out-of-sample rewards for the RL, Greedy, and fixed-factor policies under the Trend + Cycle states when $\eta=0$. So in this case, cumulative rewards are equivalent to cumulative net returns. The RL policy achieves the highest cumulative reward overall. It outperforms the Greedy policy over the sample and overtakes the fixed MOM policy around 2000. The latter further experiences a sharp decline during the 2008 financial crisis, reflecting the momentum crash.

\begin{figure}[htbp]
\caption{State Dependent Factor Rotations. Results are reported when the states are Trend $+$ Cycle, with $\eta=3$ and a switching cost of 120 bps. Panel (a) displays the percentages of out-of-sample months in which the RL policy selects among the MOM, HML, RMW, and DEF strategies, conditional on the market states classified by the momentum turning points. The right sub-plot in Panel (a) shows the destination factor conditional on exiting momentum across different market states. $N$ denotes the number of out-of-sample months in each market state, whereas $n$ denotes the number of months in which the policy exits MOM in each market state. Percentages may not sum to 100 due to rounding. Panel (b) shows the factor selected by the RL policy over the out-of-sample period.}
  \label{Figure_factor_rotation_eta3_120bps}
  \centering
    \begin{subfigure}{\textwidth}
        \centering
        \includegraphics[width=\textwidth]{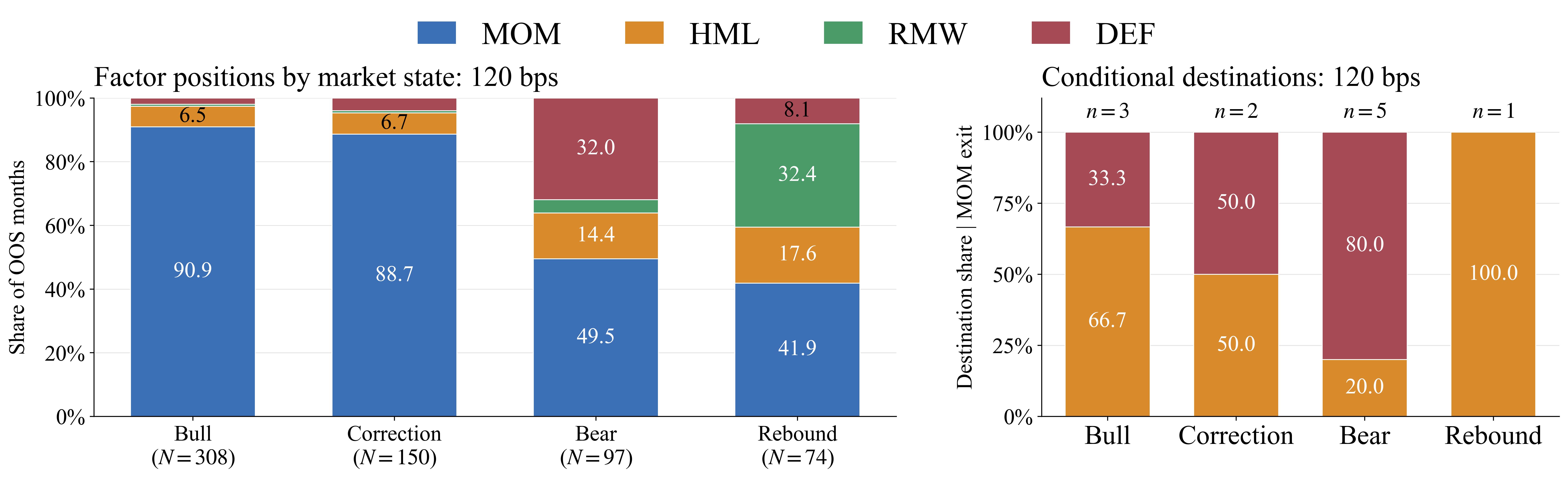}
        \caption{Market state dependent factor selection} 
    \end{subfigure}  
    \begin{subfigure}{\textwidth}
        \centering
        \includegraphics[width=\textwidth]{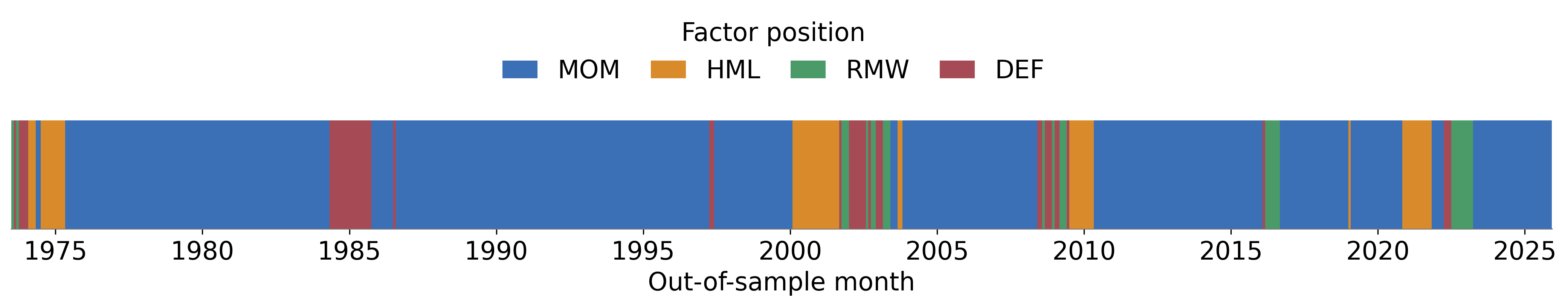}
        \caption{Factor rotation timeline} 
    \end{subfigure}
\end{figure}

Focusing on the RL policy, we find that momentum is the dominant factor in the out-of-sample allocations. We then study when investors should remain in momentum, rotate into another factor, or re-enter momentum as market conditions change. This extends Daniel and Moskowitz (2016), who mainly identify market conditions associated with momentum crash risk. 
Figure \ref{Figure_factor_rotation_eta3_120bps} shows the distribution of factor selections across market states in the out-of-sample period, as well as the destination factors chosen when the policy switches out of momentum. The results for $\eta=3$ and regularisation costs of 120 basis points are displayed.  

The upper sub-figure shows that MOM dominates the chosen factor positions in bull markets and during corrections, but its share falls to about half in bear and rebound months. The alternative factor positions vary systematically across market-cycle states. DEF is the second most frequently selected position in bear markets, whereas RMW is the second most frequently selected during rebounds. HML is selected less often and is spread across states rather than being associated with a particular market state.
Conditional on exiting MOM, the RL policy primarily switches to HML and DEF. During bull markets, two of the three exits transition to HML, compared with one of the two exits during corrections; all remaining exits transition to DEF. During bear markets, most exits are to DEF, with only one transition to HML. During rebounds, the sole exit is to HML. Overall, exits to DEF are concentrated in correction and bear markets, whereas exits to HML are more common during bull and rebound markets. Results at 20 and 80 bps do not alter these findings, except that the policy switches more often, leading to more exits from MOM and the emergence of rotations into RMW during bull and rebound markets. We discuss these patterns in Section \ref{Section_empirical} of the Electronic Supplement.

The lower sub-figure displays factor rotations over the out-of-sample period. Overall, together with Figure \ref{Figure_cumulative_rewards}, these results suggest that the policy rotates out of momentum defensively when market conditions deteriorate, shifts toward value or profitability during rebounds, and rarely exits momentum during bull markets. Re-entry into momentum under the RL policy is most concentrated in bull markets, followed by correction markets, rebounds, and bear markets. 

\section{Analysis of the Algorithm under Simplifying Conditions\label{Section_asymptoticsSimple}}

We provide here a simplified analysis of the algorithm that yields a transparent convergence result. We first state the assumptions underlying the general theoretical framework and then impose an additional condition eliminating approximation error. The Electronic Supplement relaxes this condition and provides the more general results.

\subsection{Additional Notation}

We use the notation introduced in Section \ref{Section_algorithmMainText}. For each $a\in\mathcal A$, let $\theta^a(S_t)$ summarise how strategy $a$, given the current state $S_t$, determines the next-period reward. For example, it may represent portfolio weights or a model prediction. Then, we write the reward as $R_{t+1}^{a}=R\left(S_{t+1}^{a};\theta^{a}\left(S_{t}\right)\right)$, where, conditioning on $\theta^{a}$, $R\left(s;\theta^{a}\left(r\right)\right)$ is a deterministic, possibly unknown function of $r,s\in\mathcal{S}$. This technical notation allows us to use operators and prove rather general results. In practical terms it is not restrictive. We make two remarks. First, treating the reward as a deterministic function of the state is without loss of generality for our purposes. The state can always be augmented so that the sigma-field generated by $S_{t+1}^{a}$ and $\theta^{a}\left(S_{t}\right)$ contains the sigma-field generated by the reward. The reward can then be represented as a measurable function $R\left(S_{t+1}^{a};\theta^{a}\left(S_{t}\right)\right)$. The enlargement of the state must, however, preserve the exogeneity condition of the raw state: the choice of $a$ at any point in time does not affect the evolution of the raw state $\left(\tilde{S}_{t}\right)$. This requirement is not restrictive. Any additional randomness needed to represent the reward can be included through exogenous variables whose distribution does not depend on the chosen strategy. Second, the role of $\theta^{a}\left(S_{t}\right)$ is to make explicit how the reward depends on the strategy being selected. For clarity, consider the following simple example. Let $\theta^{a}\left(s\right)$, where $s=\left(\tilde{s},a'\right)$, depend only on $a$ and $a'$; hence it is invariant with respect to $\tilde{s}$. Then, $\theta^{a}\left(s\right)$ records the strategy $a'$ used in the previous period and the strategy $a$ selected in the current period. The reward can therefore depend on both the currently selected strategy and the strategy from which we are switching. In more general applications, $\theta^{a}\left(S_{t}\right)$ may also contain state-dependent quantities such as portfolio weights or model outputs. Next, we give an example (see Section \ref{Section_remarksConditions} for more examples).

\begin{example}\label{Example_thetaRewardCharateristicPortfolios}(Characteristic-based portfolios). Let $Y_{t}=\left(Y_{t,1},\ldots,Y_{t,N}\right)^{\mathrm{T}}$ be the vector of returns on $N$ assets at time $t$, and let $X_{t}$ be an $N\times|\mathcal{A}|$ matrix of asset characteristics, where both $Y_{t}$ and $X_{t}$ are included in the raw state $\tilde{S}_{t}$. Some entries of $X_{t}$ may be unavailable and are recorded as N/A. For each $a\in\mathcal{A}$, let $w^{a}\left(\cdot\right)$ be a known deterministic rule that maps $X_{t}$ into portfolio weights using its $a$th column. For example, $w^{a}\left(\cdot\right)$ may remove assets for which the characteristic is unavailable, sort the remaining assets according to the $a$th column of $X_{t}$, and assign equal positive weights to assets in the upper quantile and equal negative weights to assets in the lower quantile. Define $\theta^{a}\left(S_{t}\right)=w^{a}\left(X_{t}\right)$. Thus, $\theta^{a}\left(S_{t}\right)$ is the vector of portfolio weights of the characteristic-sorted portfolio associated with strategy $a$. The reward from selecting strategy $a$ at time $t$ is $R\left(S_{t+1}^{a};\theta^{a}\left(S_{t}\right)\right)=Y_{t+1}^{\mathrm{T}}\theta^{a}\left(S_{t}\right)$. In this example, $\theta^{a}\left(\cdot\right)$ is not estimated, but is a known deterministic map from the state into portfolio weights. Its value is nevertheless state-dependent because the portfolio weights vary with the cross-section of characteristics in $X_{t}$.

\end{example}

We also need to define some classes of functions. The space of approximating functions for the action-value function is 
\begin{equation}
\mathcal{F}:=\left\{ f:f\left(s\right)=b^{\mathrm{T}}\varphi\left(s\right),s\in\mathcal{S},b\in\mathbb{R}^{K}\right\}.\label{EQ_spaceFunctionsMainText}
\end{equation}
Because the theory uses capping at $B$, we define the capped class
\[
\mathcal{F}_{B}:=\left\{ h=\left[f\right]_{B}:f\in\mathcal{F}\right\} ,
\]
which are capped versions of the linear functions in $\mathcal{F}$ with range in $\left[-B,B\right]$. Then we define 
\[
\mathcal{F}_{B}^{\max}:=\left\{ h\left(s\right)=\max_{a\in\mathcal{A}}f_{a}\left(s\right):f_{a}\in\mathcal{F}_{B},s\in\mathcal{S},a\in\mathcal{A}\right\} 
\]
The latter is the class of value functions obtained by taking the maximum across approximating action-value functions. For example, suppose we have three strategies. Then, $\mathcal{A}=\left\{ a_{1},a_{2},a_{3}\right\}$, and the algorithm produces three estimators of the action-value function in $\left\{ f_{a}\in\mathcal{F}_{B},a\in\mathcal{A}\right\} $. The value function is approximated by the maximum of these functions and belongs to $\mathcal{F}_{B}^{\max}$. We state a set of sufficient conditions tailored to the main convergence result. Weaker and more general conditions are given in the Electronic Supplement.

\subsection{Simplified Assumptions\protect\label{Section_assumptionsMainText}}

In what follows, recall the notation $R_{t+1}^{a}:=R\left(S_{t+1}^{a};\theta^{a}\left(S_{t}\right)\right)$

\begin{condition}\label{Condition_stateActionPair}(State Action Pair) The simulated sequence is i.i.d. uniformly distributed in $\mathcal{A}$ which has bounded cardinality, $S_{t}=\left(\tilde{S}_{t},A_{t-1}\right)$ where $\tilde{S}_{t}$ takes values in some set $\tilde{\mathcal{S}}$, and $\left(\tilde{S}_{t}\right)_{t\geq1}$ is independent of $\left(A_{t}\right)_{t\geq0}$.\end{condition}

\begin{condition}\label{Condition_mixing}(Markov Assumption) The sequence of states $S:=\left(S_{t}\right)_{\geq1}$ is a strictly stationary Markov process with exponentially decaying beta mixing coefficients.\end{condition}

\begin{condition}\label{Condition_moment}(Moments) We have that $\max_{a\in\mathcal{A}}\mathbb{E}\left|R\left(S_{t+1}^{a};\theta^{a}\left(S_{t}\right)\right)\right|^{v}<\infty$ for $v\geq2$.\end{condition}

\begin{condition}\label{Condition_theta}(Rewards Parameter) The map $\mathcal{S}\ni s\mapsto\theta^{a}\left(s\right)$ is possibly random, but independent of the sequence of states $\left(S_{t}\right)_{t\geq1}$ and measurable at time $t=0$.\end{condition}

\begin{condition}\label{Condition_functionClass}(Approximating Function Class) The functions $\varphi=\left\{ \varphi_{l}:l=1,2,...,K\right\} $ in (\ref{EQ_spaceFunctions}) are uniformly bounded and the population Gram matrix $P\varphi\varphi^{\mathrm{T}}$ is positive definite, where $\varphi$ is arranged as a column vector. Moreover, $K\geq2$, $\sup_{s\in\mathcal{S}}\varphi\left(s\right)^{\mathrm{T}}G^{-1}\varphi\left(s\right)=O\left(K\right)$ such that $r_{n}=\sqrt{n^{-1}K\left(\ln K\right)\left(\ln n\right)}\rightarrow0$.
\end{condition}

We refer to the following as the Simplified Assumptions to distinguish them from the slightly more general Assumptions in Section \ref{Section_assumptions} of the Electronic Supplement. 

\subsection{A Convergence Result \label{Section_convergenceResultSimple}}

In what follows, $V^{\left(J\right)}:=V^{\pi_{J}}$, where $V^{\pi}$ is the true value function of following policy $\pi$ (as in (\ref{EQ_valueFunctionMainText})) and $\pi_{J}$ is the policy estimated using the estimation algorithm in Section \ref{Section_algorithmMainText} after $J$ iterations. Furthermore, recall that $V^{*}$ is the true optimal value function (as in (\ref{EQ_optimalValueFunctionMainText})). For simplicity, the bound does not account for $\left|\mathcal{A}\right|$. This is equivalent to $\left|\mathcal{A}\right|$ being relatively small, which is the case for the applications that we have in mind. If the functions $\left\{ \theta^{a}:a\in\mathcal{A}\right\}$ are random but satisfy Assumption \ref{Condition_theta}, the results that follow are understood to be conditional on $\left\{ \theta^{a}:a\in\mathcal{A}\right\} $.

We state a result under the assumption of zero approximation error for the class of Bellman equations. More general results are stated in Section \ref{Section_convergenceResults} of the Electronic Supplement and not here to avoid the use of complex notation. In what follows, recall that $\gamma$ is the discount rate, $\rho$ is the ridge penalty, $K$ is the dimension of the vector space and $B$ is the capping value.

\begin{theorem}\label{Theorem_MainCorollarywithZeroDistance} Suppose that the Assumptions in Section \ref{Section_assumptionsMainText} hold. Assume that 
\[
\mathbb{E}\left[R_{t+1}^{a}+\gamma f\left(S_{t+1}^{a}\right)|S_{t}=s\right]\in\mathcal{F}
\]
for any $f\in\mathcal{F}_{B}^{\max}$ and $a\in\mathcal{A}$, and that $\sup_{t\geq0}\mathbb{E}\left[\left|R_{t+1}^{a}\right|S_{t}\right]\leq\left(1-\gamma\right)B$, $\gamma\in\left(0,1\right)$, for fixed $B$. Then, for every $J\geq1$,
\[
\left|V^{*}-V^{\left(J\right)}\right|_{2,P}=O_{P}\left(\left(1+B\right)\left[r_{n}+\rho\left|G^{-1}\right|_{{\rm op}}+\gamma^{J}\right]\right).
\]
\end{theorem}

The theorem assumes that conditional expectation of the rewards, not the rewards themselves, is bounded and in the span of the basis functions $\varphi_{l}$. The other assumption is that the conditional expectation of the functions in $\mathcal{F}_{B}^{\max}$ is also in this linear span. This is usually a property of the transition kernel (see Example \ref{Example_zeroApproximationError} in the Appendix). If the basis functions are orthonormal with respect to the law $P$, then $\left|G^{-1}\right|_{{\rm op}}=O\left(1\right)$. For the B-spline basis in Corollary \ref{Corollary_withPolynomialApprox} of the Electronic Supplement, we have $\left|G^{-1}\right|_{{\rm op}}=O\left(K\right)$. Theorem \ref{Theorem_MainCorollarywithZeroDistance} is a corollary to a more general result stated in Section \ref{Section_convergenceResults} of the Electronic Supplement. In Corollary \ref{Corollary_withPolynomialApprox} of the Electronic supplement we show that it is sufficient to satisfy the condition of zero approximation error asymptotically for $K\rightarrow\infty$ choosing a suitable set of functions $\varphi$, as in the case of sieve estimators. We also provide a bound when the functions $\left\{ \theta^{a}:a\in\mathcal{A}\right\}$ are estimated (Corollary \ref{Corollary_withEstimatedTheta} in the Electronic Supplement). These results require more complex notation and the use of operators. For this reason, we only state the simpler Theorem \ref{Theorem_MainCorollarywithZeroDistance} in the main text.

\section{Conclusion\protect\label{Section_conclusion}}
We develop a framework for dynamic strategy selection when relative performance depends on observable state variables and the selection rule must be estimated. A central implication is that switching costs can play an important role as a regularisation device. Actual transaction costs may be too small to prevent switches driven by estimation error, so the switching penalty used to estimate the policy may need to be substantially larger than the economic cost of trading. This distinction is important in applications in which small estimated differences across strategies can otherwise generate unstable decisions.

Our empirical application to U.S. equity factors illustrates several consequences of this framework. First, sufficiently strong regularisation materially improves out-of-sample performance. Second, allowing rewards to reflect risk preferences can lead to substantial improvements in risk-adjusted returns. Third, accounting for continuation values through RL can outperform the myopic Greedy policy. This difference is not explained by lower switching frequency. In our best-performing specification, RL and Greedy switch with similar frequency, but RL delivers a substantially higher Sharpe ratio. Moreover, these Sharpe ratios are computed after deducting the full switching penalty, which is deliberately set larger than expected actual transaction costs.
Implementing the same estimated policy at a lower actual transaction cost would therefore yield higher net returns, so our reported results provide a conservative assessment of the gains from dynamic strategy selection.
The estimated policy also identifies economically interpretable factor rotations, including systematic exits from and re-entries into momentum as market conditions change. 

The theoretical results support this empirical evidence. We show that, under suitable conditions, the value of the estimated policy converges to the unknown optimal value, with a rate that can be close to the parametric rate. In the simulations reported in Section \ref{Section_simulations} of the Electronic Supplement, as the estimation sample increases, the RL policy approaches the performance of the oracle policy that knows the data-generating process. On the other hand, the Greedy policy does not exhibit the same improvement. The simulations also illustrate the cost of enlarging the state space with irrelevant predictors. This reinforces the empirical finding that additional information can be outweighed by additional estimation error. 

More generally, the framework applies whenever a decision maker repeatedly selects among competing strategies and estimated differences in their conditional performance may lead to excessive switching. A natural extension is to incorporate selection of the state variables directly into the estimation procedure.  Doing so would make the asymptotic analysis more involved and could itself be the subject of future research.

\section*{Appendix\protect\label{Section_Appendix}}

\setcounter{figure}{0} 
\global\long\def\thefigure{A.\arabic{figure}}%

\setcounter{table}{0} 
\global\long\def\thetable{A.\arabic{table}}%

\setcounter{equation}{0} 
\global\long\def\theequation{A.\arabic{equation}}%

\setcounter{lemma}{0} 
\global\long\def\thelemma{A.\arabic{lemma}}%

\setcounter{equation}{0} 
\global\long\def\theexample{A.\arabic{example}}%

\setcounter{section}{0} 
\global\long\def\thesection{A.\arabic{section}}%

\section{Remarks on Assumptions\protect\label{Section_remarksConditions}}

\paragraph{Assumption \ref{Condition_stateActionPair}. }

In econometric terms, the condition implies that $\left(\tilde{S}_{t}\right)_{t\geq1}$
is exogenous and given. The i.i.d. uniformly distributed condition
is used for simplicity of exposition. In Section \ref{Section_stateActionPair}
of the Electronic Supplement we allow for dependent $\left(A_{t}\right)$
where each $A_{t}$ is not uniformly distributed. This is beneficial
if we want to place different weight on different strategies. For
example, we may know that a strategy is usually preferred to others
and so should have a higher weight, i.e. $A_{t}$ should take the
value associated to it more frequently. 

\paragraph{Assumption \ref{Condition_mixing}.}

The beta mixing coefficient can be defined as 
\[
\beta_{q}:=\frac{1}{2}\sup\sum_{A_{i}\in\mathcal{C},B_{j}\in\mathcal{D}}\left|\Pr\left(A_{i}\cap B_{j}\right)-\Pr\left(A_{i}\right)\Pr\left(B_{j}\right)\right|,\,q\geq1
\]
where the supremum is over all partitions of the sample space, and
$\mathcal{C}$ is the sigma algebra generated by $\left(S_{s}\right)_{s\leq t}$
and $\mathcal{D}$ is the sigma algebra generated by $\left(S_{s}\right)_{s\geq t+q}$.
It is worth noting that the condition with exponentially decaying
coefficients is satisfied by many Markov chains with transition distribution
possessing a density w.r.t. the Lebesgue measure (Doukhan, 1995, Ch.2.4).

\paragraph{Assumption \ref{Condition_moment}.}

The moment condition on the reward is the weakest possible in this setup, and is satisfied as soon as the rewards have finite variance. 

\paragraph{Assumption \ref{Condition_theta}.}

The functions $\left\{ \theta^{a}:a\in\mathcal{A}\right\} $ can be
estimated, but are assumed to be independent of the states in order
to keep the analysis of the problem manageable. 

\begin{example}\label{Example_linearModelThetaEstimatedIndependent}
Let $X_{t}$ be a vector of predictors for a random variable $Y_{t+1}$. Consider the models $X_{t}^{{\rm T}}\beta^{a}$ for predicting $Y_{t+1}$ where $\beta^{a}$ is a restricted vector of regression coefficients for each $a\in\mathcal{A}$. Suppose that the value of $\beta^{a}$ and the restrictions are not exactly known for each $a\in\mathcal{A}$. For example, restrictions could include certain sparsity constraints (Lasso regression), constraints within an ellipsoid (ridge regression), or being inside a simplex. Suppose that $\left(Y_{t+1}',X_{t}'\right)_{t\geq1}$ has the same distribution as $\left(Y_{t+1},X_{t}\right)_{t\geq1}$, but it is independent of it. A sample from $\left(Y_{t+1}',X_{t}'\right)_{t\geq1}$ is used to estimate $\beta^{a}$ and the restrictions. The estimator is denoted by $\hat{\beta}^{a}$. We define $\theta^{a}\left(S_{t}\right)=X_{t}^{{\rm T}}\hat{\beta}^{a}$. Then, the rewards are constructed conditional on $\left\{ \hat{\beta}^{a}:a\in\mathcal{A}\right\} $, which implicitly means conditioning on a statistic from a sample from $\left(Y_{t+1}',X_{t}'\right)_{t\geq1}$.\end{example}

In some problems,
estimation of the model is based on the full sample where a second
independent sample is not available. In this case, we write $\left\{ \hat{\theta}^{a}:a\in\mathcal{A}\right\} $
to stress the fact that this is now sample dependent. Under some additional
technical assumptions that include uniform convergence of $\hat{\theta}^{a}$
towards a nonstochastic element $\theta^{a}$ and a Lipschitz continuity
condition on $R$, we shall be able to derive consistency of our procedure
(Corollary \ref{Corollary_withEstimatedTheta}). It is worth noting
that the goal is to obtain a policy that is close to the optimal one,
where optimal is for a value function based on the rewards $R\left(S_{t+1};\theta^{a}\left(S_{t}\right)\right)$
where $\theta^{a}$ is either nonstochastic or measurable at time
$t=0$ and independent of the sample, but otherwise arbitrary. We
illustrate this with a simple variation of Example \ref{Example_linearModelThetaEstimatedIndependent}.

\begin{example}\label{Example_linearModelThetaEstimated}Suppose
that $X_{t}$ is a subset of the elements in $S_{t}$. Consider estimated
prediction models $X_{t}^{{\rm T}}\hat{\beta}^{a}$, where $\hat{\beta}^{a}$
is an estimator based on the sample data $\left\{ S_{1},S_{2},...,S_{n+1}\right\} $,
$a\in\mathcal{A}$. We set $\hat{\theta}^{a}\left(S_{t}\right)=\left(X_{t}^{{\rm T}}\hat{\beta}^{a},X_{t}^{{\rm T}}\hat{\beta}^{A_{t-1}}\right)$,
$a\in\mathcal{A}$. Information about $A_{t-1}$ is included in $S_{t}=\left(\tilde{S}_{t},A_{t-1}\right)$, and we may find it useful to keep track of $X_{t}^{{\rm T}}\hat{\beta}^{A_{t-1}}$.
Suppose that $X_{t}$ is uniformly bounded and that $\max_{a\in\mathcal{A}}\left|\hat{\beta}^{a}-\beta^{a}\right|_{2}=O_{P}\left(n^{-1/2}\right)$,
where $\beta^{a}$ and $\left|\cdot\right|_{2}$ is the Euclidean
norm. Define $\theta^{a}\left(S_{t}\right)=\left(X_{t}^{\mathrm{T}}\beta^{a},X_{t}^{\mathrm{T}}\beta^{A_{t-1}}\right)$
and similarly for $\hat{\theta}^{a}$ with $\beta^{a}$ replaced by
$\hat{\beta}^{a}$. We have that 
\[
\max_{a\in\mathcal{A}}\left|\hat{\theta}^{a}-\theta^{a}\right|_{\infty}\leq\sup_{s\in\mathcal{S}}\left|X\left(s\right)\right|_{2}\max_{a\in\mathcal{A}}\left|\hat{\beta}^{a}-\beta^{a}\right|_{2}=O_{P}\left(n^{-1/2}\right).
\]

Our methodology will produce an estimator for the optimal policy for
the unobservable reward function $R\left(T^{a}\cdot;\theta^{a}\left(\cdot\right)\right)$.
The rewards are unobservable because we can only use the estimated
reward function $R\left(T^{a}\cdot;\hat{\theta}^{a}\left(\cdot\right)\right)$.\end{example}

\paragraph{Assumption \ref{Condition_functionClass}.}

Writing $\mathcal{S}=\tilde{\mathcal{S}}\times\mathcal{A}$, the simplest
possible set of functions is given by linear functions in $\tilde{s}\in\tilde{\mathcal{S}}=\mathbb{R}^{L}$
times indicators of the action $a'\in\mathcal{A}$. In this case,
Algorithm \ref{Algorithm_Sieve} loops over $a\in\mathcal{A}$ and
for each $a$ estimates $f_{a}\in\mathcal{F}$ (as in (\ref{EQ_spaceFunctions}))
where

\begin{equation}
f_{a}\left(\left(\tilde{s},a'\right)\right)=\sum_{l=1}^{L}\sum_{a''\in\mathcal{A}}\tilde{s}_{l}1_{\left\{ a'=a''\right\} }b_{l,a''}^{a}.\label{EQ_generalLinearSpecification}
\end{equation}
Here, $f_{a}\left(\left(\tilde{s},a'\right)\right)$ is an approximation
for $Q\left(\left(\tilde{s},a'\right),a\right)$. In this case, $K=L\left|\mathcal{A}\right|$.

For $\mathcal{A}:=\left\{ a_{1},a_{2},...,a_{\left|\mathcal{A}\right|}\right\} $,
an additive specification can be written as 
\begin{equation}
f_{a}\left(\left(\tilde{s},a'\right)\right)=\sum_{l=1}^{L}\tilde{s}_{l}b_{l}^{a}+\sum_{j=1}^{\left|\mathcal{A}\right|}b_{L+j}^{a}1_{\left\{ a'\neq a_{j}\right\} }\label{EQ_linearSpecification}
\end{equation}
so that $K=L+\left|\mathcal{A}\right|$.

It is simple to see that these specifications are in (\ref{EQ_spaceFunctions})
for suitable definition of the functions $\varphi_{k}$. The linear terms in the summation over $l$ may also be replaced by polynomial terms, for example including both linear and squared terms, to capture nonlinearities while retaining linearity in the coefficients.  In some circumstances, we may wish for a more parsimonious specification, making the functions specific to each action over which we loop over in Algorithm \ref{Algorithm_Sieve}.
The notation does not account for this. However, this is a trivial extension and our results carry over to this special case. For the sake of definiteness, suppose that instead of $\mathcal{F}$ in (\ref{EQ_spaceFunctions})
we consider $\mathcal{F}^{a}$ where $f_{a}\in\mathcal{F}^{a}$ if
\begin{equation}
f_{a}\left(\left(\tilde{s},a'\right)\right)=\sum_{l=1}^{L}\tilde{s}_{l}b_{l}^{a}+b_{L+1}^{a}1_{\left\{ a'\neq a\right\} },\label{EQ_linearSpecificationSimple}
\end{equation}
The function class is now dependent on $a\in\mathcal{A}$ via the
map $a'\mapsto1_{\left\{ a'\neq a\right\} }$. It is not difficult
to see that this is indeed a restriction of (\ref{EQ_linearSpecification})
with an intercept. Given that the complexity of the class of functions
does not change with $a\in\mathcal{A}$, for notational simplicity
we use (\ref{EQ_spaceFunctions}) in our results.

The condition $\sup_{s\in\mathcal{S}}\varphi\left(s\right)^{\mathrm{T}}G^{-1}\varphi\left(s\right)=O\left(K\right)$
controls the maximum squared norm of the normalised basis functions.
Note that $P\varphi{}^{\mathrm{T}}G^{-1}\varphi=K$. Hence, the condition
requires the supremum to be of the same order as its population average.
The condition holds, for example, for uniformly bounded orthonormal
bases and for B-spline bases considered in Corollary \ref{Corollary_withPolynomialApprox}
in the Electronic Supplement.

We conclude with an example that satisfies the additional conditions
of Theorem \ref{Theorem_MainCorollarywithZeroDistance}. 

\begin{example}\label{Example_zeroApproximationError}Let $\tilde{S}_{t}\in\left[0,1\right]$
and $\mathcal{A}=\left\{ 1,2\right\} $. Suppose that $\tilde{S}_{t+1}$
conditional on $\tilde{S}_{t}=\tilde{s}$ has distribution $\left(1-h\left(\tilde{s}\right)\right)F_{0}+h\left(\tilde{s}\right)F_{1}$
where $F_{0}$ and $F_{1}$ are two distribution functions on $[0,1]$
and $h$ is a function in $\mathcal{F}$ with values in $[0,1]$. To define
the rewards, consider the augmented raw state $\left(\tilde{S}_{t},\varepsilon_{t}\right)$
where $\left(\varepsilon_{t}\right)$ is a sequence of i.i.d. random
variables independent of $\left(S_{t}\right)$ with mean zero and
finite variance. Consider rewards 
\[
R_{t+1}^{a}=\alpha_{R}^{a}+\beta_{R}^{a}\tilde{S}_{t}+\varepsilon_{t+1}-c1_{\left\{ a\neq A_{t-1}\right\} }.
\]
The conditional expectation of $R_{t+1}^{a}$ is linear in $\tilde{S}_{t}$
and $1_{\left\{ a\neq A_{t-1}\right\} }$. We suppose that we only
use $\tilde{S}_{t}$ for the predictions. For any $f\in\mathcal{F}_{B}^{\max}$,
\begin{align*}
\mathbb{K}^{a}f\left(\left(\tilde{s},a\right)\right)= & \left(1-h\left(\tilde{s}\right)\right)\int f\left(\left(\tilde{s},a\right)\right)dF_{0}\left(\tilde{s}\right)+h\left(\tilde{s}\right)\int f\left(\left(\tilde{s},a\right)\right)dF_{1}\left(\tilde{s}\right)\\
= & \alpha_{f}^{a}+b_{f}^{a}h\left(\tilde{s}\right)
\end{align*}
for suitable coefficients. Hence, if the constant, the linear function
and $1_{\left\{ a\neq\cdot\right\} }$ is also one of the basis functions
$\varphi_{l}$, we clearly have that $\mathbb{T}^{a}f=\mathbb{K}^{a}R+\gamma\mathbb{K}^{a}f\in\mathcal{F}$.
Since the raw state is bounded, we also have $|\mathbb{T}^{a}f|_{\infty}\le B$
for every $a\in\mathcal{A}$ and $f\in\mathcal{F}_{B}^{\max}$ for
$B$ large enough. \end{example}

\newpage{}

\section*{Electronic Supplement to ``Dynamic Strategy Selection'' by F.
Cordoni, A. Sancetta and Y. Zhao\protect\label{Section_Supplement}}

\setcounter{figure}{0} 
\global\long\def\thefigure{S.\arabic{figure}}%

\setcounter{table}{0} 
\global\long\def\thetable{S.\arabic{table}}%

\setcounter{equation}{0} 
\global\long\def\theequation{S.\arabic{equation}}%

\setcounter{lemma}{0} 
\global\long\def\thelemma{S.\arabic{lemma}}%

\setcounter{lemma}{0} 
\global\long\def\thecorollary{S.\arabic{corollary}}%

\setcounter{theorem}{0}
\global\long\def\thetheorem{S.\arabic{theorem}}%

\setcounter{lemma}{0} 
\global\long\def\thealgorithm{S.\arabic{algorithm}}%

\setcounter{condition}{0} 
\global\long\def\thecondition{S.\arabic{condition}}%

\setcounter{lemma}{0} 
\global\long\def\theexample{S.\arabic{example}}%

\setcounter{page}{1}

\setcounter{section}{0} 
\global\long\def\thesection{S.\arabic{section}}%

\section{The Decision Problem }

At each point in time, the expected total discounted reward is given by the value function 
\begin{equation}
V^{\pi}\left(s\right)=\mathbb{E}^{\pi}\left[\sum_{i=1}^{\infty}\gamma^{i-1}R_{t+i}^{\pi}|S_{t}^{\pi}=s\right]\label{EQ_valueFunction}
\end{equation}
for some $\gamma\in\left(0,1\right)$. Here, $s=\left(\tilde{s},a\right)\in\mathcal{S}$ so that $S_{t}^{\pi}$ is the state where the previous action is fixed to $a\in\mathcal{A}$. The expectation is with respect to (w.r.t.) a Markov transition distribution under the policy $\pi$. We rely on a number of operators in order to describe the problem in a Markovian framework. Let $\mathbb{K}^{a}$ be the Markov decision operator with kernel $\kappa^{a}$ so that for any map $\mathcal{S}\ni s\mapsto f\left(s\right)$, $\mathbb{K}^{a}f\left(r\right)=\int_{\mathcal{S}}f\left(s\right)\kappa^{a}\left(r,ds\right)$. In our context this means that integration is w.r.t. $s$ keeping the last entry in $s$ fixed to $a$. Its interpretation is as the expectation of $f\left(S_{t+1}^{a}\right)$ conditional on $S_{t}^{a'}=r$ for some $a'\in\mathcal{A}$, whose value is implicitly set according to $r$. Finally, we use $\mathbb{K}^{\pi}f$ for the operator s.t. $\mathbb{K}^{\pi}f\left(r\right)=\int_{\mathcal{S}}f\left(s\right)\kappa^{\pi}\left(r,ds\right)$ where integration is taken keeping the last entry in $s$ equal to $\pi\left(r\right)$. We overload the notation and for the reward function we write $\mathbb{K}^{\pi}R\left(r\right)=\int_{\mathcal{S}}R\left(s;\theta^{\pi}\left(r\right)\right)\kappa^{\pi}\left(r,ds\right)$, $r\in\mathcal{S}$. With this notation, we have that $V^{\pi}\left(s\right)=\sum_{t=1}^{\infty}\gamma^{t-1}\left(\left(\mathbb{K}^{\pi}\right)^{t}R\right)\left(s\right)$
where $\left(\mathbb{K}^{\pi}\right)^{t}=\mathbb{K}^{\pi}\left(\mathbb{K}^{\pi}\right)^{t-1}$
and $\left(\mathbb{K}^{\pi}\right)^{0}$ is the identity. Note that
we are implicitly assuming stationary states. Then, \textbf{$V^{\pi}$
}satisfies the Bellman equation $V^{\pi}\left(s\right)=\mathbb{K}^{\pi}R\left(s\right)+\gamma\mathbb{K}^{\pi}V^{\pi}\left(s\right)$.
The optimal policy is the one that maximizes these quantities w.r.t.
$\pi$ within a given set of policies. We write 
\begin{equation}
V^{*}\left(s\right)=\max_{\pi}\sum_{t=1}^{\infty}\gamma^{t-1}\left(\left(\mathbb{K}^{\pi}\right)^{t}R\right)\left(s\right)\label{EQ_optimalValueFunction}
\end{equation}
where the maximum is over any policy on $\mathcal{S}$ such that the
above is well defined. In our setup this will always be well defined.
The Bellman equation says that 
\begin{equation}
V^{\pi}\left(s\right)=\mathbb{K}^{\pi}R\left(s\right)+\gamma\mathbb{K}^{\pi}V^{\pi}\left(s\right).\label{EQ_BellmanEq}
\end{equation}
With this notation, we introduce the concept of a $Q$ function: 
\begin{equation}
Q^{\pi}\left(s,a\right)=\mathbb{K}^{a}R\left(s\right)+\gamma\mathbb{K}^{a}V^{\pi}\left(s\right).\label{EQ_QRecursion}
\end{equation}
This means that we are in state $s$ and for that time only we shall
take decision $a\in\mathcal{A}$ rather than decision $\pi\left(s\right)$.
For example, $Q^{\pi}\left(s,a\right)-V^{\pi}\left(s\right)$ tells
us the total difference incurred by using $a\in\mathcal{A}$ instead
of $\pi$, for one period only, when we are in state $s$. Recall
that $s$ can include information on the previous action. By definition,
$Q^{\pi}\left(s,\pi\left(s\right)\right)=V^{\pi}\left(s\right)$.
For the optimal policy that gives (\ref{EQ_optimalValueFunction}), we have the following relation $V^{*}\left(s\right)=\max_{a\in\mathcal{A}}Q^{*}\left(s,a\right)$.

\section{Revisiting the Algorithm\protect\label{Section_algorithm}}

The algorithm requires to augment the data with a sequence of randomly
generated actions. This is to ensure that in the estimation stage
we can condition on a state that includes a previous action. This
ensures that we can estimate consistently our best action conditioning
on the state and some previous action. The recursive nature of the
problem requires us to do so. Next, we show how to augment the data
for this purpose. We then introduce operators that are used in the
implementation of the Algorithm. The Algorithm is finally introduced
in Section \ref{Section_algoDescription}.

\subsection{The Baseline State Action Pair\protect\label{Section_stateActionPair}}

For the purpose of estimation, we need to define what we call raw
states and the baseline state action pair. We suppose that the user
observes a sequence of raw states $\left(\tilde{S}_{t}\right)_{t\geq1}$
taking values in a set $\tilde{S}$. The raw states are exogenous
to the model selection process. We suppose that the user has simulated
a sequence of random actions $\left(A_{t}\right)_{t\geq0}$ with values
in $\mathcal{A}$. These actions are generated for estimation purposes only, and we shall mention how to construct them momentarily. We then
augment $\tilde{S}_{t}$ with $A_{t-1}$ and define $S_{t}=\left(\tilde{S}_{t},A_{t-1}\right)$.
For all practical purposes, we can think of $S_{t}$ as a vector where
the last entry is equal to $A_{t-1}$. Then, in Section \ref{Section_problemSetup},
when we wrote $S_{t}^{a}$ we meant $S_{t}^{a}=\left(\tilde{S}_{t},a\right)$.
Essentially, we do not use a superscript in $S_{t}^{a}$ when $a=A_{t-1}$.
The role of $S_{t}$ is as conditioning variable in our estimation
algorithm (Algorithm \ref{Algorithm_Sieve}, in Section \ref{Section_algoDescription}).

Now, we discuss how $\left(A_{t}\right)_{t\geq0}$ can be constructed.
Let $\mathcal{A}\times\mathcal{S}\ni\left(a,s\right)\mapsto\alpha\left(a|s\right)$
be a map such that $\alpha\left(\cdot|s\right)$ is a mass function
on $\mathcal{A}$ for all $s\in\mathcal{S}$. Then, we suppose that
we observe the state action pair $\left(S_{t},A_{t}\right)_{t\geq1}$,
where, conditioning on $S_{t}=s$, $A_{t}$ is a random variable with
mass function $\alpha\left(\cdot|s\right)$. The value of $A_{t}$
for $t=0$, $A_{0}$, is also a random variable on $\mathcal{A}$.
We require that there is a constant $\underline{\alpha}>0$ such that
\begin{equation}
\inf_{a\in\mathcal{A},s\in\mathcal{S}}\Pr\left(A_{t}=a|S_{t}=s\right)\geq\underline{\alpha}\text{ and }\inf_{a\in\mathcal{A}}\Pr\left(A_{0}=a\right)\geq\underline{\alpha}.\label{EQ_conditionStateActionPair}
\end{equation}
This means that all actions receive a nonzero probability irrespective
of the state. For this sequence of states, we shall not use any superscript.
Hence, we assume that the state action pair is given. In practice,
this sample is constructed in a simple way by the researcher.

\begin{example}\label{Example_stateActionPairDecomposition}The initial
sequence of raw states $\left(\tilde{S}_{t}\right)_{t\geq1}$ with
values in $\mathcal{\tilde{S}}\subset\mathcal{S}$ is given. The mass function $\alpha\left(\cdot|s\right)$ is the uniform distribution on $\mathcal{A}$ (i.e. independent of $s$), and $A_{0}$ is also a uniform random variable with values in $\mathcal{A}$. This means that the user specified an initial policy that is independent of the raw state. This choice is what we shall use in our empirical illustration in Section \ref{Section_empirical}. It just requires us to generate i.i.d. uniform random variables in $\mathcal{A}$.\end{example}

With no further notice, in what follows, we shall use the notation
introduced here, and decompose $S_{t}$ as $\left(\tilde{S}_{t},A_{t-1}\right)$
when applicable, i.e. we shall always write $S_{t}^{a}=S_{t}$ when
$a=A_{t-1}$. Letting $\alpha\left(\cdot|s\right)$ be independent
of $s\in\mathcal{S}$ and uniform over the action space $\mathcal{A}$
is the least informative choice. The most informative choice is to
choose $\alpha\left(\cdot|s\right)$ equal to the point mass at $\pi^{*}\left(s\right)$
where $\pi^{*}$ is the optimal policy. However, this is not known,
and in fact it is ruled out by our assumptions above. In between these
two extreme choices, we could use $\alpha\left(\cdot|s\right)$ to
be a convex combination of the uniform distribution on $\mathcal{A}$
and the point mass on some guess prior policy. Our methodology is
consistent in this case as well, as long as the prior is independent
of the estimation sample. We can view $\alpha$ as a prior conditional
probability. To summarize, the sole purpose of the state action pair
$\left(S_{t},A_{t}\right)_{t\geq1}$ is to ensure that we have an
initial set of data such that all states and actions are visited eventually
as the sample goes to infinity. This is necessary for proper exploration
of the state action space. However, given that $\left(\tilde{S}_{t}\right)_{t\geq1}$
is independent of the actions, we shall only need a single sample
realization to achieve consistency.

\subsection{Norms and Expectations\protect\label{Section_normsExpectations}}

Denote by $P$ the law of $S_{t}$, where $S_{t}=\left(\tilde{S}_{t},A_{t-1}\right)$
as in Section \ref{Section_stateActionPair}. The empirical law is
denoted by $P_{n}=\frac{1}{n}\sum_{t=1}^{n}\delta_{S_{t}}$ where
$\delta_{S_{t}}$ is the point mass at $S_{t}$. This means that for
$f:\mathcal{S}\rightarrow\mathbb{R}$, $Pf=\int_{\mathcal{S}}f\left(s\right)dP\left(s\right)$
and $P_{n}f=\frac{1}{n}\sum_{t=1}^{n}f\left(S_{t}\right)$. In general,
for any measure $\mathbb{Q}$ and function $f$ on some set $\mathcal{X}$,
$\mathbb{Q}f=\int_{\mathcal{X}}f\left(x\right)d\mathbb{Q}\left(x\right)$
and $\left|\cdot\right|_{p,\mathbb{Q}}$ is the $L_{p}$ norm w.r.t
$\mathbb{Q}$, i.e. $\left|f\right|_{p,\mathbb{Q}}=\left(\mathbb{Q}\left|f\right|^{p}\right)^{1/p}$.
Moreover, $\left|f\right|_{\infty}=\sup_{x\in\mathcal{X}}\left|f\left(x\right)\right|$
is the uniform norm. We denote by $T$ the shift operator such that
$TS_{t}=S_{t+1}$. To avoid notational trivialities, throughout, we
suppose that we have a sample of $n+1$ states, which means that we
have a sample of $n$ observations $S_{t+1}=TS_{t}$, $t=1,2,...,n$.
We denote by $T^{a}$ the shift operator that evaluates the last entry
in $S_{t}$ to $a$. This means that $T^{a}S_{t}=S_{t+1}^{a}$, i.e.
we replace $A_{t-1}$ with $a$ in $S_{t}$. Finally, $T^{\pi}$ is
the shift operator s.t. $T^{\pi}S_{t}=S_{t+1}^{a}$ with $a=\pi\left(S_{t}\right)$.
Note that this is not the same as $S_{t+1}^{\pi}$ because $S_{t+1}^{\pi}$
is $S_{t+1}^{a}$ with $a=\pi\left(S_{t}^{\pi}\right)$. The sequence
$\left(S_{t}^{\pi}\right)_{t\geq1}$ is what appears in (\ref{EQ_valueFunction})
and is such that each state $S_{t}^{\pi}$ depends on the previous
one according to the policy $\pi$. Hence, this latter sequence can
only be defined conditioning on some initial state, e.g. $S_{1}^{\pi}=s\in\mathcal{S}$.
Also note that for any $s\in\mathcal{S}$, $T^{a}s=s^{a}$ and $T^{\pi}s=s^{\pi\left(s\right)}$,
i.e. a variable that does not depend on $t\geq1$ is returned as it
is, but with its last entry modified accordingly.

To avoid notational complexities, we overload the meaning of the above
probability measures when applied to the reward $R$. In particular,
we write $P_{n}R\left(T^{a}\cdot;\theta^{a}\left(\cdot\right)\right)=\frac{1}{n}\sum_{t=1}^{n}R\left(S_{t+1}^{a};\theta^{a}\left(S_{t}\right)\right)$,
and $PR\left(T^{a}\cdot;\theta^{a}\left(\cdot\right)\right)=\int_{\mathcal{S}}R\left(s;\theta^{a}\left(r\right)\right)\kappa^{a}\left(r,ds\right)dP\left(r\right)$.
The overloaded notation means that $PR\left(T^{a}\cdot;\theta^{a}\left(\cdot\right)\right)=P\mathbb{K}^{a}R$.

\subsection{Operators and Classes of Functions for the Estimation Problem\protect\label{Section_operatorsFunctionClasses}}

This section is necessarily technical. The reader can skim through
it and refer to it when reading the description of Algorithm \ref{Algorithm_Sieve},
in Section \ref{Section_algorithm}. The main objects of interest
to understand Algorithm \ref{Algorithm_Sieve} are two operators,
$\Pi_{n,B}$, and $\Gamma_{n}^{a}$, which we shall define momentarily.

Let $\left\{ \varphi_{k}:\mathcal{S}\rightarrow\mathbb{R},k=1,2,...,K\right\} $
be a set of functions. Define 
\begin{equation}
\mathcal{F}:=\left\{ f:f\left(s\right)=\sum_{k=1}^{K}b_{k}\varphi_{k}\left(s\right),s\in\mathcal{S}\right\} \label{EQ_spaceFunctions}
\end{equation}
to be the space of approximating functions. For any $f:\mathcal{S}\rightarrow\mathbb{R}$,
$\Pi_{n}$ is the operator such that 
\begin{equation}
\Pi_{n}f=\arg\inf_{h\in\mathcal{F}}\left\{ \left|f-h\right|_{2,P_{n}}^{2}+{\rm Pen}\left(h\right)\right\} \label{EQ_projPenalised}
\end{equation}
where ${\rm Pen}\left(\cdot\right):\mathcal{F}\rightarrow\left[0,\infty\right)$
is a penalty. In particular, we only consider the $\ell_{2}$ penalty
on the linear coefficients $b_{k}$, i.e. for $h\in\mathcal{F}$ such
that $h\left(\cdot\right)=\sum_{k=1}^{K}b_{k}\varphi_{k}\left(\cdot\right)$,
${\rm Pen}\left(h\right)=\rho\sum_{k=1}^{K}b_{k}^{2}$ for some $\rho\geq0$.
For $\rho=0$, we use the Moore\textendash Penrose generalised inverse
of the empirical Gram matrix to ensure $\Pi_{n}$ remains linear.
We define the cap operator on $\mathbb{R}$ such that for $x\in\mathbb{R}$,
and a cap value $B>0$, $\left[x\right]_{B}={\rm sign}\left(x\right)\min\left\{ \left|x\right|,B\right\} $.
Hence, we define the capped class of functions 
\begin{equation}
\mathcal{F}_{B}:=\left\{ h=\left[f\right]_{B}:f\in\mathcal{F}\right\} .\label{EQ_cappedSpaceFunctions}
\end{equation}
Note that for a function $h$ on $\mathcal{S}$ let $\left|h\right|_{\infty}=\sup_{s\in\mathcal{S}}\left|h\right|$.
Then, for any probability measure $\mathbb{Q}$, including $P_{n}$,
\begin{equation}
\left[f\right]_{B}=\arg\inf_{h:\left|h\right|_{\infty}\leq B}\left|f-h\right|_{2,\mathbb{Q}}^{2}.\label{EQ_projBoxConstraint}
\end{equation}
This follows from the fact that minimisation in $\left\{ h:\left|h\right|_{\infty}\leq B\right\}$ can be performed for each individual value of $h\left(s\right)$, $s\in\mathcal{S}$, $\mathbb{Q}$-almost surely. We define the operator $\Pi_{n,B}$ on $\mathcal{F}$ s.t. $\Pi_{n,B}f=\left[\Pi_{n}f\right]_{B}$. For the interested reader, Lemmas \ref{Lemma_projBasicIneq} and \ref{Lemma_projDiffIneq} in the Electronic Supplement state a few properties of the operator $\Pi_{n,B}$.

For any $f:\mathcal{S}\rightarrow\mathbb{R}$, let $\Gamma_{n}^{a}$
be the operator such that $\Gamma_{n}^{a}f=R\left(T^{a}\cdot;\theta^{a}\left(\cdot\right)\right)+\gamma f\left(T^{a}\cdot\right)$
for $f$ evaluated at $\left\{ S_{t}:t=1,2,...,n\right\} $, and zero
otherwise. This is because we are supposing a sample of $n+1$ observations.
With these definitions, for $f:\mathcal{S}\rightarrow\mathbb{R}$,
\[
\Pi_{n}\Gamma_{n}^{a}f=\arg\min_{h\in\mathcal{F}}\left\{ \frac{1}{n}\sum_{t=1}^{n}\left|R\left(S_{t+1}^{a};\theta^{a}\left(S_{t}\right)\right)+\gamma f\left(S_{t+1}^{a}\right)-h\left(S_{t}\right)\right|^{2}+{\rm Pen}\left(h\right)\right\} .
\]
Hence, $\Pi_{n,B}\Gamma_{n}^{a}f\left(\cdot\right)=\left[\sum_{k=1}^{K}b_{k}\varphi_{k}\left(\cdot\right)\right]_{B}$
where the coefficients $b_{k}$ are the regression coefficients obtained
regressing $R\left(S_{t+1}^{a};\theta^{a}\left(S_{t}\right)\right)+\gamma f\left(S_{t+1}^{a}\right)$
on $\varphi_{1}\left(S_{t}\right),\varphi_{2}\left(S_{t}\right),...,\varphi_{K}\left(S_{t}\right)$
for $t=1,...,n$.

Additionally, define the following population counterpart of the operator
$\Gamma_{n}^{a}$. For any $f:\mathcal{S}\rightarrow\mathbb{R}$ and
$a\in\mathcal{A}$, $\mathbb{T}^{a}$ is such that $\mathbb{T}^{a}f\left(s\right)=\mathbb{K}^{a}R\left(s\right)+\gamma\mathbb{K}^{a}f\left(s\right)$
and $\mathbb{T}f\left(s\right)=\max_{a\in\mathcal{A}}\mathbb{T}^{a}f\left(s\right)$.
This notation allows us to write the r.h.s. of (\ref{EQ_BellmanEq}) and (\ref{EQ_QRecursion}) more compactly as $V^{\pi}\left(s\right)=\max_{a\in\mathcal{A}}\mathbb{T}^{a}V^{\pi}\left(s\right)$ and $Q^{\pi}\left(s,a\right)=\mathbb{T}^{a}V^{\pi}\left(s\right)$, respectively. Moreover, for any fixed $f:\mathcal{S}\rightarrow\mathbb{R}$,
we have that 
\[
\left(\Gamma_{n}^{a}-\mathbb{T}^{a}\right)f\left(S_{t}\right)=R\left(S_{t+1}^{a};\theta^{a}\left(S_{t}\right)\right)-\mathbb{K}^{a}R\left(S_{t}\right)+\gamma\left(f\left(T^{a}S_{t}\right)-\mathbb{K}^{a}f\left(S_{t}\right)\right)
\]
 is mean zero conditioning on $S_{t}$.

In order to describe the approximation error of Algorithm \ref{Algorithm_Sieve},
we need to introduce some additional classes of functions with some
basic motivation. For each $a\in\mathcal{A}$, $\Pi_{n,B}\Gamma_{n}^{a}f$
is an element of $\mathcal{F}_{B}$ in (\ref{EQ_cappedSpaceFunctions}).
We shall extend functions on $\mathcal{S}$ to functions on $\mathcal{S}\times\mathcal{A}$
using indicator functions. This can be used to approximate the action
value function $\mathcal{S}\times\mathcal{A}\ni\left(s,a\right)\mapsto Q\left(s,a\right)$.
Hence, we define the class of functions 
\begin{equation}
\mathcal{F}_{B}^{\mathcal{A}}:=\left\{ h\left(s,a\right)=\sum_{a'\in\mathcal{A}}f_{a'}\left(s\right)1_{\left\{ a'=a\right\} }:s\in\mathcal{S},a\in\mathcal{A},f_{a'}\in\mathcal{F}_{B}\right\} .\label{EQ_classFunctionsFBA}
\end{equation}
Our results will be stated in terms of the value function. Hence,
to approximate the value function, we need to take the maximum w.r.t.
$a\in\mathcal{A}$, i.e. $\max_{a\in\mathcal{A}}\Pi_{n,B}\Gamma_{n}^{a}f$.
This means that we shall be interested in the following class of functions
\begin{equation}
\mathcal{F}_{B}^{\max}:=\left\{ h\left(s\right)=\max_{a\in\mathcal{A}}\sum_{a'\in\mathcal{A}}f_{a'}\left(s\right)1_{\left\{ a'=a\right\} }:s\in\mathcal{S},f_{a'}\in\mathcal{F}_{B}\right\} .\label{EQ_classFunctionsFBMax}
\end{equation}

\subsection{Description of the Algorithm using Operators\protect\label{Section_algoDescription}}

The algorithm to choose among competing models approximates (\ref{EQ_QRecursion}), using sample quantities. We use an iterative algorithm. At iteration $j$, the policy is $\pi_{j}$, and we write $V^{\left(j\right)}=V^{\pi_{j}}$ for convenience to denote the true value function when using policy $\pi_{j}$. Note that $\pi_{j}$ is random because it is an estimator generated by the algorithm. At each iteration, we start with an estimated value function $\hat{V}^{\left(j-1\right)}=\hat{V}^{\pi_{j-1}}$ which depends on a policy $\pi_{j-1}$. For a given action $a$ that chooses a model, we compute the performance in terms of the rewards $R\left(S_{t+1}^{a};\theta^{a}\left(S_{t}\right)\right)$,
$t=1,2,...,n$. We regress $R\left(S_{t+1}^{a};\theta^{a}\left(S_{t}\right)\right)+\gamma\hat{V}^{\left(j-1\right)}\left(S_{t+1}^{a}\right)$
on the functions $\varphi_{1}\left(S_{t}\right),\varphi_{2}\left(S_{t}\right),...,\varphi_{K}\left(S_{t}\right)$,
$t=1,2,...,n$, using a ridge penalty. Once capped by $B$, the regression
function produces the approximation to the Q function, $\hat{Q}^{\left(j\right)}\left(\cdot,a\right):=\Pi_{n,B}\Gamma_{n}^{a}\hat{V}^{\left(j-1\right)}$,
the optimal policy, $\pi_{j}=\arg\max_{a\in\mathcal{A}}\hat{Q}^{\left(j\right)}\left(\cdot,a\right)$,
and the value function, $\hat{V}^{\left(j\right)}\left(\cdot\right)=\max_{a\in\mathcal{A}}\hat{Q}^{\left(j\right)}\left(\cdot,a\right)$.
Note that $\hat{V}^{\left(j\right)}\left(s\right)=\max_{a\in\mathcal{A}}\hat{Q}^{\left(j\right)}\left(s,a\right)=\hat{Q}^{\left(j\right)}\left(s,\pi_{j}\left(s\right)\right)$.
The recursion starts with $\hat{V}^{\left(0\right)}:=0$. Algorithm
\ref{Algorithm_Sieve} summarizes the procedure. For the sake of clarity,
in Algorithm \ref{Algorithm_Sieve}, $\Gamma_{n}^{a}\hat{V}^{\left(j-1\right)}\left(S_{t}\right)=R\left(S_{t+1}^{a};\theta^{a}\left(S_{t}\right)\right)+\gamma\hat{Q}^{\left(j-1\right)}\left(S_{t+1}^{a},\pi_{j-1}\left(S_{t+1}^{a}\right)\right)$;
note that $\hat{Q}^{\left(j-1\right)}\left(S_{t+1}^{a},\pi_{j-1}\left(S_{t+1}^{a}\right)\right)=\max_{a'\in\mathcal{A}}\hat{Q}^{\left(j-1\right)}\left(S_{t+1}^{a},a'\right)$,
which is $\hat{V}^{\left(j-1\right)}\left(S_{t+1}^{a}\right)$.

\begin{algorithm}
\caption{Q-Function Estimation for Model Selection.}
\label{Algorithm_Sieve}

Start with $\hat{V}^{\left(0\right)}=0$.

Set $\epsilon>0$, ${\rm cond}={\rm True}$ and $j=1$.

A sample time series of states $\left\{ S_{t}:t=1,2,...,n+1\right\} $
is given.

While ${\rm cond}$ is True:

For $a\in\mathcal{A}$:

Compute $\left\{ R_{t+1}^{a}=R\left(S_{t+1}^{a};\theta^{a}\left(S_{t}\right)\right):t=1,...,n\right\} ;$

Find $\hat{Q}^{\left(j\right)}\left(\cdot,a\right)=\Pi_{n,B}\Gamma_{n}^{a}\hat{V}^{\left(j-1\right)}\left(\cdot\right)$

End of For.

Set $\pi_{j}\left(\cdot\right)=\arg\max_{a\in\mathcal{A}}\hat{Q}^{\left(j\right)}\left(\cdot,a\right)$
and $\hat{V}^{\left(j\right)}\left(s\right)=\hat{Q}^{\left(j\right)}\left(s,\pi_{j}\left(s\right)\right)$
for $s\in\mathcal{S}$.

$j=j+1$

${\rm cond}=\left|\hat{V}^{\left(j\right)}-\hat{V}^{\left(j-1\right)}\right|_{2,P_{n}}>\epsilon$.

End of While.
\end{algorithm}

\paragraph{Averaging Across Simulated Actions.}

Algorithm \ref{Algorithm_Sieve} uses a single action state sample
$\left(S_{t},A_{t}\right)_{t\geq1}$. Recall that $S_{t}$ includes
$A_{t-1}$ as part of its values and the sequence of actions is randomly
generated (see Section \ref{Section_stateActionPair}). This can make
results dependent on the single realization of $\left(A_{t}\right)_{t\geq0}$.
However, the actions $\left(A_{t}\right)_{t\geq0}$ in the definition
of the augmented state are computed independently of the sample data.
Hence, we can generate $N_{A}$ augmented states $S_{t}^{\left(u\right)}=\left(\tilde{S}_{t},A_{t-1}^{\left(u\right)}\right)$,
$u=1,2,...,N_{A}$.\textbf{ }We can then apply Algorithm \ref{Algorithm_Sieve}
to each of these augmented states in parallel. The final estimator
of $\hat{Q}_{N_{A}}^{\left(J\right)}$ is the average across the estimators
$\hat{Q}^{\left(J,u\right)}$, $u=1,2,...,N_{A}$, i.e. 
\begin{equation}
\hat{Q}_{N_{A}}^{\left(J\right)}\left(\cdot,a\right)=\frac{1}{N_{A}}\sum_{u=1}^{N_{A}}\hat{Q}^{\left(J,u\right)}\left(\cdot,a\right).\label{EQ_averageQ}
\end{equation}
The final estimator for the policy is given by $\pi_{J,N_{A}}\left(\cdot\right):=\arg\max_{a\in\mathcal{A}}\hat{Q}_{N_{A}}^{\left(J\right)}\left(\cdot,a\right)$
and the value function is $\hat{V}_{N_{A}}^{\left(J\right)}$. Simulations
carried out by the authors showed better behaved results when we use
(\ref{EQ_averageQ}) with $N_{A}>1$. However, given the recurrent
nature of the methodology, we were unable to derive a theoretical
result that supports this claim.

\section{Analysis of the Algorithm\protect\label{Section_asymptotics}}

We shall state the assumptions under which the algorithm will be analysed. We then include remarks on such assumptions. Finally, we state a number of consistency results based on such assumptions. We use l.h.s. and r.h.s. to mean left and right hand side, respectively. The symbol $\lesssim$ means that the l.h.s. is less than a constant times the r.h.s. and $\asymp$ means that the l.h.s. is bounded by positive constants times the r.h.s. 

\subsection{Assumptions\protect\label{Section_assumptions}}

In what follows, Assumptions \ref{Condition_stateActionPair-1} and
\ref{Condition_moment-1} weaken the corresponding ones in the main
text. We provide the full set for ease of exposition. 

\begin{condition}\label{Condition_stateActionPair-1}(State Action
Pair) The state action pair $\left(S_{t},A_{t}\right)_{t\geq1}$ and
$A_{0}$ satisfy (\ref{EQ_conditionStateActionPair}) where $\left|\mathcal{A}\right|$
is a bounded integer, $S_{t}=\left(\tilde{S}_{t},A_{t-1}\right)$
where $\tilde{S}_{t}$ takes values in some set $\tilde{\mathcal{S}}$,
and $\left(\tilde{S}_{t}\right)_{t\geq1}$ evolves according to a
Markov transition law that does not depend on $\left(A_{t}\right)_{t\geq0}$.
\end{condition}

\begin{condition}\label{Condition_mixing-1}(Markov Assumption) The
sequence of states $S:=\left(S_{t}\right)_{\geq1}$ is a strictly
stationary Markov process with beta mixing coefficients $\left(\beta_{q}\right)_{q\geq1}$
satisfying $\beta_{q}\lesssim e^{-C_{\beta}q}$ for some constant
$C_{\beta}>0$.\end{condition}

\begin{condition}\label{Condition_moment-1}(Moments) We have that
$\max_{a\in\mathcal{A}}\left|R\left(T^{a}\cdot;\theta^{a}\left(\cdot\right)\right)\right|_{v,P}<\infty$
for $v\geq2$.\end{condition}

\begin{condition}\label{Condition_theta-1}(Rewards Parameter) The
map $\mathcal{S}\ni s\mapsto\theta^{a}\left(s\right)$ is possibly
random, but independent of the sequence of states $\left(S_{t}\right)_{t\geq1}$
and measurable at time $t=0$.\end{condition}

\begin{condition}\label{Condition_functionClass-1}(Approximating
Function Class) The functions $\varphi=\left\{ \varphi_{l}:l=1,2,...,K\right\} $
in (\ref{EQ_spaceFunctions}) are uniformly bounded and the population
Gram matrix $P\varphi\varphi^{\mathrm{T}}$ is positive definite,
where $\varphi$ is arranged as a column vector. Moreover, $K\geq2$,
$\sup_{s\in\mathcal{S}}\varphi\left(s\right)^{\mathrm{T}}G^{-1}\varphi\left(s\right)\lesssim K$
such that $r_{n}=\sqrt{n^{-1}K\left(\ln K\right)\left(\ln n\right)}\rightarrow0$.
\end{condition}

Assumption \ref{Condition_stateActionPair-1} is more general than
Assumption \ref{Condition_stateActionPair} as discussed in Section
\ref{Section_stateActionPair}. 

The only difference from Assumption \ref{Condition_moment} and Assumption
\ref{Condition_moment-1} is that we may only require a second moment,
depending on the result. Note that $\max_{a\in\mathcal{A}}\left|R\left(T^{a}\cdot;\theta^{a}\left(\cdot\right)\right)\right|_{v,P}<\infty$
is the same as $\max_{a\in\mathcal{A}}\left(\mathbb{E}\left|R\left(S_{t+1}^{a};\theta^{a}\left(S_{t}\right)\right)\right|^{v}\right)^{1/v}<\infty$
for some $v\geq2$. For example, random variables following a GARCH
model with conditionally Gaussian errors have unconditional distribution
with Pareto tails (Basrak et al., 2002). A weak moment assumption
is necessary for some problems. By Jensen inequality, $\left|\mathbb{K}^{a}R\right|_{v,P}\leq\left|R\right|_{v,P}$.
Hence, Assumption \ref{Condition_moment} implies the conditional
moment condition $\max_{a\in\mathcal{A}}\left|\mathbb{K}^{a}R\right|_{v,P}<\infty$
for $v\geq2$. 

\subsection{Convergence Analysis of Algorithm \ref{Algorithm_Sieve}\protect\label{Section_convergenceResults}}

Recall the definitions of $V^{\left(J\right)}:=V^{\pi_{J}}$ and $V^{*}$  (see (\ref{EQ_valueFunction}) and (\ref{EQ_optimalValueFunction}) in the main text, or  (\ref{EQ_valueFunction}) and (\ref{EQ_optimalValueFunction} in this supplement)). Moreover, $\pi_{J}$ is the policy estimated from Algorithm \ref{Algorithm_Sieve} after $J$ iterations. If the functions $\left\{ \theta^{a}:a\in\mathcal{A}\right\}$ are random but satisfy Assumption \ref{Condition_theta-1}, the results that follow are understood to be conditional on $\left\{ \theta^{a}:a\in\mathcal{A}\right\} $.

\begin{theorem}\label{Theorem_convergenceAlgo} Suppose Assumptions
hold with $v>2$. Then for every $J\ge1$, 

\[
\left|V^{*}-V^{\left(J\right)}\right|_{2,P}\leq\frac{2}{\underline{\alpha}\left(1-\gamma\right)^{2}}\sqrt{\left|\mathcal{A}\right|}\left(Aprox+Cap+Est\right)+\frac{2\gamma^{J}}{\underline{\alpha}\left(1-\gamma\right)}\left|V^{*}\right|_{2,P},
\]
where $Approx$, $Cap$ and $Est$ are approximation error, error
incurred from capping, and estimation error, respectively and in particular,
\begin{align*}
Approx= & \max_{a\in\mathcal{A}}\sup_{f\in\mathcal{F}_{B}^{\max}}\inf_{h\in\mathcal{F}}\left|\mathbb{T}^{a}f-h\right|_{2,P}\\
Cap= & O\left(B^{-(v-2)/2}\right)\\
Est= & O_{P}\left(\left(1+\gamma B\right)\left(r_{n}+\rho\left|G^{-1}\right|_{{\rm op}}\right)\right)
\end{align*}
When $v=2$, we can replace $O\left(B^{-(v-2)/2}\right)$ by $\max_{a\in\mathcal{A}}\left|\left(\left|\mathbb{K}^{a}R\right|-\left(1-\gamma\right)B\right)_{+}\right|_{2,P}$,
which is $o(1)$ as $B\to\infty$. When $|\mathbb{T}^{a}f|_{\infty}\le B$
uniformly over $a\in\mathcal{A}$ and $f\in\mathcal{F}_{B}^{\max}$,
the capping error $Cap$ can be set to zero. \end{theorem}

The result does not require the optimal value function $V^{*}$ to
belong to $\mathcal{F}$. The approximation condition concerns the
functions $\mathbb{T}^{a}f=\mathbb{K}^{a}R+\gamma\mathbb{K}^{a}f$
for $a\in\mathcal{A}$ and $f\in\mathcal{F}_{B}^{\max}$. The term
$V^{*}$ on the r.h.s. enters the bound through the initialisation
error.

By judicious choice of $B$, we derive a simpler bound. For simplicity,
it does not make explicit the dependence on $\left|\mathcal{A}\right|$.

\begin{corollary}\label{Corollary_withMinimumDistance}Suppose that
the Assumptions hold with $v>2$. Set $B\asymp r_{n}^{-2/v}$ where
$r_{n}$ is as in Assumption \ref{Condition_functionClass-1}. Then
\begin{align}
\left|V^{*}-V^{\left(J\right)}\right|_{2,P} & \lesssim\max_{a\in\mathcal{A}}\sup_{f\in\mathcal{F}_{B}^{\max}}\inf_{h\in\mathcal{F}}\left|\mathbb{T}^{a}f-h\right|_{2,P}\nonumber \\
 & +O_{P}\left(r_{n}^{(v-2)/v}+\rho\left|G^{-1}\right|_{{\rm op}}r_{n}^{-2/v}+\gamma^{J}\left|V^{*}\right|_{2,P}\right).\label{EQ_corollaryWithMinDistanceBound}
\end{align}
\end{corollary}

Under conditions that include uniform boundedness of the conditional
rewards, we do not incur an approximation error for large enough capping. 

\begin{corollary}\label{Corollary_withZeroDistance} Under the conditions
of Theorem \ref{Theorem_convergenceAlgo} (including its $v=2$ extension),
suppose that for a fixed $B>0$, uniformly in $a\in\mathcal{A}$ and
$f\in\mathcal{F}_{B}^{\max}$, we have that $\mathbb{T}^{a}f\in\mathcal{F}$
and $\left|\mathbb{T}^{a}f\right|_{\infty}\leq B$. Then
\[
\left|V^{*}-V^{\left(J\right)}\right|_{2,P}=O_{P}\left(\left(1+B\right)\left[r_{n}+\rho\left|G^{-1}\right|_{{\rm op}}+\gamma^{J}\left|V^{*}\right|_{2,P}\right]\right).
\]
 \end{corollary}

A sufficient condition for $\left|\mathbb{T}^{a}f\right|_{\infty}\leq B$
is that $B\geq\max_{a\in\mathcal{A}}\left|\mathbb{K}^{a}R\right|_{\infty}/\left(1-\gamma\right)$.
Then, $\left|\mathbb{T}^{a}f\right|_{\infty}\leq\left(1-\gamma\right)B+\gamma B=B$ for any $f\in\mathcal{F}_{B}^{\max}$. The condition $\mathbb{T}^{a}f\in\mathcal{F}$ also means that the conditional expectation of the rewards and the conditional expectation of the functions in $\mathcal{F}_{B}^{\max}$ are in $\mathcal{F}$: $\mathbb{K}^{a}R,\mathbb{K}^{a}f\in\mathcal{F}$.
The corollary does not require $K$ to be fixed. When $K$ is fixed, assuming the population Gram matrix is full rank $G$ we can choose $\rho=0$.

\paragraph{Sieve Approximation.}

Note that $\mathcal{F}$ in (\ref{EQ_spaceFunctions}) depends on
$K$. If $\mathbb{T}^{a}f$ can only be approximated by $\mathcal{F}$
as $K\to\infty$, the speed of convergence depends on the approximation
rate. We give an example below. To simplify notation, suppose that
$\mathcal{S}=[0,1]\times\mathcal{A}$, so that there is only one raw-state
variable in addition to the previous action. 

We assume that the functions $\varphi_{l}$ in $\mathcal{F}$ are
constructed from a B-spline basis of order $q$. For a fixed integer
$q\ge1$, let $e_{1},\ldots,e_{L+q-1}$ be the standard B-spline basis
of order $q$, with $L+1$ equally spaced knots in $\left[0,1\right]$.
These are piecewise polynomials of degree $q-1$. For $q=1$, these
are piecewise constant functions. Consider the basis 
\begin{equation}
\varphi_{l,a'}\left(\left(\tilde{s},a\right)\right)=e_{l}\left(\tilde{s}\right)1_{\left\{ a=a'\right\} }\label{EQ_polyFuncClass}
\end{equation}
for $l=1,2,...,L+q-1$ and $a'\in\mathcal{A}$ so that $K=(L+q-1)|\mathcal{A}|$. 

\begin{corollary}\label{Corollary_withPolynomialApprox} Suppose
Assumptions \ref{Condition_stateActionPair-1}\textendash{} \ref{Condition_theta-1}
hold with $v>2$. Assume that the law of the raw state has a density
bounded away from zero. Suppose that for some fixed finite integer
$q\geq1$, the derivatives of $\mathbb{K}^{a}R\left(\left(\tilde{s},a'\right)\right)$
and $\mathbb{K}^{a}g\left(\left(\tilde{s},a'\right)\right)$ w.r.t.
$\tilde{s}$ up to order $q$ are uniformly bounded in $\tilde{s}\in\left[0,1\right]$
and $g:\mathcal{S}\rightarrow\left[-1,1\right]$. Fix $B$ such that
$B\ge\max_{a\in\mathcal{A}}|\mathbb{K}^{a}R|_{\infty}/\left(1-\gamma\right)$.
Use the basis functions in (\ref{EQ_polyFuncClass}) with $L\asymp\left(n/\left(\ln n\right)^{2}\right)^{1/\left(2q+1\right)}$
and set the ridge parameter to $\rho=O\left(\left[\left(\ln n\right)^{2}/n\right]^{\left(q+1\right)/\left(2q+1\right)}\right)$.
Then 
\[
\left|V^{*}-V^{\left(J\right)}\right|_{2,P}=O_{P}\left(\left(\frac{\left(\ln n\right)^{2}}{n}\right)^{q/\left(2q+1\right)}+\gamma^{J}\left|V^{*}\right|_{2,P}\right).
\]
\end{corollary}

\paragraph{Rewards Depending on Estimated Sample Quantities.}

We can allow the functions $\left\{ \theta^{a}:a\in\mathcal{A}\right\} $
to be estimated on the same sample used for estimation of the optimal
policy in Algorithm \ref{Algorithm_Sieve}. We assume that the sample
is used to construct a preliminary estimator $\hat{\theta}^{a}$ in
a set $\Theta$, $a\in\mathcal{A}$. This could be a function of a
model that is estimated using the same sample data, as in Example
\ref{Example_linearModelThetaEstimated} (Section \ref{Section_remarksConditions}).
We consider the case where the estimator converges to a nonstochastic
element, e.g. a pseudo true value. This is weaker than requiring consistency.
To control the error caused by estimating $\left\{ \theta^{a}:a\in\mathcal{A}\right\} $
we assume the following commonly used Lipschitz condition (e.g. Andrews,
1994, eq.4.3), 
\begin{equation}
\left|R\left(s;\theta_{1}^{a}\left(r\right)\right)-R\left(s;\theta_{2}^{a}\left(r\right)\right)\right|\leq\Delta\left(s\right)\left|\theta_{1}^{a}-\theta_{2}^{a}\right|_{\infty}\label{EQ_liptchitzReward}
\end{equation}
$s,r\in\mathcal{S}$. Let $\theta^{\pi}\left(s\right)=\theta^{a}\left(s\right)$
with $a=\pi\left(s\right)$, $s\in\mathcal{S}$. To make the dependence
on $\theta^{\pi}$ explicit, we write $\mathbb{K}^{\pi}R\left(r;\theta^{\pi}\right)=\int_{\mathcal{S}}R\left(s;\theta^{\pi}\left(r\right)\right)\kappa^{\pi}\left(r,ds\right)$
and similarly for $\mathbb{K}^{a}R\left(r;\theta^{a}\right)$, $a\in\mathcal{A}$.

\begin{condition}\label{Condition_thetaEstimated}The following hold
true:

1. The Lipschitz condition in (\ref{EQ_liptchitzReward}) holds and
satisfies $\max_{a}\left|\Delta\left(T^{a}\cdot\right)\right|_{v,P}<\infty$
for the same $v\geq2$ as in Assumption (\ref{Condition_moment-1}).

2. There is a nonstochastic set of functions $\left\{ \theta^{a}:a\in\mathcal{A}\right\} $
such that $\max_{a\in\mathcal{A}}\left|\hat{\theta}^{a}-\theta^{a}\right|_{\infty}=O_{P}\left(\sqrt{\frac{C_{\Theta}}{n}}\right)$,
for some fixed $C_{\Theta}<\infty$, where $\left\{ \hat{\theta}^{a}:a\in\mathcal{A}\right\} $
is the estimated set of functions.

\end{condition}

We need to make explicit the dependence of the rewards on $\left\{ \theta^{a}:a\in\mathcal{A}\right\} $.
Hence, we write $V^{\pi}\left(\cdot\right)=V^{\pi}\left(\cdot;\theta^{\pi}\right)=\sum_{t=1}^{\infty}\gamma^{t-1}\left(\mathbb{K}^{\pi}\right)^{t}R\left(\cdot;\theta^{\pi}\right)$,
using (\ref{EQ_valueFunction}). As usual, $V^{*}\left(\cdot\right):=\max_{\pi}V^{\pi}\left(\cdot\right)$
where the maximum is over all policies on $\mathcal{S}$. When we
use rewards dependent on a sample parameter, we replace $\theta^{\pi}$
with $\hat{\theta}^{\pi}$ in the above and write $V^{\pi}\left(\cdot;\hat{\theta}^{\pi}\right)$.
After $J$ iterations, Algorithm \ref{Algorithm_Sieve} produces a
policy $\pi_{J}$ for rewards based on $R\left(T^{a}\cdot;\hat{\theta}^{a}\left(\cdot\right)\right)$,
$a\in\mathcal{A}$. The next result shows that $\pi_{J}$ is optimal
in the sense that the value function $V^{\pi_{J}}\left(\cdot\right)=V^{\pi_{J}}\left(\cdot;\theta^{\pi_{J}}\right)$
is close to $V^{*}$ in $L_{2}$ norm. Unlike Theorem \ref{Theorem_convergenceAlgo},
we explicitly write $V^{\pi_{J}}$ instead of $V^{\left(J\right)}$
to ensure that it is clear that $\pi_{J}$ here is not the same as
in Theorem \ref{Theorem_convergenceAlgo}, as the rewards are dependent
on the estimators $\left\{ \hat{\theta}^{a}:a\in\mathcal{A}\right\} $.

\begin{corollary}\label{Corollary_withEstimatedTheta} Suppose the
assumptions of Theorem \ref{Theorem_convergenceAlgo} hold, with Assumption
\ref{Condition_theta-1} replaced by Assumption \ref{Condition_thetaEstimated}.
Assume that the limiting reward parameters $\theta^{a}$ are nonstochastic
and satisfy the moment condition in Assumption \ref{Condition_moment-1}.
Let $\pi_{J}$ be the policy obtained by using $R(T^{a}\cdot;\widehat{\theta}^{a}(\cdot))$
in Algorithm \ref{Algorithm_Sieve}, and let $V^{\pi_{J}}$ be evaluated
under the limiting rewards $R(T^{a}\cdot;\theta^{a}(\cdot))$. Then
the conclusion of Theorem \ref{Theorem_convergenceAlgo} continues
to hold with an additional $O_{P}\left(\sqrt{C_{\Theta}/n}\right)$
term in the estimation error $Est$. \end{corollary}

\section{Proofs\protect\label{Section_proofs}}

The proofs of the main result use a standard coupling for beta mixing random variables. This allows us to adapt existing inequalities for i.i.d. observations to the dependent case (Section \ref{Section_inequalitiesDependentRV}). The inequalities are based on the covering number of some function classes within which the various estimated quantities lie (Section \ref{Section_CoveringNumbers}).

When stating results, we shall control the complexity of function classes using covering numbers. Recall the notation from Section \ref{Section_normsExpectations}.
For a norm $\left|\cdot\right|_{p,\mathbb{Q}}$, the $\epsilon$-covering
number of a class of functions $\mathcal{G}$ is the minimum number
of balls of radius $\epsilon$, under $\left|\cdot\right|_{p,\mathbb{Q}}$,
required to cover $\mathcal{G}$. We denote this covering number by
$N\left(\epsilon,\mathcal{G},\left|\cdot\right|_{p,\mathbb{Q}}\right)$.
An envelope function for $\mathcal{G}$ is a function $F$ such that
$\left|f\right|\leq F$ for all $f\in\mathcal{G}$. The minimal envelope
function is $F:=\sup_{f\in\mathcal{G}}\left|f\right|$. We shall assume
that every quantity is measurable in what follows given that our sets
are finite dimensional and often compact. Finally, we use $\left|\cdot\right|_{2}$
for the Euclidean norm, and $\lambda_{\max}\left(\cdot\right)$, $\lambda_{\min}\left(\cdot\right)$
for the maximum and minimum eigenvalue of the matrix argument. For
positive definite matrices, the maximum eigenvalue is the same as
the operator norm which we denote by $\left|\cdot\right|_{{\rm op}}$. 

\subsection{A Maximal Inequality for Dependent Random Variables\protect\label{Section_inequalitiesDependentRV}}

Throughout this section, the notation is intended to be local with
no reference to the rest of the paper. This is to avoid the introduction
of too many new symbols. We use $P_{n}$ to denote the empirical law
of a stationary sample $\left\{ X_{i}:i=1,2,..,n\right\} $ with values
in a set $\mathcal{X}$. The following blocking technique is standard
(Yu, 1994, Rio, 2017, Ch.8). Divide the sample into $2m$ nonoverlapping
blocks of size $q$ each, plus a residual block of size $n-2mq$.
To avoid distracting technicalities and notational trivialities in the remaining of the paper we suppose that $2mq=n$. The general case with a remainder is dealt with in the aforementioned references. Define $H_{i}=\left\{ j\leq n:1+2\left(i-1\right)q\leq j\leq\left(2i-1\right)q\right\}$, $i\in\left\{ 1,2,...,m\right\} $. Let $\left(\tilde{X}_{j}\right)_{j\in H_{i}}$ be random variables with the same law as $\left(X_{j}\right)_{j\in H_{i}}$
and s.t. $\left(\tilde{X}_{j}\right)_{j\in H_{i}}$ and $\left(\tilde{X}_{j}\right)_{j\in H_{k}}$
are independent for $i\neq k$. Define $P_{q,m}$ to be the measure
that assigns weight $2/n$ to each variable in $\left\{ \left(\tilde{X}_{j}\right):j\in\bigcup_{i=1}^{m}H_{i}\right\} $.
Hence, the empirical measure $P_{q,m}$ is over variables within blocks
of cardinality $q$, separated by $q$ variables and such that each
block of variables is independent of each other. The following is
a rephrasing of Lemma 4.1 Yu (1994).

\begin{lemma}\label{Lemma_couplingArconesYu}Using the above notation,
let 
\[
X_{H}:=\left\{ \left(X_{j}\right):j\in\bigcup_{i=1}^{m}H_{i}\right\} \text{ and }\tilde{X}_{H}:=\left\{ \left(\tilde{X}_{j}\right):j\in\bigcup_{i=1}^{m}H_{i}\right\} .
\]
For any measurable function $G$ on $\mathcal{X}^{qm}$, uniformly
bounded in $\left[0,1\right]$, $\mathbb{E}G\left(X_{H}\right)\leq\mathbb{E}G\left(\tilde{X}_{H}\right)+m\beta_{q}.$
\end{lemma}

\begin{lemma}\label{Lemma_maximalIneqQDep}Suppose that $\mathcal{F}$
is a class of functions and that, for every discrete probability measure
$\mathbb{Q}$ on $\mathcal{X}$, there is a number $D_{\mathbb{Q}}$
such that $\sup_{f\in\mathcal{F}}|f|_{2,\mathbb{Q}}\leq D_{\mathbb{Q}}$
and $\sup_{\mathbb{Q}:D_{\mathbb{Q}}>0}\ln N\left(\epsilon D_{\mathbb{Q}},\mathcal{F},|\cdot|_{2,\mathbb{Q}}\right)\leq A_{1}\ln\left(A_{2}/\epsilon\right)$
for some $A_{1}>1$, $A_{2}>e$; $\epsilon\in\left(0,1\right)$. Then,
for any $x>0$,

\[
\Pr\left(\sup_{f\in\mathcal{F}}\left|\left(P_{q,m}-P\right)f\right|\geq x\right)\lesssim\frac{\mathbb{E}D_{P_{q,m}}\sqrt{1+A_{1}\ln A_{2}}}{x\sqrt{m}}.
\]
When there is an envelope function $F$, we can replace $\ln N\left(\epsilon D_{\mathbb{Q}},\mathcal{F},|\cdot|_{2,\mathbb{Q}}\right)$
with $\ln N\left(\epsilon\left|F\right|_{2,\mathbb{Q}},\mathcal{F},|\cdot|_{2,\mathbb{Q}}\right)$
and $\mathbb{E}D_{P_{q,m}}$ with $\left|F\right|_{2,P}$ in the above
display.\end{lemma}

\begin{proof}Define $Z_{i,q}:=\left(\tilde{X}_{j}\right)_{j\in H_{i}}$
and let $P_{m|Z}$ to be the law that assigns weight $1/m$ to each
variable in $\left\{ Z_{i,q}:i=1,2,...,m\right\} $. Hence, the variables
$Z_{i,q}$ are independent of each other. Let $P_{Z}$ be the law
of $Z_{i,q}$. For any class of functions $\mathcal{F}$ on $\mathcal{X}$,
let $\mathcal{F}\left(q\right):=\left\{ g\left(z\right)=\frac{1}{q}\sum_{l=1}^{q}f\left(z_{l}\right):f\in\mathcal{F},z\in\mathcal{X}^{q}\right\} $;
$\mathcal{X}^{q}=\otimes_{l=1}^{q}\mathcal{X}$ and if $z$ is a $q$-dimensional
tuple of elements in $\mathcal{X}$, $z_{l}$ is the $l^{th}$ entry.
Then, the set $\left\{ \sup_{f\in\mathcal{F}}\left|\left(P_{q,m}-P\right)f\right|\geq x\right\} $
is contained in the set $\left\{ \sup_{f\in\mathcal{F}\left(q\right)}\left|P_{m|Z}-P_{Z}f\right|\geq x\right\} $.
From the first display on page 240 of van der Vaart and Wellner (2000)
we deduce that 
\[
\mathbb{E}\left|\sqrt{m}\sup_{f\in\mathcal{F}\left(q\right)}\left|\left(P_{m|Z}-P_{Z}\right)f\right|\right|\lesssim\mathbb{E}\int_{0}^{\eta\left(q\right)}\sqrt{1+\ln N\left(\epsilon,\mathcal{F}\left(q\right),\left|\cdot\right|_{2,P_{m|Z}}\right)}d\epsilon,
\]
where $\eta\left(q\right):=\sup_{f\in\mathcal{F}\left(q\right)}\left|f\right|_{2,P_{m|Z}}$.
By Jensen inequality, $\eta\left(q\right)\leq\eta:=\sup_{f\in\mathcal{F}}|f|_{2,P_{q,m}}$.
Moreover, for $h_{1},h_{2}\in\mathcal{F}\left(q\right)$, such that
$h_{k}\left(z\right)=\frac{1}{q}\sum_{l=1}^{q}f_{k}\left(z_{l}\right)$,
$f_{k}\in\mathcal{F}$, $k=1,2$, using again Jensen inequality, $\left|h_{1}-h_{2}\right|_{2,P_{m|Z}}^{2}\leq\left|f_{1}-f_{2}\right|_{2,P_{q,m}}^{2}$,
using the definition of $P_{q,m}$. We can then deduce that the r.h.s.
of the above display is less than 
\[
\mathbb{E}\int_{0}^{\eta}\sqrt{1+\ln N\left(\epsilon,\mathcal{F},\left|\cdot\right|_{2,P_{q,m}}\right)}d\epsilon\leq\mathbb{E}\left[D_{P_{q,m}}\int_{0}^{1}\sqrt{1+\ln N\left(\epsilon D_{P_{q,m}},\mathcal{F},\left|\cdot\right|_{2,P_{q,m}}\right)}d\epsilon\right]
\]
where the second inequality follows by change of variables and using the upper bound $\eta\leq D_{P_{q,m}}$. Using the covering-number assumption, the above display is bounded by
\[
\mathbb{E}D_{P_{q,m}}\int_{0}^{1}\sqrt{1+A_{1}\ln\left(\frac{A_{2}}{\epsilon}\right)}d\epsilon\lesssim\mathbb{E}D_{P_{q,m}}\sqrt{1+A_{1}\ln A_{2}}.
\]
Applying Markov's inequality we obtain the result. 

The last statement in the lemma is obtained bounding $\eta$ by $|F|_{2,P_{q,m}}$
and using Jensen inequality as in the proof of Theorem 2.14.1 of van
der Vaart and Wellner (2000).\end{proof}

\subsection{Bounds on Uniform Covering Numbers of Function Classes\protect\label{Section_CoveringNumbers} }

We need to introduce some additional notation. In particular, for
later reference, we introduce a relatively long list of function classes
that will be used in the proof.

\subsubsection{Function Classes\protect\label{Section_functionClasses}}

We introduce the following function classes, some of which have already
been defined previously:

$\mathcal{F}_{B}$:= Real valued functions $h$ on $\mathcal{S}$
s.t. $h=\left[\sum_{k=1}^{K}b_{k}\varphi_{k}\left(s\right)\right]_{B}$
(see (\ref{EQ_cappedSpaceFunctions}));

$\mathcal{F}_{B}^{\mathcal{A}}$:= Real valued functions $h$ on $\mathcal{S}\times\mathcal{A}$
s.t. $\mathcal{S}\times\mathcal{A}\ni\left(s,a\right)\mapsto h\left(s,a\right)=\sum_{a'\in\mathcal{A}}f_{a'}\left(s\right)1_{\left\{ a'=a\right\} }$
where each $f_{a'}$ is in $\mathcal{F}_{B}$;

$\mathcal{F}_{B}^{\max}$:= Real valued functions $h$ on $\mathcal{S}$
s.t. $h\left(s\right)=\max_{a\in\mathcal{A}}f\left(s,a\right)$ where
$f\in\mathcal{F}_{B}^{\mathcal{A}}$;

$\mathbb{K}^{a}\mathcal{\mathcal{F}^{\max}}$:= Real valued functions
$h$ on $\mathcal{S}$ s.t. $h\left(s,a\right)=\left(\mathbb{K}^{a}f\right)\left(s\right)$
where $f\in\mathcal{F}_{B}^{\max}$;

$\mathbb{K}^{\max}\mathcal{\mathcal{F}^{\max}}$:= Real valued functions
$h$ on $\mathcal{S}$ s.t. $h\left(s\right)=\max_{a\in\mathcal{A}}\left(\mathbb{K}^{a}f\right)\left(s\right)$
where $f\in\mathcal{F}_{B}^{\max}$;

If $f\in\mathcal{F}_{B}^{\mathcal{A}}$, then $f\left(\cdot,a\right)=\sum_{a'\in\mathcal{A}}f_{a'}\left(\cdot\right)1_{\left\{ a'=a\right\} }=f_{a}$
where $f_{a}\in\mathcal{F}_{B}$ and this implies that $\sum_{a\in\mathcal{A}}f\left(\cdot,a\right)=\sum_{a\in\mathcal{A}}f_{a}$.
Hence, each element in $\mathcal{F}_{B}^{\mathcal{A}}$ is a collection
of $\left|\mathcal{A}\right|$ elements in $\mathcal{F}_{B}$ where
the second argument in $f\in\mathcal{F}_{B}^{\mathcal{A}}$ says which
element is to be picked. We also have that $\Pi_{n,B}f\in\mathbf{\mathcal{F}}_{B}$
for any real valued function on $\mathcal{S}$. Note that $\Pi_{n,B}$
is not a linear operator.

\subsubsection{Lemmas on Covering Numbers of the Function Classes}

The aim of this section is to compute the covering number of the classes
of functions defined in Section \ref{Section_functionClasses}. At
first, we break down the calculation in terms of simpler function
classes. In what follows, for any classes of functions $\mathcal{F}$
and $\mathcal{G}$, $\mathcal{F}\oplus\mathcal{G}:=\left\{ h=f+g:f\in\mathcal{F},g\in\mathcal{G}\right\} $
and $\mathcal{F}\otimes\mathcal{G}:=\left\{ h=fg:f\in\mathcal{F},g\in\mathcal{G}\right\} $.

\begin{lemma}\label{Lemma_coveringSumProd}Suppose that $\mathbb{Q}$
is a probability measure on a set $\mathcal{X}$ and $\mathcal{F}$
and $\mathcal{G}$ are classes of functions on $\mathcal{X}$ with
an envelope function $F$ and $G$, respectively. Then, for any strictly
positive constants $c_{1}$, $c_{2}$ s.t. $c_{1}+c_{2}=1$, we have
that $N\left(\epsilon,\mathcal{F}\oplus\mathcal{G},\left|\cdot\right|_{p,\mathbb{Q}}\right)\leq N\left(c_{1}\epsilon,\mathcal{F},\left|\cdot\right|_{p,\mathbb{Q}}\right)N\left(c_{2}\epsilon,\mathcal{G},\left|\cdot\right|_{p,\mathbb{Q}}\right)$.
If the envelope functions are uniformly bounded, we have that $N\left(\epsilon,\mathcal{F}\otimes\mathcal{G},\left|\cdot\right|_{p,\mathbb{Q}}\right)\leq N\left(\epsilon c_{1}\left|G\right|_{\infty}^{-1},\mathcal{F},\left|\cdot\right|_{p,\mathbb{Q}}\right)N\left(\epsilon c_{2}\left|F\right|_{\infty}^{-1},\mathcal{G},\left|\cdot\right|_{p,\mathbb{Q}}\right)$
\end{lemma}

\begin{proof}For the $L_{1}$ empirical norm $\left|\cdot\right|_{1,P_{n}}$,
the statements are Theorems 29.6 and 29.7 in Devroye et al. (1997).
Inspection of the proofs show that the extension to $\left|\cdot\right|_{p,\mathbb{Q}}$
is immediate by application of the triangle inequality.\end{proof}

\begin{lemma}\label{Lemma_coveringFB}Let $F$ be an envelope function
for $\mathcal{F}_{B}$. Then, $\sup_{\mathbb{Q}}\ln N\left(\epsilon\left|F\right|_{2,\mathbb{Q}},\mathcal{F}_{B},\left|\cdot\right|_{2,\mathbb{Q}}\right)\lesssim K\ln\left(\frac{C}{\epsilon}\right)$,
$\epsilon\in\left(0,1\right)$ for some $C>1$, where the supremum
is over all discrete probability measures on $\mathcal{S}$ s.t. $\left|F\right|_{2,\mathbb{Q}}>0$.
\end{lemma}

\begin{proof}The functions in $\mathcal{F}_{B}$ are nondecreasing
transformation of the functions in $\mathcal{F}$. Hence, they have
the same VC index/pseudo dimension (Theorem 11.3 in Anthony and Bartlett,
1999; see also Lemma 2.6.18(viii) in van der Vaart and Wellner, 2000).
Given that $\mathcal{F}$ is a $K$-dimensional vector, its pseudo
dimension is $K$ (Theorem 11.4 in Anthony and Bartlett, 1999; see
also Lemma 2.6.15 in van der Vaart and Wellner, 2000). Then, we apply
Theorem 2.6.7 in van der Vaart and Wellner (2000) to bound the uniform
covering number by the pseudo dimension, as given in the statement
of the lemma.\end{proof}

The above result allows us to control the following function classes.

\begin{lemma}\label{Lemma_coveringBasicClasses}Under the assumptions
of Lemma \ref{Lemma_coveringFB} we have the following where $\epsilon\in\left(0,1\right)$:\\
 1. ${\displaystyle \sup_{\mathbb{Q}}\ln N\left(\epsilon B,\mathcal{F}_{B}^{\max},|\cdot|_{p,\mathbb{Q}}\right)\lesssim|\mathcal{A}|K\ln\left(\frac{|\mathcal{A}|}{\epsilon}\right)}$;\\
 2. ${\displaystyle \sup_{\mathbb{Q}}\ln N\left(\epsilon B,\mathbb{K}^{a}\mathcal{F}_{B}^{\max},|\cdot|_{p,\mathbb{Q}}\right)\lesssim|\mathcal{A}|K\ln\left(\frac{|\mathcal{A}|}{\epsilon}\right)}$;\\
 3. ${\displaystyle \sup_{\mathbb{Q}}\ln N\left(\epsilon B,\mathbb{K}^{\max}\mathcal{F}_{B}^{\max},|\cdot|_{p,\mathbb{Q}}\right)\lesssim|\mathcal{A}|K\ln\left(\frac{|\mathcal{A}|}{\epsilon}\right)}$;\\
 4. ${\displaystyle \ln N\left(\epsilon B,h\mathcal{F}_{B},|\cdot|_{p,\mathbb{Q}}\right)\lesssim K\ln\left(\frac{M}{\epsilon}\right)}$
where $h\mathcal{F}_{B}$ is the set of functions $hf$ where $f\in\mathcal{F}_{B}$
and $h$ is a fixed function whose absolute value is bounded by $M$; $\epsilon<M$ if $M<1$;\\
 5. $N\left(\epsilon,\mathbb{K}^{\max}f,\left|\cdot\right|_{p,\mathbb{Q}}\right)=1$
for any fixed function $f$.\end{lemma}

\begin{proof}We prove each point separately.

\paragraph{Point 1. }

Note that $\max_{a\in\mathcal{A}}f\left(\cdot,a\right)\in\mathcal{F}_{B}^{\max}$
for $f\in\mathcal{F}_{B}^{\mathcal{A}}$ . For any $f,g\in\mathcal{F}_{B}^{\mathcal{A}}$,
\begin{align*}
\left|\max_{a\in\mathcal{A}}f\left(\cdot,a\right)-\max_{a\in\mathcal{A}}g\left(\cdot,a\right)\right|_{p,\mathbb{Q}}\leq & \left|\max_{a\in\mathcal{A}}\left|f\left(\cdot,a\right)-g\left(\cdot,a\right)\right|\right|_{p,\mathbb{Q}}\\
\leq & \left|\mathcal{A}\right|^{1/p}\max_{a\in\mathcal{A}}\left|f\left(\cdot,a\right)-g\left(\cdot,a\right)\right|_{p,\mathbb{Q}}.
\end{align*}
Then, note that for $f,g\in\mathcal{F}_{B}^{\mathcal{A}}$, we have
that $f\left(\cdot,a\right)=f_{a}$ and $g\left(\cdot,a\right)=g_{a}$,
where $f_{a},g_{a}\in\mathcal{F}_{B}$. This follows from the definition
of $\mathcal{F}_{B}^{\mathcal{A}}$. Then, by definition of functions
in $\mathcal{F}_{B}^{\max}$, the $\epsilon$-covering number of $\mathcal{F}_{B}^{\max}$
is bounded by $\left[N\left(\epsilon/\left|\mathcal{A}\right|^{1/p},\mathcal{F}_{B},\left|\cdot\right|_{p,\mathbb{Q}}\right)\right]^{\left|\mathcal{A}\right|}$.
Using Lemma \ref{Lemma_coveringFB} and taking the logarithm gives
the result.

\paragraph{Point 2.}

\noindent For any $f,g\in\mathcal{F}_{B}^{\max}$, Jensen's inequality
gives
\[
\left|\mathbb{K}^{a}f-\mathbb{K}^{a}g\right|_{p,\mathbb{Q}}^{p}\leq\mathbb{Q}\mathbb{K}^{a}\left|f-g\right|^{p}\leq\left|f-g\right|_{p,\mathbb{Q}\mathbb{K}^{a}}^{p},
\]
 where $\mathbb{Q}\mathbb{K}^{a}$ the probability measure such that
$\mathbb{Q}\mathbb{K}^{a}f=\int_{\mathcal{S}}\int_{\mathcal{S}}f\left(s\right)\kappa^{a}\left(r,ds\right)d\mathbb{Q}\left(r\right)$.
Therefore, 
\[
N\left(\epsilon B,\mathbb{K}^{a}\mathcal{F}_{B}^{\max},|\cdot|_{p,\mathbb{Q}}\right)\leq N\left(\epsilon B,\mathcal{F}_{B}^{\max},|\cdot|_{p,\mathbb{Q}\mathbb{K}^{a}}\right)
\]
Since Point 1 holds for any probability measure on $\mathcal{S}$,
the result follows.

\paragraph{Proof of Point 3. }

\noindent For any $f,g\in\mathcal{F}_{B}^{\max}$, 
\[
\left|\max_{a\in\mathcal{A}}\mathbb{K}^{a}f-\max_{a\in\mathcal{A}}\mathbb{K}^{a}g\right|_{p,\mathbb{Q}}^{p}\leq\sum_{a\in\mathcal{A}}\left|\mathbb{K}^{a}f-\mathbb{K}^{a}g\right|_{p,\mathbb{Q}}^{p}\leq\sum_{a\in\mathcal{A}}\mathbb{Q}\mathbb{K}^{a}\left|f-g\right|^{p}.
\]
Define the probability measure $\bar{\mathbb{Q}}=|\mathcal{A}|^{-1}\sum_{a\in\mathcal{A}}\mathbb{Q}\mathbb{K}^{a}$.
Then, 
\[
\left|\max_{a\in\mathcal{A}}\mathbb{K}^{a}f-\max_{a\in\mathcal{A}}\mathbb{K}^{a}g\right|_{p,\mathbb{Q}}\leq\left|\mathcal{A}\right|^{1/p}\left|f-g\right|_{p,\bar{\mathbb{Q}}}.
\]
Therefore, $N\left(\epsilon B,\mathbb{K}^{\max}\mathcal{F}_{B}^{\max},\left|\cdot\right|_{p,\mathbb{Q}}\right)\leq N\left(\frac{\epsilon B}{|\mathcal{A}|^{1/p}},\mathcal{F}_{B}^{\max},\left|\cdot\right|_{p,\bar{\mathbb{Q}}}\right)$.
Then, apply Point 1 to the $\left|\mathcal{A}\right|^{-1/p}\epsilon B$-covering
number instead of the $\epsilon B$-covering number. Taking logarithms,
the $1/p$ exponent can be absorbed in the implicit constant in $\lesssim$.

\paragraph{Proof of Point 4.}

This is a special case of Lemma \ref{Lemma_coveringSumProd} because
$h$ is fixed and bounded.

\paragraph{Proof of Point 5.}

Given that $f$ is fixed, also $\mathbb{K}^{\max}f$ is fixed, so
that there is only one element and its covering number is one. \end{proof}

\subsection{Convergence Rates for the Capped Projection Operator }

At first, we introduce some properties of the capped projection operator.
The next is a property also shared by standard projections and useful
to bound their estimation error.

\begin{lemma}\label{Lemma_projBasicIneq} For any $f:\mathcal{S}\to\mathbb{R}$,
\begin{equation}
P_{n}f\Pi_{n}f=\left|\Pi_{n}f\right|_{2,P_{n}}^{2}+{\rm Pen}\left(\Pi_{n}f\right).\label{EQ_ridgeBasicEquality}
\end{equation}
\end{lemma} 

\begin{proof} Write $\Pi_{n}f=\varphi^{\mathrm{T}}\hat{b}$ where
$\hat{b}$ is the ridge regression coefficient that minimises $P_{n}\left(f-\varphi^{\mathrm{T}}b\right)^{2}+\rho b^{\mathrm{T}}b$.
The first order conditions give $P_{n}\varphi\left(f-\varphi^{\mathrm{T}}\hat{b}\right)=\rho\hat{b}$.
Multiplying both sides by $\hat{b}^{\mathrm{T}}$ we have that $P_{n}\hat{b}^{\mathrm{T}}\varphi f-P_{n}\hat{b}^{\mathrm{T}}\varphi\varphi^{\mathrm{T}}\hat{b}=\rho P_{n}\hat{b}^{\mathrm{T}}\hat{b}$.
Rearranging, we obtain (\ref{EQ_ridgeBasicEquality}). \end{proof}

The capped projection operator satisfies the following inequality.

\begin{lemma}\label{Lemma_projDiffIneq}For any real valued functions
$f,g$ on $\mathcal{S}$, and $B\geq0$, we have that
\[
|\Pi_{n,B}f\left(s\right)-\Pi_{n,B}g\left(s\right)|\le|\Pi_{n,2B}\left(f-g\right)\left(s\right)|\le|\Pi_{n}\left(f-g\right)\left(s\right)|
\]
 uniformly in $s\in\mathcal{S}$. Moreover, $|\Pi_{n,B}f\left(s\right)|\leq|\Pi_{n}f\left(s\right)|$.
\end{lemma} 

\begin{proof}By definition of the operator $\Pi_{n,B}$ and linearity
of $\Pi_{n}$, we have that $\Pi_{n,2B}\left(f-g\right)=\left[\Pi_{n}f-\Pi_{n}g\right]_{2B}$.
To deduce the result, note that $\left|\left[x\right]_{B}-\left[y\right]_{B}\right|\leq\left|\left[x-y\right]_{2B}\right|$
for any $x,y\in\mathbb{R}$. The last assertion is trivial because
$\Pi_{n,B}f=\left[\Pi_{n}f\right]_{B}$.\end{proof}

The result below is applied to the estimated value function. In fact,
the bounds hold for sample-dependent $f=\widehat{V}^{(j)}$ simultaneously
for all $j$, because the supremum is over the deterministic class
$\mathcal{F}_{B}^{\max}$.

\begin{lemma}\label{Lemma_projectionGeneralConvergence} Suppose
that the Assumptions hold. Define $\psi=G^{-1/2}\varphi$, where $G$
is the population Gram matrix as in the Assumptions. Recall that $\left|\cdot\right|_{2}$
is the Euclidean norm. We have that, uniformly in $a\in\mathcal{A}$,
\begin{align*}
\sup_{f\in\mathcal{F}_{B}^{\max}}\left|\left(P_{n}-P\right)\psi f\left(T^{a}\cdot\right)\right|_{2} & =O_{P}\left(B\sqrt{n^{-1}K\ln n}\right),\\
\sup_{f\in\mathcal{F}_{B}^{\max}}\left|\left(P_{n}-P\right)\psi\mathbb{K}^{a}f\right|_{2} & =O_{P}\left(B\sqrt{n^{-1}K\ln n}\right),\\
\left|\left(P_{n}-P\right)\psi R\left(T^{a}\cdot;\theta^{a}\left(\cdot\right)\right)\right|_{2} & =O_{P}\left(\sqrt{n^{-1}K\ln n}\right),\\
\left|\left(P_{n}-P\right)\psi\mathbb{K}^{a}R\right|_{2} & =O_{P}\left(\sqrt{n^{-1}K\ln n}\right).
\end{align*}
\end{lemma}

\begin{proof} Let 
\[
\mathcal{G}_{1}^{a}:=\left\{ b^{\mathrm{T}}\psi f\left(T^{a}\cdot\right):|b|_{2}\le1,\ f\in\mathcal{F}_{B}^{\max}\right\} 
\]
\[
\mathcal{G}_{2}^{a}:=\left\{ b^{\mathrm{T}}\psi\mathbb{K}^{a}f:|b|_{2}\le1,f\in\mathcal{F}_{B}^{\max}\right\} .
\]
Recall that for any $x\in\mathbb{R}^{K}$, $\left|x\right|_{2}=\sup_{|b|_{2}\leq1}|b^{\mathrm{T}}x|$.
Consequently, to bound the first two expressions in the lemma, we
consider the classes of functions in the above display. First, we
find an estimate of the covering numbers of $\mathcal{G}_{a}$. For
an arbitrary probability measure $\mathbb{Q}$ on $\mathcal{S}$,
let $\Lambda_{\mathbb{Q}}=\lambda_{\max}\left(\mathbb{Q}\psi\psi^{\mathrm{T}}\right)$,
where $\lambda_{\max}\left(\cdot\right)$ is the largest eigenvalue
of its matrix argument. For each $a\in\mathcal{A}$, we verify the
weighted entropy bound 
\begin{equation}
\sup_{\mathbb{Q}}\log N\left(\epsilon B\sqrt{\Lambda_{\mathbb{Q}}},\mathcal{G}_{j}^{a},|\cdot|_{2,\mathbb{Q}}\right)\lesssim|\mathcal{A}|K\ln\left(|\mathcal{A}|/\epsilon\right)\quad j\in\left\{ 1,2\right\} .\label{EQ_weightedCoveringNumber}
\end{equation}
If $\Lambda_{\mathbb{Q}}=0$, $\psi=0$ $\mathbb{Q}$-almost surely
so that all functions in $\mathcal{G}_{a}$ vanish $\mathbb{Q}$-almost
surely.

The difference between two elements in $\mathcal{G}_{a,1}$ can be
written as $b^{\mathrm{T}}\psi f\left(T^{a}\cdot\right)-c^{\mathrm{T}}\psi g\left(T^{a}\cdot\right)$
where $b,c\in\left\{ b:\left|b\right|_{2}\le1\right\} $ and $f,g\in\mathcal{F}_{B}^{\max}$.
Then, 
\begin{equation}
\left|b^{\mathrm{T}}\psi f\left(T^{a}\cdot\right)-c^{\mathrm{T}}\psi g\left(T^{a}\cdot\right)\right|_{2,\mathbb{Q}}\leq\left|\left(b-c\right)^{\mathrm{T}}\psi f\left(T^{a}\cdot\right)\right|_{2,\mathbb{Q}}+\left|c^{\mathrm{T}}\psi\left(f\left(T^{a}\cdot\right)-g\left(T^{a}\cdot\right)\right)\right|_{2,\mathbb{Q}}.\label{EQ_differenceGa1Functions}
\end{equation}
The square of the first term on the r.h.s. of the above display is
\[
\mathbb{Q}\left[\left(b-c\right)^{\mathrm{T}}\psi f\left(T^{a}\cdot\right)\right]^{2}\leq B^{2}\mathbb{Q}\left[\left(b-c\right)^{\mathrm{T}}\psi\right]^{2}\leq B^{2}\left|b-c\right|_{2}^{2}\Lambda_{\mathbb{Q}}.
\]
The square of the second term in (\ref{EQ_differenceGa1Functions})
is 
\[
\mathbb{Q}\left[c^{\mathrm{T}}\psi\left(f\left(T^{a}\cdot\right)-g\left(T^{a}\cdot\right)\right)\right]^{2}=\mathbb{Q}\left(c^{\mathrm{T}}\psi\right)^{2}\mathbb{Q}_{c}\left(f\left(T^{a}\cdot\right)-g\left(T^{a}\cdot\right)\right)^{2}
\]
using the change of measure $d\mathbb{Q}_{c}=\left[\mathbb{Q}\left(c^{\mathrm{T}}\psi\right)^{2}\right]^{-1}\left(c^{\mathrm{T}}\psi\right)^{2}d\mathbb{Q}$.
We can assume that the denominator is positive. If it is zero, then
the corresponding function in $\mathcal{G}_{a,1}$ is zero and we
do not need to bound that function. Then, $\mathbb{Q}_{c}f=\left[\mathbb{Q}\left(c^{\mathrm{T}}\psi\right)^{2}\right]^{-1}\mathbb{Q}\left(c^{\mathrm{T}}\psi\right)^{2}f$.
Now, we use the fact that $\mathbb{Q}\left(c^{\mathrm{T}}\psi\right)^{2}\leq\Lambda_{\mathbb{Q}}$
because $\left|c\right|_{2}\leq1$ and 
\[
\mathbb{Q}_{c}\left(f\left(T^{a}\cdot\right)-g\left(T^{a}\cdot\right)\right)^{2}=\left|f-g\right|_{2,\mathbb{Q}_{c}\circ\left(T^{a}\right)^{-1}}^{2}
\]
to see that 
\[
\mathbb{Q}\left[c^{\mathrm{T}}\psi\left(f\left(T^{a}\cdot\right)-g\left(T^{a}\cdot\right)\right)\right]^{2}\leq\Lambda_{\mathbb{Q}}\left|f-g\right|_{2,\mathbb{Q}_{c}\circ\left(T^{a}\right)^{-1}}^{2}
\]
Putting everything together, we deduce that (\ref{EQ_differenceGa1Functions})
is bounded above by 
\[
\sqrt{\Lambda_{\mathbb{Q}}}\left[B\left|b-c\right|_{2}+\left|f-g\right|_{2,\mathbb{Q}_{c}\circ\left(T^{a}\right)^{-1}}\right].
\]
The above bound for (\ref{EQ_differenceGa1Functions}) implies that
an $\epsilon/2$-net of $\left\{ b:\left|b\right|_{2}\le1\right\} $
and an $\epsilon B/2$-cover of $\mathcal{F}_{B}^{\max}$ under the
measure $\mathbb{Q}_{c}\circ\left(T^{a}\right)^{-1}$ produces an
$\epsilon B\sqrt{\Lambda_{\mathbb{Q}}}$-cover of $\mathcal{G}_{1}^{a}$.
The $\epsilon/2$-net of the $K$-dimensional Euclidean ball has entropy
number proportional to $K\ln\left(1/\epsilon\right)$ and the $\epsilon B/2$-cover
of $\mathcal{F}_{B}^{\max}$ under the measure $\mathbb{Q}_{c}\circ\left(T^{a}\right)^{-1}$
has entropy number less than a constant multiple of $|\mathcal{A}|K\ln\left(|\mathcal{A}|/\epsilon\right)$
according to Point 1 in Lemma \ref{Lemma_coveringBasicClasses} because
the measure $\mathbb{Q}$ in that lemma is arbitrary, thus if can be taken to be $\mathbb{Q}_{c}\circ\left(T^{a}\right)^{-1}$. This
proves (\ref{EQ_weightedCoveringNumber}) for $j=1$. For the class
$\mathcal{G}_{2}^{a}$ we can use the same arguments applied to $\mathbb{K}^{a}f$
where $f\in\mathcal{F}_{B}^{\max}$ instead of $f\left(T^{a}\cdot\right)$.
By Jensen's inequality, as done in the proof of Point 2 we derive
the same result under the measure $\mathbb{Q}_{c}\mathbb{K}^{a}$
instead of $\mathbb{Q}_{c}\circ\left(T^{a}\right)^{-1}$. This proves
(\eqref{EQ_weightedCoveringNumber}).

We now apply the results in \ref{Section_inequalitiesDependentRV}.
To avoid inconsequential complications, we suppose that $2qm=n$.
Define the event 
\[
\mathcal{E}_{j}=\left\{ \sup_{f\in\mathcal{G}_{j}^{a}}|\left(P_{n}-P\right)f|>x\right\} \quad j\in\left\{ 1,2\right\} .
\]
By Lemma \ref{Lemma_couplingArconesYu} applied to the indicator function
$1_{\mathcal{E}_{j}}$, we deduce that 
\[
\Pr\left(\mathcal{E}_{j}\right)\leq2\Pr\left(\sup_{f\in\mathcal{G}_{j}^{a}}\left|\left(P_{q,m}-P\right)f\right|>x\right)+2m\beta_{q-1}.
\]
Recall that $P_{q,m}$ is the empirical law of the independent coupled blocks as defined in Section \ref{Section_inequalitiesDependentRV}.
We use $\beta_{q-1}$ and not $\beta_{q}$ because the use of the
shift operator the class definition reduces the distance between blocks
from $q$ to $q-1$. By Assumption \ref{Condition_mixing}, we choose
$q=\left(1/C_{\beta}\right)\ln n$ ($C_{\beta}$ as in Assumption
\ref{Condition_mixing}) so that $m\beta_{q}\rightarrow0$. This implies
that $m\asymp n/\ln n$ and we shall replace $m$ with the r.h.s.,
in due course. We now apply Lemma \ref{Lemma_maximalIneqQDep} to
the first term on the r.h.s. using the entropy number in (\eqref{EQ_weightedCoveringNumber})
to deduce that 
\begin{equation}
\Pr\left(\sup_{f\in\mathcal{G}_{j}^{a}}\left|\left(P_{q,m}-P\right)f\right|>x\right)\lesssim\frac{\left[\mathbb{E}B\sqrt{\Lambda_{P_{q,m}}}\right]\sqrt{1+|\mathcal{A}|K\ln\left(|\mathcal{A}|\right)}}{x\sqrt{m}}\label{EQ_maximalIneqGja}
\end{equation}
We have used the fact that $\sup_{f\in\mathcal{G}_{j}^{a}}\left|f\right|_{\mathbb{Q}}\leq B\sqrt{\Lambda_{\mathbb{Q}}}$
so that Lemma \ref{Lemma_maximalIneqQDep} is applied with $D_{\mathbb{Q}}=B\sqrt{\Lambda_{\mathbb{Q}}}$.
Recall that $\Lambda_{P_{q,m}}=\lambda_{\max}\left(P_{q,m}\psi\psi^{\mathrm{T}}\right)$.
By Lemma\ref{Lemma_normEmpiricalBound} the expectation of this quantity
is of order one. Thus, using Jensen inequality, and $m\asymp n/\ln n$,
(\ref{EQ_maximalIneqGja}) is of order $B\sqrt{n^{-1}K\ln n}$, which
proves the first two results in the lemma. 

The last two bounds in the lemma follow by the same arguments, applied
to the fixed reward functions, so there is no need to estimate the
entropy integral. In particular, repeating the calculations for a
single function $f$ with finite second moment, we have that 
\[
\mathbb{Q}\left[\left(b-c\right)^{\mathrm{T}}\psi f\right]^{2}\leq\mathbb{Q}\left[\left(b-c\right)^{\mathrm{T}}\psi f\right]^{2}\leq\left|b-c\right|_{2}^{2}\lambda_{\max}\left(\mathbb{Q}\psi\psi^{\mathrm{T}}f^{2}\right)
\]
which replaces the bound on the first term on the r.h.s. of (\ref{EQ_differenceGa1Functions});
the second term on the r.h.s. of (\ref{EQ_differenceGa1Functions})
is zero because the class of functions includes only $f$. So Lemma
\ref{Lemma_maximalIneqQDep} simplifies to 
\[
\Pr\left(\sup_{f\in\mathcal{G}_{j}^{a}}\left|\left(P_{q,m}-P\right)f\right|>x\right)\lesssim\frac{\mathbb{E}D_{P_{q,m}}}{x\sqrt{m}}
\]
with $D_{P_{q,m}}=\sqrt{\lambda_{\max}\left(P_{q,m}\psi\psi^{\mathrm{T}}f^{2}\right)}$.
The largest eigenvalue of a positive definite matrix is bounded by
the trace of the matrix so that it is sufficient to bound $\mathbb{E}\sqrt{P_{q,m}\left|\psi\right|_{2}f^{2}}$.
By definition of $\psi$ and the Assumptions $\left|\psi\right|_{2}^{2}\lesssim K$.
Hence it is sufficient to bound $\mathbb{E}\sqrt{KP_{q,m}f^{2}}$
which is $O\left(\sqrt{K}\right)$ by Jensen inequality and the moment
condition in the Assumptions.\end{proof}

The next lemma controls the normalised Gram matrix.

\begin{lemma}\label{Lemma_normEmpiricalBound} Define $\psi=G^{-1/2}\varphi$,
where $G$ is the population Gram matrix as in the Assumptions. Under
the Assumptions, 
\[
\left|P_{n}\psi\psi^{\top}-I_{K}\right|_{{\rm op}}=O_{P}\left(\sqrt{\frac{K\ln K\ln n}{n}}\right),
\]
$\Pr\left(\lambda_{\min}\big(P_{n}\psi\psi^{\top}\big)\ge\tfrac{1}{2}\right)\rightarrow1$
and, for $P_{q,m}$ as in Section \ref{Section_inequalitiesDependentRV}
with $m\asymp n^{-1}\ln n$, $\mathbb{E}\lambda_{\max}\left(P_{q,m}\psi\psi^{\mathrm{T}}\right)\lesssim1$.\end{lemma}

\begin{proof}Construct independent blocks as in the proof of Lemma \ref{Lemma_projectionGeneralConvergence} so that the error incurred in the substitution is negligible, in probability assuming $n=2qm$ for simplicity with $m\asymp n/\ln n$. As discussed in the proof of Lemma \ref{Lemma_projectionGeneralConvergence}, $P_{q,m}\psi\psi^{\mathrm{T}}$
is the average of $m$ independent matrices where each matrix is itself the average of $q$ matrices with expectation $I_{K}$ and maximum eigenvalue bounded by a constant multiple of $K$. This means that $P_{q,m}\psi\psi^{\mathrm{T}}$ can be written as $\frac{1}{m}\sum_{i=1}^{m}M_{i}\left(q\right)$
where $M_{i}\left(q\right)=\frac{1}{q}\sum_{j\in H_{i}}\psi\left(S_{j}\right)\psi^{\mathrm{T}}\left(S_{j}\right)$
with $H_{i}$ as in Section \ref{Section_inequalitiesDependentRV}.
Then, $M_{i}\left(q\right)-I_{K}$ has mean zero, largest eigenvalue
equal to a constant multiple of $K$. Moreover, $\left|\mathbb{E}\left(M_{i}\left(q\right)-I_{K}\right)^{2}\right|_{{\rm op}}=\left|\mathbb{E}M_{i}\left(q\right)^{2}-I_{K}\right|_{{\rm op}}$.
Then, for any conformable column-vector $x$ with unit Euclidean norm,
$x^{\mathrm{T}}\left(\mathbb{E}M_{i}\left(q\right)^{2}-I_{K}\right)x\leq x^{\mathrm{T}}\mathbb{E}M_{i}\left(q\right)^{2}x$.
Since $M_{i}\left(q\right)$ is positive definite and the largest
eigenvalue of $M_{i}\left(q\right)$ is a constant multiple of $K$,
we have that $x^{\mathrm{T}}\mathbb{E}M_{i}\left(q\right)^{2}x\lesssim Kx^{\mathrm{T}}\mathbb{E}M_{i}\left(q\right)x=K$
because $\mathbb{E}M_{i}\left(q\right)=I_{K}$. This implies that
$\left|\mathbb{E}\left(M_{i}\left(q\right)-I_{K}\right)^{2}\right|_{{\rm op}}\lesssim K$.
Hence, we apply Remark 6.5 in Tropp (2012) to the average of these
$m$ independent matrices to deduce that 
\[
\mathbb{E}\left|P_{q,m}\psi\psi^{\mathrm{T}}-I_{K}\right|_{\mathrm{op}}\lesssim\sqrt{\frac{K\ln K}{m}}+\frac{K\ln K}{m}.
\]
Since $m\asymp n/\ln n$, the result follows. The second assertion
follows from the fact that the above display goes to zero and $P_{q,m}\psi\psi^{\mathrm{T}}=I_{K}+\left(P_{q,m}\psi\psi^{\mathrm{T}}-I_{K}\right)$. 

To show the third statement, we use the triangle inequality and the
inequality in the above display: 
\[
\mathbb{E}\left|P_{q,m}\psi\psi^{\mathrm{T}}\right|_{\mathrm{op}}\leq\left|I_{K}\right|_{\mathrm{op}}+\mathbb{E}\left|P_{q,m}\psi\psi^{\mathrm{T}}-I_{K}\right|_{\mathrm{op}}\lesssim1
\]
noting that $m\asymp n^{-1}\ln n$ and that, by the Assumptions, $n^{-1}K\ln K\ln n=o\left(1\right)$.
\end{proof}

The following bounds the norm of a ridge projection by the Euclidean
norm of a partial sum. 

\begin{lemma}\label{Lemma_projPopNorm2EmpNorm}Define $\psi=G^{-1/2}\varphi$,
where $G$ is the population Gram matrix as in the Assumptions. Under
the Assumptions, on the event $\mathcal{E}:=\left\{ \lambda_{\min}\left(P_{n}\psi\psi^{\mathrm{T}}\right)\ge1/2\right\} $,
for any real valued $f$ on $\mathcal{S}$, $\left|\Pi_{n}f\right|_{2,P}\leq2\left|P_{n}\psi f\right|_{2}$.\end{lemma}

\begin{proof}The ridge projection can be written as $\Pi_{n}f=\hat{b}^{\mathrm{T}}\psi$,
where
\begin{equation}
\hat{b}=\left(P_{n}\psi\psi^{\mathrm{T}}+\rho G^{-1}\right)^{-1}P_{n}\psi f,\label{EQ_ridgeCoefficientMatRep}
\end{equation}
exploiting the representation in terms of $\psi=G^{-1/2}\varphi$;
$\rho$ is the ridge penalty. Computing the norm, we have that 
\[
\left|\Pi_{n}f\right|_{2,P}^{2}=\left|\hat{b}^{\mathrm{T}}\psi\right|_{2,P}^{2}=\hat{b}^{\mathrm{T}}\left(P\psi\psi^{\mathrm{T}}\right)\hat{b}=\left|\hat{b}\right|_{2}^{2}
\]
because $P\psi\psi^{\mathrm{T}}=I_{K}$ by construction; recall that
$\left|\cdot\right|_{2}$ is the Euclidean norm. Since $\rho G^{-1}$
is positive semidefinite, on the event $\mathcal{E}:=\left\{ \lambda_{\min}\left(P_{n}\psi\psi^{\mathrm{T}}\right)\ge1/2\right\} $,
we have that $\left|\left(P_{n}\psi\psi^{\mathrm{T}}+\rho G^{-1}\right)^{-1}\right|_{{\rm op}}\leq2$.
On $\mathcal{E}$, using (\ref{EQ_ridgeCoefficientMatRep}), we have
that $\left|\hat{b}\right|_{2}\leq2\left|P_{n}\psi f\right|_{2}$.\end{proof}

\subsection{Proof of Theorem \ref{Theorem_convergenceAlgo}}

We shall show that the estimation error can be bounded using the control
of the quantities stated in the next few lemmas.

\begin{lemma}\label{Lemma_projectionConvergence} Under the Assumptions
\[
\max_{a\in\mathcal{A}}\sup_{f\in\mathcal{F}_{B}^{\max}}\left|\Pi_{n,B}\Gamma_{n}^{a}f-\Pi_{n,B}\mathbb{T}^{a}f\right|_{2,P}=O_{P}\left(\left(1+\gamma B\right)\sqrt{\frac{K\ln n}{n}}\right).
\]
\end{lemma} 

\begin{proof} By Lemma \ref{Lemma_projDiffIneq}, 
\[
\max_{a\in\mathcal{A}}\sup_{f\in\mathcal{F}_{B}^{\max}}\left|\Pi_{n,B}\Gamma_{n}^{a}f-\Pi_{n,B}\mathbb{T}^{a}f\right|_{2,P}\le\max_{a\in\mathcal{A}}\sup_{f\in\mathcal{F}_{B}^{\max}}\left|\Pi_{n}\left(\Gamma_{n}^{a}-\mathbb{T}^{a}\right)f\right|_{2,P}.
\]
On the event $\mathcal{E}$ of Lemma \ref{Lemma_projPopNorm2EmpNorm},
applying that lemma to the r.h.s. of the above display, we deduce
that 
\[
\max_{a\in\mathcal{A}}\sup_{f\in\mathcal{F}_{B}^{\max}}\left|\Pi_{n,B}\Gamma_{n}^{a}f-\Pi_{n,B}\mathbb{T}^{a}f\right|_{2,P}\leq2\max_{a\in\mathcal{A}}\sup_{f\in\mathcal{F}_{B}^{\max}}\left|P_{n}\psi\left(\Gamma_{n}^{a}-\mathbb{T}^{a}\right)f\right|_{2}
\]
where 
\[
\left(\Gamma_{n}^{a}-\mathbb{T}^{a}\right)f=\left[R\left(T^{a}\cdot;\theta^{a}\left(\cdot\right)\right)-\mathbb{K}^{a}R\right]+\gamma\left[f\left(T^{a}\cdot\right)-\mathbb{K}^{a}f\right]
\]
by definition of the operators. The population mean of each quantity
in the square brackets is zero by the conditional expectation operator
$\mathbb{K}^{a}$. Hence, the bounds of Lemma \ref{Lemma_projectionGeneralConvergence}
give the claimed rate. Finally, the event $\mathcal{E}$ has probability
tending to one by Lemma \ref{Lemma_normEmpiricalBound}. \end{proof}

\begin{lemma}\label{Lemma_approximationError} Under the Assumptions,
\begin{align*}
\max_{a\in\mathcal{A}}\sup_{f\in\mathcal{F}_{B}^{\max}}\left|\Pi_{n,B}\mathbb{T}^{a}f-\mathbb{T}^{a}f\right|_{2,P}\le & \max_{a\in\mathcal{A}}\sup_{f\in\mathcal{F}_{B}^{\max}}\inf_{h\in\mathcal{F}}\left|\mathbb{T}^{a}f-h\right|_{2,P}\\
 & +\max_{a\in\mathcal{A}}\left|\big(|\mathbb{K}^{a}R|-(1-\gamma)B\big)_{+}\right|_{2,P}\\
 & +O_{P}\left(\left(1+\gamma B\right)\left(\sqrt{\frac{K\ln K\ln n}{n}}+\rho\left|G^{-1}\right|_{{\rm op}}\right)\right).
\end{align*}
If $v>2$, $\max_{a\in\mathcal{A}}\left|\big(|\mathbb{K}^{a}R|-(1-\gamma)B\big)_{+}\right|_{2,P}=O\left(\left[\left(1-\gamma\right)B\right]^{-\left(v-2\right)/2}\right)$.
\end{lemma} 

\begin{proof} Fix $a,f$ and put $g=\mathbb{T}^{a}f$. Let $\Pi g$
be the population projection onto $\mathcal{F}$ with zero ridge coefficient.
By the triangle inequality, and then the fact that the capping function
is Lipschitz, we deduce that
\begin{align*}
\left|\Pi_{n,B}g-g\right|\leq & \left|\left[\Pi_{n}g\right]_{B}-\left[g\right]_{B}\right|+\left|\left[g\right]_{B}-g\right|\\
\leq & \left|\Pi_{n}g-g\right|+\left|\left[g\right]_{B}-g\right|\\
\leq & \left|\Pi_{n}g-\Pi g\right|+\left|\Pi g-g\right|+\left|\left[g\right]_{B}-g\right|
\end{align*}
where the last inequality follows from adding and subtracting $\Pi g$.
Using the above we bound 
\begin{align}
\max_{a\in\mathcal{A}}\sup_{f\in\mathcal{F}_{B}^{\max}}\left|\Pi_{n,B}\mathbb{T}^{a}f-\mathbb{T}^{a}f\right|_{2,P}\leq & \max_{a\in\mathcal{A}}\sup_{f\in\mathcal{F}_{B}^{\max}}\left|\left(\Pi_{n}-\Pi\right)\mathbb{T}^{a}f\right|_{2,P}\nonumber \\
 & +\max_{a\in\mathcal{A}}\sup_{f\in\mathcal{F}_{B}^{\max}}\inf_{h\in\mathcal{F}}\left|g-h\right|_{2,P}\nonumber \\
 & +\max_{a\in\mathcal{A}}\left|\left(\left|\mathbb{K}^{a}R\right|-\left(1-\gamma\right)B\right)_{+}\right|_{2,P},\label{EQ_decompositionEmpiricalProjection}
\end{align}
where we used $\left|\left[g\right]_{B}-g\right|=\left(\left|g\right|-B\right)_{+}\leq\left(\left|\mathbb{K}^{a}R\right|-\left(1-\gamma\right)B\right)_{+}$
because $g=\mathbb{T}^{a}f$ equals $\mathbb{K}^{a}R+\gamma\mathbb{K}^{a}f$
and $f\in\mathcal{F}_{B}^{\max}$ is bounded by $B$; for any variable
$z$ and constant $c$, $\left(z-c\right)_{+}=\max\left\{ z-c,0\right\} $.
We shall control each term in the above display, separately. To ease
notation continue to use $g=\mathbb{T}^{a}f$. Note that $\Pi g=\psi^{\top}b$
with $b=P\psi g$ and $\Pi_{n}g=\psi^{\top}\hat{b}$ where $\hat{b}$
is as in (\ref{EQ_ridgeCoefficientMatRep}). By linearity, $\left(\Pi_{n}-\Pi\right)g=\psi^{\mathrm{T}}\left(\hat{b}-b\right)$,
implying $\left|\left(\Pi_{n}-\Pi\right)g\right|_{2,P}=\left|\hat{b}-b\right|_{2}$
because $P\psi\psi^{\mathrm{T}}=I_{K}$. By similar arguments as in
the proof of Lemma \ref{Lemma_projectionConvergence}, the coefficients
satisfy 
\begin{equation}
\left(P_{n}\psi\psi^{\mathrm{T}}+\rho G^{-1}\right)\left(\widehat{b}-b\right)=P_{n}\psi\left(g-\Pi g\right)-\rho G^{-1}b.\label{EQ_bhatScaledEstimationError}
\end{equation}
To control the Euclidean norm of the difference in between the coefficients,
we first control the first term on the r.h.s. of the above display.
Adding and subtracting $b=P\psi g$ we have that
\[
P_{n}\psi\left(g-\Pi g\right)=P_{n}\left(\psi g-\psi\psi^{\top}b\right)=\left(P_{n}-P\right)\psi g-\left(P_{n}\psi\psi^{\top}-I_{K}\right)b.
\]
Substituting $g=\mathbb{T}^{a}f$ in the above display, we obtain
\begin{align*}
P_{n}\psi\left(g-\Pi g\right)= & \left(P_{n}-P\right)\psi\mathbb{K}^{a}R+\gamma\left(P_{n}-P\right)\psi\mathbb{K}^{a}f-\left(P_{n}\psi\psi^{\mathrm{T}}-I_{K}\right)b.
\end{align*}
By Lemma \ref{Lemma_projectionGeneralConvergence}, the Euclidean
norm of the first two terms on the r.h.s. is $O_{P}\left(\sqrt{n^{-1}K\ln n}\right)$
and $O_{P}\left(\gamma B\sqrt{n^{-1}K\ln n}\right)$, respectively,
uniformly in $f\in\mathcal{F}_{B}^{\max}$. By Lemma \ref{Lemma_normEmpiricalBound}
the last term on the r.h.s. of the display has Euclidean norm $O_{P}\left(\left|b\right|_{2}\sqrt{\frac{K\ln K\ln n}{n}}\right)$.
Since $b=P\psi g$, we have that $\left|b\right|_{2}\le\left|g\right|_{2,P}\leq\left|R\left(T^{a}\cdot;\theta^{a}\left(\cdot\right)\right)\right|_{2,P}+\gamma B$.
Now, the Euclidean norm of $\rho G^{-1}b$ is bounded by a quantity
$O\left(\left(1+\gamma B\right)\rho\left|G^{-1}\right|_{{\rm op}}\right)$.
This completes the control of the quantities on the r.h.s. of (\ref{EQ_bhatScaledEstimationError}).
Hence, on the set $\mathcal{E}:=\left\{ \lambda_{\min}\left(P_{n}\psi\psi^{\mathrm{T}}\right)\ge1/2\right\} $,
we have that, arguing as in the proof of Lemma \ref{Lemma_projectionConvergence},
we have that the first term in (\ref{EQ_decompositionEmpiricalProjection})
is bounded by 
\[
\left|\hat{b}-b\right|_{2}=O_{P}\left(\left(1+\gamma B\right)\left(\sqrt{\frac{K\ln K\ln n}{n}}+\rho\left|G^{-1}\right|_{{\rm op}}\right)\right).
\]
If Assumption \ref{Condition_moment-1} holds for $v>2$, we can further
simplify the last term in (\ref{EQ_decompositionEmpiricalProjection}).
In this case, note that for any positive random variable $Z$ with
finite $v>2$ moment and constant $C>0$, we have that $\mathbb{E}\left(Z-C\right)_{+}^{2}\leq\mathbb{E}Z^{2}1_{\left\{ Z>C\right\} }\lesssim C^{-\left(v-2\right)}$
using Holder inequality. We deduce that $\max_{a\in\mathcal{A}}\left|\left(\left|\mathbb{K}^{a}R\right|-\left(1-\gamma\right)B\right)_{+}\right|_{2,P}\lesssim\left[\left(1-\gamma\right)B\right]^{-\left(v-2\right)/2}$.
This completes the proof of the lemma.\end{proof}

The following is a crucial estimate of the error incurred by Algorithm
\ref{Algorithm_Sieve}. 

\begin{lemma}\label{Lemma_estimtionErrorUniform} Under the Assumptions,
for the fitted iterates of Algorithm \ref{Algorithm_Sieve}, for any
$J\geq1$,
\begin{align*}
\max_{j\le J}\left|\max_{a\in\mathcal{A}}\left|\hat{Q}^{(j)}\left(\cdot,a\right)-\mathbb{T}^{a}\hat{V}^{(j-1)}\right|\right|_{2,P}\leq & \sqrt{\left|\mathcal{A}\right|}\max_{a\in\mathcal{A}}\sup_{f\in\mathcal{F}_{B}^{\max}}\left|\Pi_{n,B}\Gamma_{n}^{a}f-\Pi_{n,B}\mathbb{T}^{a}f\right|_{2,P}\\
 & +\sqrt{\left|\mathcal{A}\right|}\max_{a\in\mathcal{A}}\sup_{f\in\mathcal{F}_{B}^{\max}}\left|\Pi_{n,B}\mathbb{T}^{a}f-\mathbb{T}^{a}f\right|_{2,P}.
\end{align*}
In particular, the r.h.s. is at most 
\begin{align*}
 & \sqrt{\left|\mathcal{A}\right|}\max_{a\in\mathcal{A}}\sup_{f\in\mathcal{F}_{B}^{\max}}\inf_{h\in\mathcal{F}}\left|\mathbb{T}^{a}f-h\right|_{2,P}+\sqrt{\left|\mathcal{A}\right|}\max_{a\in\mathcal{A}}\left|\big(|\mathbb{K}^{a}R|-(1-\gamma)B\big)_{+}\right|_{2,P}\\
 & +O_{P}\left(\left(1+\gamma B\right)\left(\sqrt{\frac{K\ln K\ln n}{n}}+\rho\left|G^{-1}\right|_{{\rm op}}\right)\right)
\end{align*}
If $v>2$, $\max_{a\in\mathcal{A}}\left|\big(|\mathbb{K}^{a}R|-(1-\gamma)B\big)_{+}\right|_{2,P}=O\left(\left[\left(1-\gamma\right)B\right]^{-\left(v-2\right)/2}\right)$.\end{lemma} 

\begin{proof}Using the definition $\hat{Q}^{(j)}\left(\cdot,a\right):=\Pi_{n,B}\Gamma_{n}^{a}\hat{V}^{(j-1)}$,
adding and subtracting $\Pi_{n,B}\mathbb{T}^{a}\hat{V}^{(j-1)}$ and
applying the triangle inequality we have that 
\[
\left|\hat{Q}^{(j)}\left(\cdot,a\right)-\mathbb{T}^{a}\hat{V}^{(j-1)}\right|\leq\left|\Pi_{n,B}\Gamma_{n}^{a}\hat{V}^{(j-1)}-\Pi_{n,B}\mathbb{T}^{a}\hat{V}^{(j-1)}\right|+\left|\Pi_{n,B}\mathbb{T}^{a}\hat{V}^{(j-1)}-\mathbb{T}^{a}\hat{V}^{(j-1)}\right|.
\]
The next step is to take the maximum over $a\in\mathcal{A}$, then
the $\left|\cdot\right|_{2,P}$ norm and then use the fact that the
squared maximum is bounded by the sum of squares. Performing these
calculations, we deduce that 
\begin{align*}
\left|\max_{a\in\mathcal{A}}\left|\hat{Q}^{(j)}\left(\cdot,a\right)-\mathbb{T}^{a}\hat{V}^{(j-1)}\right|\right|_{2,P}\leq & \sqrt{\left|\mathcal{A}\right|}\max_{a\in\mathcal{A}}\left|\Pi_{n,B}\Gamma_{n}^{a}\hat{V}^{(j-1)}-\Pi_{n,B}\mathbb{T}^{a}\hat{V}^{(j-1)}\right|_{2,P}\\
 & +\sqrt{\left|\mathcal{A}\right|}\max_{a\in\mathcal{A}}\left|\Pi_{n,B}\mathbb{T}^{a}\hat{V}^{(j-1)}-\mathbb{T}^{a}\hat{V}^{(j-1)}\right|_{2,P}.
\end{align*}
Every $\hat{V}^{(j-1)}$ belongs to $\mathcal{F}_{B}^{\max}$ (including
$\hat{V}^{(0)}=0$). Hence, taking the supremum for $\hat{V}^{(j-1)}\in\mathcal{F}_{B}^{\max}$
gives the first statement. The second statement is obtained applying
Lemmas \ref{Lemma_projectionConvergence} and \ref{Lemma_approximationError}.\end{proof}

The next result is a stability result for the transition kernel of
$S_{t}$. It also implies that the optimal value function $V^{*}$
in (\ref{EQ_optimalValueFunction}) is always well defined.

\begin{lemma}\label{Lemma_transitionStability} Define $\mathbb{K}f=\max_{a\in\mathcal{A}}\mathbb{K}^{a}f$
for any real valued function $f$ on $\mathcal{S}$ so that $\left(\mathbb{K}\right)^{j}f=\mathbb{K}\left(\left(\mathbb{K}\right)^{j-1}f\right)$,
$\left(\mathbb{K}\right)^{0}f=f$. Under the Assumptions, for any
square integrable real valued function $f$ on $\mathcal{S}$ and
every fixed integer $j\ge1$, we have that $\left|\left(\mathbb{K}\right)^{j}f\right|_{2,P}\le\underline{\alpha}^{-1/2}\left|f\right|_{2,P}$.

Moreover, for any policy $\pi$ and square integrable and real valued
function $f$ on $\mathcal{S}$, $\left|\left(\mathbb{K}^{\pi}\right)^{j}f\right|_{2,P}\le\underline{\alpha}^{-1/2}\left|f\right|_{2,P}$.\end{lemma} 

\begin{proof}By independence of the raw states from the simulated
actions, 
\[
\mathbb{K}f\left(S_{t}\right)=\max_{a\in\mathcal{A}}\mathbb{E}\left[f\left(\tilde{S}_{t+1},a\right)|S_{t}\right]\leq\mathbb{E}\left[\max_{a\in\mathcal{A}}f\left(\tilde{S}_{t+1},a\right)|S_{t}\right].
\]
Recall that $S_{t}=S_{t}^{A_{t-1}}$ where $A_{t-1}$ is the randomly
simulated action and that $S_{t}^{a}=\left(\tilde{S}_{t},a\right)$
for any $a\in\mathcal{A}$. Applying the same argument twice, 
\[
\left(\mathbb{K}\right)^{2}f\left(S_{t}\right)=\max_{a\in\mathcal{A}}\mathbb{E}\left[\left(\mathbb{K}f\right)\left(\tilde{S}_{t+1},a\right)|S_{t}\right]\leq\max_{a_{1}\in\mathcal{A}}\mathbb{E}\left[\mathbb{E}\left[\max_{a_{2}\in\mathcal{A}}f\left(\tilde{S}_{t+2},a_{2}\right)|S_{t+1}^{a_{1}}\right]|S_{t}\right].
\]
By the Markov assumption and the independence of the raw states from
the simulated actions, in the above display, $\tilde{S}_{t+2}$ only
depends on $\tilde{S}_{t+1}$ and not on $a_{1}$. Hence, conditioning
on $S_{t+1}^{a_{1}}$ is the same as conditioning on $\tilde{S}_{t+1}$,
making the outer maximum irrelevant. By the tower law, we have that
the r.h.s. is equal to $\mathbb{E}\left[\max_{a\in\mathcal{A}}f\left(\tilde{S}_{t+2},a\right)|S_{t}\right]$.
By induction, we can show that the argument holds for any $j\geq1$,
implying that 
\[
\left|\left(\mathbb{K}\right)^{j}f\left(S_{t}\right)\right|\leq\left(\mathbb{K}\right)^{j}\left|f\left(S_{t}\right)\right|\leq\mathbb{E}\left[\max_{a\in\mathcal{A}}\left|f\left(\tilde{S}_{t+j},a\right)\right||S_{t}\right].
\]
By Jensen's inequality and stationarity, 
\[
\left|\left(\mathbb{K}\right)^{j}f\right|_{2,P}^{2}\leq\mathbb{E}\max_{a\in\mathcal{A}}f\left(\tilde{S}_{t},a\right)^{2}\leq\mathbb{E}\sum_{a\in\mathcal{A}}f\left(\tilde{S}_{t},a\right)^{2}.
\]
Now, by the Assumptions, $\Pr\left(A_{t}=a|S_{t}\right)=\alpha\left(a|S_{t}\right)\geq\underline{\alpha}$,
implying that
\[
\mathbb{E}f\left(S_{t+1}\right)^{2}=\mathbb{E}\sum_{a\in\mathcal{A}}\alpha\left(a|S_{t}\right)f\left(\tilde{S}_{t+1},a\right)^{2}\geq\text{\underbar{\ensuremath{\alpha}}}\mathbb{E}\sum_{a\in\mathcal{A}}f\left(\tilde{S}_{t+1},a\right)^{2}.
\]
By stationarity, the above two displays imply that $\left|\left(\mathbb{K}\right)^{j}f\right|_{2,P}^{2}\leq\underbar{\ensuremath{\alpha}}^{-1}Pf^{2}$,
which gives the result once we take the square root. To show the second
statement, note that 
\[
\left|\left(\mathbb{K}^{\pi}\right)^{j}f\right|\leq\left(\mathbb{K}^{\pi}\right)^{j}\left|f\right|\leq\left(\mathbb{K}\right)^{j}\left|f\right|.
\]
In consequence, $\left|\left(\mathbb{K}^{\pi}\right)^{j}f\right|_{2,P}^{2}\leq\left|\left(\mathbb{K}\right)^{j}\left|f\right|\right|_{2,P}^{2}$
and the result follows from the first statement. \end{proof}

The following lemma follows the general error-propagation approach
of Munos and Szepesv\'{a}ri (2008, Lemmas 3-4), adapted to the fitted
action-value estimates and the state-action sampling structure considered
here.

\begin{lemma}\label{Lemma_recursionBound} Under the Assumptions,
for the policy produced by Algorithm \ref{Algorithm_Sieve} and any
$J\ge1$,
\[
\left|V^{*}-V^{\left(J\right)}\right|_{2,P}\leq\frac{2\underline{\alpha}^{-1}}{\left(1-\gamma\right)^{2}}\max_{j\le J}\left|\max_{a\in\mathcal{A}}\left|\hat{Q}^{\left(j\right)}\left(\cdot,a\right)-\mathbb{T}^{a}\hat{V}^{\left(j-1\right)}\right|\right|_{2,P}+\frac{2\underline{\alpha}^{-1}\gamma^{J}}{1-\gamma}\left|V^{*}\right|_{2,P},
\]
\end{lemma} 

\begin{proof}We need to bound $V^{*}-V^{\left(J\right)}$, which
is non negative by definition of $V^{*}$. To this end, recall the
notation $V^{(J)}=V^{\pi_{J}}$ and note that, by the Bellman equation
(\ref{EQ_BellmanEq}), $V^{\pi_{J}}=\mathbb{K}^{\pi_{J}}R+\gamma\mathbb{K}^{\pi_{J}}V^{\pi_{J}}$.
To ease notation, we define $\mathbb{T}^{\pi_{J}}f=\mathbb{K}^{\pi_{J}}R+\gamma\mathbb{K}^{\pi_{J}}f$
for any real valued function on $\mathcal{S}$. Adding and subtracting
$\mathbb{T}^{\pi_{J}}V^{*}$, we have that 
\begin{align*}
V^{*}-V^{\left(J\right)}=V^{*}-\mathbb{T}^{\pi_{J}}V^{\pi_{J}}= & \left(V^{*}-\mathbb{T}^{\pi_{J}}V^{*}\right)+\left(\mathbb{T}^{\pi_{J}}V^{*}-\mathbb{T}^{\pi_{J}}V^{\pi_{J}}\right)\\
= & \left(V^{*}-\mathbb{T}^{\pi_{J}}V^{*}\right)+\gamma\mathbb{K}^{\pi_{J}}\left(V^{*}-V^{\pi_{J}}\right),
\end{align*}
where the last equality follows from the fact that, for any functions
$f,g$ on $\mathcal{S}$, 
\begin{equation}
\mathbb{T}^{a}f-\mathbb{T}^{a}g=\gamma\mathbb{K}^{a}\left(f-g\right)\label{EQ_TdiffKdif}
\end{equation}
by definition of $\mathbb{T}^{a}$. Since, $V^{(J)}=V^{\pi_{J}}$,
the above implies the following recursion which can be iterated:
\begin{align*}
V^{*}-V^{\left(J\right)}= & \gamma\mathbb{K}^{\pi_{J}}\left(V^{*}-V^{\left(J\right)}\right)+\left(V^{*}-\mathbb{T}^{\pi_{J}}V^{*}\right)\\
= & \sum_{l=0}^{\infty}\gamma^{l}\left(\mathbb{K}^{\pi_{J}}\right)^{l}\left(V^{*}-\mathbb{T}^{\pi_{J}}V^{*}\right).
\end{align*}
Taking the $\left|\cdot\right|_{2,P}$ norm, and using Minkowski inequality,
\begin{align}
\left|V^{*}-V^{\left(J\right)}\right|_{2,P}\leq & \sum_{l=0}^{\infty}\gamma^{l}\left|\left(\mathbb{K}^{\pi_{J}}\right)^{l}\left(V^{*}-\mathbb{T}^{\pi_{J}}V^{*}\right)\right|_{2,P}\nonumber \\
\leq & \frac{\underline{\alpha}^{-1/2}}{\left(1-\gamma\right)}\left|V^{*}-\mathbb{T}^{\pi_{J}}V^{*}\right|_{2,P}\label{EQ_mainErrorBoundTh1}
\end{align}
where the last inequality follows using the fact that $\left(V^{*}-\mathbb{T}^{\pi_{J}}V^{*}\right)\geq0$
and by replacing $\mathbb{K}^{\pi_{J}}$ with $\mathbb{K}=\max_{a}\mathbb{K}^{a}$
and using Lemma \ref{Lemma_transitionStability}. 

We shall bound the quantity in the norm on the r.h.s. of the display.
Note that $V^{*}=\max_{a\in\mathcal{A}}\mathbb{T}^{a}V^{*}$ and let
$\pi_{*}$ be the optimal true, but unknown policy. For any fixed
but arbitrary $s\in\mathcal{S}$, let $a_{*}=\pi_{*}\left(s\right)$
be the maximiser of $\mathbb{T}^{a}V^{*}\left(s\right)$ so that $V^{*}\left(s\right)=\mathbb{T}^{a_{*}}V^{*}\left(s\right)$.
Let $a_{J}=\pi_{J}\left(s\right)$ and note that, by definition, $\hat{Q}^{\left(J\right)}\left(s,a_{J}\right)\geq\hat{Q}^{\left(J\right)}\left(s,a\right)$
for $a\in\mathcal{A}$. Hence, 
\begin{align*}
V^{*}\left(s\right)-\mathbb{T}^{\pi_{J}}V^{*}\left(s\right)= & \mathbb{T}^{a_{*}}V^{*}\left(s\right)-\mathbb{T}^{a_{J}}V^{*}\left(s\right)\\
\leq & \left(\mathbb{T}^{a_{*}}V^{*}\left(s\right)-\hat{Q}^{\left(J\right)}\left(s,a_{*}\right)\right)+\left(\hat{Q}^{\left(J\right)}\left(s,a_{J}\right)-\mathbb{T}^{a_{J}}V^{*}\left(s\right)\right)\\
\leq & 2\max_{a\in\mathcal{A}}\left|\mathbb{T}^{a}V^{*}\left(s\right)-\hat{Q}^{\left(J\right)}\left(s,a\right)\right|.
\end{align*}
The inequality is valid for any $s\in\mathcal{S}$, implying 
\begin{align}
V^{*}-\mathbb{T}^{\pi_{J}}V^{*}\leq & 2\max_{a\in\mathcal{A}}\left|\mathbb{T}^{a}V^{*}-\hat{Q}^{\left(J\right)}\left(\cdot,a\right)\right|\nonumber \\
\leq2 & \max_{a\in\mathcal{A}}\left|\mathbb{T}^{a}V^{*}-\mathbb{T}^{a}\hat{V}^{\left(J-1\right)}\right|+2\max_{a\in\mathcal{A}}\left|\mathbb{T}^{a}\hat{V}^{\left(J-1\right)}-\hat{Q}^{\left(J\right)}\left(\cdot,a\right)\right|\nonumber \\
\leq & 2\gamma\max_{a\in\mathcal{A}}\mathbb{K}^{a}\left|V^{*}-\hat{V}^{\left(J-1\right)}\right|+2\max_{a\in\mathcal{A}}\left|\mathbb{T}^{a}\hat{V}^{\left(J-1\right)}-\hat{Q}^{\left(J\right)}\left(\cdot,a\right)\right|\label{EQ_intermediateRecursionBound}
\end{align}
where in the second inequality we added and subtracted $\mathbb{T}^{a}\hat{V}^{\left(J-1\right)}$
together with the triangle inequality and in the third inequality,
we used (\ref{EQ_TdiffKdif}) together with Jensen inequality. 

We bound the first term on the r.h.s. of the display. By the Bellman
identities, $V^{*}=\max_{a\in\mathcal{A}}\mathbb{T}^{a}V^{*}$ and
$\hat{V}^{\left(j\right)}=\max_{a\in\mathcal{A}}\hat{Q}^{\left(j\right)}\left(\cdot,a\right)$.
Adding and subtracting $\mathbb{T}^{a}\hat{V}^{\left(j-1\right)}$
and using the triangle inequality, 
\begin{align*}
\left|V^{*}-\hat{V}^{\left(j\right)}\right|= & \left|\max_{a\in\mathcal{A}}\mathbb{T}^{a}V^{*}-\max_{a\in\mathcal{A}}\hat{Q}^{\left(j\right)}\left(\cdot,a\right)\right|\leq\max_{a\in\mathcal{A}}\left|\mathbb{T}^{a}V^{*}-\hat{Q}^{\left(j\right)}\left(\cdot,a\right)\right|\\
\leq & \max_{a\in\mathcal{A}}\left|\mathbb{T}^{a}V^{*}-\mathbb{T}^{a}\hat{V}^{\left(j-1\right)}\right|+\max_{a\in\mathcal{A}}\left|\mathbb{T}^{a}\hat{V}^{\left(j-1\right)}-\hat{Q}^{\left(j\right)}\left(\cdot,a\right)\right|.
\end{align*}
From (\ref{EQ_TdiffKdif}) and Jensen inequality, the first term on
the r.h.s. is bounded as follows, 
\[
\max_{a\in\mathcal{A}}\left|\mathbb{T}^{a}V^{*}-\mathbb{T}^{a}\hat{V}^{\left(j-1\right)}\right|=\gamma\max_{a\in\mathcal{A}}\left|\mathbb{K}^{a}\left(V^{*}-\hat{V}^{\left(j-1\right)}\right)\right|\leq\gamma\max_{a\in\mathcal{A}}\mathbb{K}^{a}\left|V^{*}-\hat{V}^{\left(j-1\right)}\right|,
\]
To ease notation, define $d_{j}:=\left|V^{*}-\hat{V}^{\left(j\right)}\right|$
and $\xi_{j}:=\max_{a\in\mathcal{A}}\left|\hat{Q}^{\left(j\right)}\left(\cdot,a\right)-\mathbb{T}^{a}\hat{V}^{\left(j-1\right)}\right|$.
The above two displays imply the following recursive inequality, $d_{j}\leq\gamma\mathbb{K}d_{j-1}+\xi_{j}$,
where $\mathbb{K}$ is as in Lemma \ref{Lemma_transitionStability}
and $\hat{V}^{\left(0\right)}:=0$ so that $d_{0}=\left|V^{*}\right|$.
Iterating, using the monotonicity, positive homogeneity and subadditivity
of $\mathbb{K}$, we find that 
\[
d_{j}\leq\gamma^{j}\left(\mathbb{K}\right)^{j}d_{0}+\sum_{l=0}^{j-1}\gamma^{l}\left(\mathbb{K}\right)^{l}\xi_{j-l};
\]
note that $\mathbb{K}$ is a nonlinear operator due to the maximum.
We shall use this and the related notation in (\ref{EQ_intermediateRecursionBound})
to find that 
\begin{align*}
V^{*}-\mathbb{T}^{\pi_{J}}V^{*} & \leq2\gamma\mathbb{K}d_{J-1}+2\xi_{J}\\
 & \leq2\gamma^{J}\left(\mathbb{K}\right)^{J}d_{0}+2\sum_{l=0}^{J-1}\gamma^{l}\left(\mathbb{K}\right)^{l}\xi_{J-l}.
\end{align*}
Taking the $\left|\cdot\right|_{2,P}$ norm, using Minkowski inequality
and then Lemma \ref{Lemma_transitionStability}, we deduce that 
\[
\left|V^{*}-\mathbb{T}^{\pi_{J}}V^{*}\right|_{2,P}\leq\frac{2\underline{\alpha}^{-1/2}}{1-\gamma}\max_{j\leq J}\left|\xi_{j}\right|_{2,P}+2\underline{\alpha}^{-1/2}\gamma^{J}\left|V^{*}\right|_{2,P}.
\]
Substituting in (\ref{EQ_mainErrorBoundTh1}) we obtain the result.\end{proof} 

\paragraph{Proof of Theorem \ref{Theorem_convergenceAlgo}.}

This follows immediately from Lemmas \ref{Lemma_recursionBound} and \ref{Lemma_estimtionErrorUniform}. The remark on the capping error follows from inspection of the proof of Lemma \ref{Lemma_approximationError}. 

\subsection{Proof of Corollary \ref{Corollary_withMinimumDistance}}

\begin{proof} With $r_{n}$ as in Assumption \ref{Condition_functionClass-1},
choose $B\asymp r_{n}^{-2/v}$. Then $Br_{n}$ and $B^{-(v-2)/2}$
are both of order $r_{n}^{(v-2)/v}$. Substituting in Theorem \ref{Theorem_convergenceAlgo}
yields the result. \end{proof}

\subsection{Proof of Corollary \ref{Corollary_withZeroDistance} and Theorem
\ref{Theorem_MainCorollarywithZeroDistance}}

Under the stated conditions, the approximation and capping terms ($Approx$
and $Cap$) in Theorem \ref{Theorem_convergenceAlgo} are zero.

\subsection{Proof of Corollary \ref{Corollary_withPolynomialApprox}}

First, we bound from below the smallest eigenvalue of $G$ in Assumption
\ref{Condition_functionClass-1}. This will produce a bound on the
operator norm of $G^{-1}$. Let $p$ be the density of the stationary
law of the raw state. By assumption, it is on $[0,1]$ and satisfies
$p\left(\tilde{s}\right)\ge c>0$ almost everywhere, where $c$ is
independent of $L$ and $n$. We first verify the design conditions
of Theorem \ref{Theorem_convergenceAlgo}. Since $G=P\varphi\varphi^{\mathrm{T}},$
for arbitrary conformable vector $x$ of unit Euclidean norm, 
\begin{align*}
x'Gx= & \mathbb{E}\left(\sum_{a'\in\mathcal{A}}\sum_{l=1}^{L+q-1}x_{l,a'}e_{l}\left(\tilde{S}_{t}\right)1_{\left\{ A_{t-1}=a'\right\} }\right)^{2}\\
= & \mathbb{E}\mathbb{E}\left[\sum_{a\in\mathcal{A}}\alpha\left(a|S_{t-1}\right)\left(\sum_{l=1}^{L+q-1}x_{l,a}e_{l}\left(\tilde{S}_{t}\right)\right)^{2}|S_{t-1}\right]
\end{align*}
using the tower law; to avoid notational trivialities we have used
double indexing in $x$. By the Assumptions, the above is greater
than 
\[
\sum_{a\in\mathcal{A}}\underline{\alpha}\mathbb{E}\mathbb{E}\left[\left(\sum_{l=1}^{L+q-1}x_{l,a}e_{l}\left(\tilde{S}_{t}\right)\right)^{2}|S_{t-1}\right]=\sum_{a\in\mathcal{A}}\underline{\alpha}\mathbb{E}\left(\sum_{l=1}^{L+q-1}x_{l,a}e_{l}\left(\tilde{S}_{t}\right)\right)^{2},
\]
where the equality follows again by the tower law. Replacing the expectation
with the integral w.r.t. the density of $P$, and using the lower
bound, we find that the above display is greater than 
\[
\sum_{a\in\mathcal{A}}\underline{\alpha}c\int_{0}^{1}\left(\sum_{l=1}^{L+q-1}x_{l,a}e_{l}\left(\tilde{s}\right)\right)^{2}d\tilde{s}.
\]
An application of a standard result on the norm of B-splines (de Boor,
1976, Theorem 5.2) gives 
\[
\frac{\left|x\right|_{2}^{2}}{LqD_{q}^{2}}\leq\sum_{a\in\mathcal{A}}\int_{0}^{1}\left(\sum_{l=1}^{L+q-1}x_{l,a}e_{l}\left(\tilde{s}\right)\right)^{2}d\tilde{s}
\]
where $D_{q}$ is a constant depending only on $q$. By the above
two displays we deduce that the smallest eigenvalue of $G$ is bounded
below by a constant multiple of $\underline{\alpha}/\left(D_{q}^{2}Lq\right)$.
Then, $\left|G^{-1}\right|_{{\rm op}}\leq D_{q}^{2}Lq/\left(\underline{\alpha}c\right)$.
We now verify the growth condition in Assumption \ref{Condition_functionClass-1}.
Since the B-splines are nonnegative and form a partition of unity,
$\sup_{s\in\mathcal{S}}\left|\varphi\left(s\right)\right|\leq1$.
Therefore, 
\[
\sup_{s\in\mathcal{S}}\varphi\left(s\right)^{\mathrm{T}}G^{-1}\varphi\left(s\right)\lesssim\frac{D_{q}^{2}Lq}{\underline{\alpha}c}\lesssim K.
\]
Next we control the approximation error $Approx$ in Theorem \ref{Theorem_convergenceAlgo}. For fixed $q$, the local spline approximation result of de Boor (1976, Section 6, Lemma 6.1 and equation (6.6)), together with Taylor's theorem, implies that a function $g$ with bounded $q$th derivative $g^{\left(q\right)}$ can be approximated by a spline of order $q$ with error bounded by $C_{q}L^{-q}|g^{\left(q\right)}|_{\infty}$; the constant $C_{q}$ depends only on $q$ and not on $L$. We apply this result to each term on the r.h.s of $\mathbb{T}^{a}f=\mathbb{K}^{a}R+\gamma\mathbb{K}^{a}f$. By assumption $\mathbb{K}^{a}R$ has $q$ uniformly bounded derivatives, so the error rate in the approximation is of order $L^{-q}$. The uniform bound on the derivatives applies to $f/B$ for any $f\in\mathcal{F}_{B}^{\max}$. Hence, the derivatives of $\mathbb{T}^{a}f=\mathbb{K}^{a}R+\gamma\mathbb{K}^{a}f$ are uniformly bounded by a constant multiple of $1+\gamma B$, $f\in\mathcal{F}_{B}^{\max}$.
Hence, the approximation error $Approx$ is of order $L^{-q}$. Moreover,
the choice of $B$ gives $\left|\mathbb{T}^{a}f\right|_{\infty}\le\max_{a\in\mathcal{A}}\left|\mathbb{K}^{a}R\right|_{\infty}+\gamma B\le B$
so that the capping error $Cap$ is zero.

Finally, since $K=(L+q-1)|\mathcal{A}|\asymp L$ and $q$ and $|\mathcal{A}|$
are fixed, using the choice $L\asymp\left(n/\left(\ln n\right)^{2}\right)^{1/\left(2q+1\right)},$
we have that
\[
r_{n}=\sqrt{\frac{K\ln K\ln n}{n}}=O\left(\left[\frac{\left(\ln n\right)^{2}}{n}\right]^{q/(2q+1)}\right).
\]
For the specified choice of $\rho$, together with $|G^{-1}|_{\mathrm{op}}\lesssim L$,
gives 
\[
\rho|G^{-1}|_{\mathrm{op}}\lesssim\rho L=O\left(\left[\frac{\left(\ln n\right)^{2}}{n}\right]^{q/\left(2q+1\right)}\right).
\]
Substitution into Theorem \ref{Theorem_convergenceAlgo} proves the
result.

\subsection{Proof of Corollary \ref{Corollary_withEstimatedTheta}}

When $\theta^{a}$ is estimated, we have that for $j\geq1$, $\hat{Q}^{\left(j\right)}\left(\cdot,a\right)=\Pi_{n,B}\left[R\left(T^{a}\cdot;\hat{\theta}^{a}\left(\cdot\right)\right)+\gamma\hat{V}^{\left(j-1\right)}\left(T^{a}\cdot\right)\right]$
with $\hat{V}^{\left(0\right)}:=0$. Then, Algorithm \ref{Algorithm_Sieve}
applies as it is with $\Gamma_{n}^{a}$ replaced by $\hat{\Gamma}_{n}^{a}$
where for any $f$ on $\mathcal{S}$, $\hat{\Gamma}_{n}^{a}f=R\left(T^{a}\cdot;\hat{\theta}^{a}\left(\cdot\right)\right)+\gamma f\left(T^{a}\cdot\right)$.
That is $\hat{Q}^{\left(j\right)}\left(\cdot,a\right)=\Pi_{n,B}\hat{\Gamma}_{n}^{a}\hat{V}^{\left(j-1\right)}$.
We have the following crucial result. 

\begin{lemma}\label{Lemma_recursionErrorThetaBound}Consider Algorithm
\ref{Algorithm_Sieve} with $\Gamma_{n}^{a}$ replaced by $\hat{\Gamma}_{n}^{a}$.
Under the Assumptions with Assumption \ref{Condition_theta-1} replaced
by Assumption \ref{Condition_thetaEstimated}, 
\[
\max_{j\leq J}\left|\max_{a\in\mathcal{A}}\left|\hat{Q}^{\left(j\right)}\left(\cdot,a\right)-\Pi_{n,B}\Gamma_{n}^{a}\hat{V}^{\left(j-1\right)}\right|\right|_{2,P}=O_{P}\left(\sqrt{\frac{C_{\Theta}}{n}}\right).
\]

\end{lemma}

\begin{proof}Since $\hat{Q}^{\left(j\right)}\left(\cdot,a\right)=\Pi_{n,B}\hat{\Gamma}_{n}^{a}\hat{V}^{\left(j-1\right)}$,
the proof just requires to bound a norm of the difference $\Pi_{n,B}\left[\left(\hat{\Gamma}_{n}^{a}-\Gamma_{n}^{a}\right)\hat{V}^{\left(j-1\right)}\right]$.
By definition of the operators, for any function $f$ on $\mathcal{S}$,
\[
\left(\hat{\Gamma}_{n}^{a}-\Gamma_{n}^{a}\right)f=R\left(T^{a}\cdot;\hat{\theta}^{a}\left(\cdot\right)\right)-R\left(T^{a}\cdot;\theta^{a}\left(\cdot\right)\right).
\]
By these remarks and the contraction property of the capping operator
(Lemma \ref{Lemma_projDiffIneq}), 
\begin{equation}
\left|\hat{Q}^{\left(j\right)}\left(\cdot,a\right)-\Pi_{n,B}\Gamma_{n}^{a}\hat{V}^{\left(j-1\right)}\right|\leq\left|\Pi_{n}\left[R\left(T^{a}\cdot;\hat{\theta}^{a}\left(\cdot\right)\right)-R\left(T^{a}\cdot;\theta^{a}\left(\cdot\right)\right)\right]\right|\label{EQ_QIterDiffThetaEst}
\end{equation}
where the inequality holds pointwise. The objective is to bound this
in $\left|\cdot\right|_{2,P}$ norm. Since the r.h.s. does not depend
on $j$, the bound will then hold uniformly in $j\leq J$. For reasons
that will become apparent momentarily, we first bound the difference
\[
D_{n}^{2}:=\max_{a\in\mathcal{A}}P_{n}\left|R\left(T^{a}\cdot;\hat{\theta}^{a}\left(\cdot\right)\right)-R\left(T^{a}\cdot;\theta^{a}\left(\cdot\right)\right)\right|^{2}.
\]
From Assumption \ref{Condition_thetaEstimated}, 
\[
D_{n}\leq\max_{a\in\mathcal{A}}\left|\hat{\theta}^{a}-\theta^{a}\right|_{\infty}\left|\Delta\left(T^{a}\cdot\right)\right|_{2,P_{n}}=O_{P}\left(\sqrt{C_{\Theta}/n}\right).
\]
Using the fact that, by Jensen inequality, 
\[
\mathbb{E}\max_{a\in\mathcal{A}}\left|\Delta\left(T^{a}\cdot\right)\right|_{2,P_{n}}\leq\left|\mathcal{A}\right|^{1/2}\max_{a\in\mathcal{A}}\left(\mathbb{E}\left|\Delta\left(T^{a}\cdot\right)\right|_{2,P_{n}}^{2}\right)^{1/2}=\left|\mathcal{A}\right|^{1/2}\max_{a\in\mathcal{A}}\left|\Delta\left(T^{a}\cdot\right)\right|_{2,P}
\]
which is finite. Now, by Lemma \ref{Lemma_projPopNorm2EmpNorm}, on
the event $\mathcal{E}$, as defined there, for any real valued function
$f$ on $\mathcal{S}$, 
\[
\left|\Pi_{n}f\right|_{2,P}\le2\left|P_{n}\psi f\right|_{2}\leq2\left|P_{n}\psi\psi^{\mathrm{T}}\right|_{{\rm op}}^{1/2}\left|f\right|_{2,P_{n}}=O_{P}\left(\left|f\right|_{2,P_{n}}\right)
\]
because, by Lemma \ref{Lemma_normEmpiricalBound}, $\left|P_{n}\psi\psi^{\mathrm{T}}\right|_{{\rm op}}^{1/2}=O_{P}\left(1\right)$.
Applying the above inequality to the $\left|\cdot\right|_{2,P}$ norm
of the maximum over $a\in\mathcal{A}$ of (\ref{EQ_QIterDiffThetaEst}),
we deduce that, on the event $\mathcal{E}$, 
\[
\left|\max_{a\in\mathcal{A}}\left|\Pi_{n}\left[R\left(T^{a}\cdot;\hat{\theta}^{a}\left(\cdot\right)\right)-R\left(T^{a}\cdot;\theta^{a}\left(\cdot\right)\right)\right]\right|\right|_{2,P}\leq2\left|\mathcal{A}\right|^{1/2}\left|P_{n}\psi\psi^{\mathrm{T}}\right|_{{\rm op}}^{1/2}D_{n}=O_{P}\left(\sqrt{C_{\Theta}/n}\right),
\]
where the factor $\left|\mathcal{A}\right|^{1/2}$ comes from taking
the maximum inside the $\left|\cdot\right|_{2,P}$ norm. Now, by Lemma
\ref{Lemma_normEmpiricalBound}, the event $\mathcal{E}$ has probability
going to one. Hence, the result of the lemma follows.\end{proof}

\paragraph{Proof of Corollary \ref{Corollary_withEstimatedTheta}.}

Theorem \ref{Theorem_convergenceAlgo} follows from an application
of Lemma \ref{Lemma_recursionBound}. Inspection of the proof shows
that the result does not depend on whether we use $\Gamma_{n}^{a}$
or $\hat{\Gamma}_{n}^{a}$ in the definition of $\hat{Q}^{\left(j\right)}$.
However, to apply Lemma \ref{Lemma_recursionBound} we need an estimate
of 
\[
\max_{j\le J}\left|\max_{a\in\mathcal{A}}\left|\hat{Q}^{\left(j\right)}\left(\cdot,a\right)-\mathbb{T}^{a}\hat{V}^{\left(j-1\right)}\right|\right|_{2,P},
\]
which we derive here. Adding and subtracting $\Pi_{n,B}\Gamma_{n}^{a}\hat{V}^{\left(j-1\right)}$,
we find that the above display is bounded above by 
\[
\max_{j\le J}\left|\max_{a\in\mathcal{A}}\left|\hat{Q}^{\left(j\right)}\left(\cdot,a\right)-\Pi_{n,B}\Gamma_{n}^{a}\hat{V}^{\left(j-1\right)}\right|\right|_{2,P}+\max_{j\le J}\left|\max_{a\in\mathcal{A}}\left|\Pi_{n,B}\Gamma_{n}^{a}\hat{V}^{\left(j-1\right)}-\mathbb{T}^{a}\hat{V}^{\left(j-1\right)}\right|\right|_{2,P}.
\]
The first term is $O_{P}\left(\sqrt{C_{\Theta}/n}\right)$ by Lemma
\ref{Lemma_recursionErrorThetaBound}. The second term is bounded
by Lemma \ref{Lemma_estimtionErrorUniform}. Note that even when $\theta^{a}$
is estimated, $\hat{V}^{\left(j-1\right)}$ is still an element in
$\mathcal{F}_{B}^{\max}$. Then, the corollary follows.

\section{Additional Details on Empirical Analysis}\label{Section_empiricalsupp}
This section provides additional details on the empirical analysis of factor-rotation in the U.S. equity market presented in Section \ref{Section_empirical}.

\subsection{State Variables}\label{Section_statesupp}
Recall that, in Section \ref{Section_empirical}, the state variables include 12-month factor trends (Trend), market cycle indicators (Cycle) classified based on momentum turning points following Goulding et al. (2023), and macroeconomic variables (Macro). The macroeconomic information set consists of the following variables:

\begin{enumerate}
    \item $\mathrm{FEDFUNDS}_t$: the federal funds rate, which measures the short-term interest rate.
    \item $\mathrm{TermSpread}_t = \mathrm{GS10}_t-\mathrm{GS1}_t$: the term spread, defined as the difference between the ten-year and one-year Treasury yields.
    \item $\mathrm{CreditSpread}_t = \mathrm{BAA}_t-\mathrm{AAA}_t$: the credit spread, defined as the difference between Moody's BAA and AAA corporate bond yields.
    \item $\mathrm{CoreInflation}_t = 100[\log(\mathrm{CoreCPI}_t)-\log(\mathrm{CoreCPI}_{t-12})]$: the 12-month core inflation rate, where $\mathrm{CoreCPI}_t$ is measured by $\mathrm{CPILFENS}_t$, which excludes food and energy.
    \item $\mathrm{FoodEnergy}_t =
    100[\log(\mathrm{HeadlineCPI}_t)-\log(\mathrm{HeadlineCPI}_{t-12})]
    -
    100[\log(\mathrm{CoreCPI}_t)-\log(\mathrm{CoreCPI}_{t-12})]$:
    the food and energy inflation proxy, defined as the difference between 12-month headline and core inflation.
\end{enumerate}

Because the CPI for month $t$ is usually released during month $t+1$, the inflation variables are lagged by one month when used as state variables.
Except for the binary market cycle indicators, we scale factor trends and macroeconomic state variables to improve numerical stability. We scale each variable by a prespecified constant and then clip the scaled value to the interval $[-1,1]$. The scaling constants are 0.1, 20, 10, 10, 15, and 5 for factor trends, $\mathrm{FEDFUNDS}_t$, $\mathrm{TermSpread}_t$, $\mathrm{CreditSpread}_t$, $\mathrm{CoreInflation}_t$, and $\mathrm{FoodEnergy}_t$, respectively. Factor trends and rewards are expressed in decimals, so that 0.1 corresponds to 10\%, whereas the other variables are expressed in percentage points. The scaling constants were chosen a priori to represent economically meaningful bounds, with clipping limiting the influence of extreme observations. We repeated the estimation without clipping and our conclusions did not change.

\subsection{Estimation Details}\label{Section_estimationDetailsSupp}
The basis functions $\varphi_i$ are quadratic polynomials of each raw state variable, except for dummy variables. For example, for the past 12-month average momentum return, we include both its level and its square. We apply the same transformation to all raw state variables.
The model for the $Q$-function is additive as in (\ref{EQ_linearSpecification}), after mapping the relevant raw states into quadratic polynomials. However, we include an intercept. For this reason, in each function $f_a(({\tilde s},a'))$ in (\ref{EQ_linearSpecification}) we remove the dummy variable corresponding to $\{a'=a\}$ to avoid multicollinearity.
For ridge estimation, we first standardise the variables into z-scores using only observations available at the time of the decision. This ensures that the ridge penalty is applied on a common scale across variables. The ridge penalty for the intercept is set to 0. The estimated coefficients are then transformed back to the original scale when making predictions. To avoid sample dependence, we work with the ridge penalty parameter $\rho/n$. This parameter is selected by generalised cross-validation over a grid of 31 logarithmically spaced values between 0.1 and 2.5.
The estimated $Q$-values are truncated to the interval $[-0.5,0.5]$, i.e., $B=0.5$. From the definition of the $Q$-function in (\ref{EQ_QRecursion}), assuming a constant expected reward equal to $\mu$, we have $Q\simeq \mu/(1-\gamma)$. With $\gamma=0.9$, clipping $Q$ at $B=0.5$ therefore corresponds, under this approximation, to clipping the implied monthly expected reward at $(1-\gamma)B=0.05$, i.e. 5\%, in absolute value. Finally, to reduce the dependence on the simulated actions $A_t$, we use (\ref{EQ_averageQ}) with $N_A=10$. 

Recall that we use an expanding-window estimation scheme with an initial sample of 120 months, using the complete observations available after constructing the state variables. The out-of-sample evaluation therefore begins in July 1973 and contains 629 monthly decisions, corresponding to factor returns realised from August 1973 through December 2025.


\subsection{Supplementary Results for Section \ref{Section_empirical}}
\begin{figure}[htbp]
\caption{
Cumulative Out-of-Sample Rewards. The figure plots the cumulative monthly rewards, in percent, of the RL, Greedy, and four fixed-factor policies. Results are reported when the states are Trend $+$ Cycle, with $\eta=0$ and a switching cost of 20, 80, and 120 bps.}
  \label{Figure_cumulative_rewards_eta0}
  \centering 
        \includegraphics[width=\textwidth]{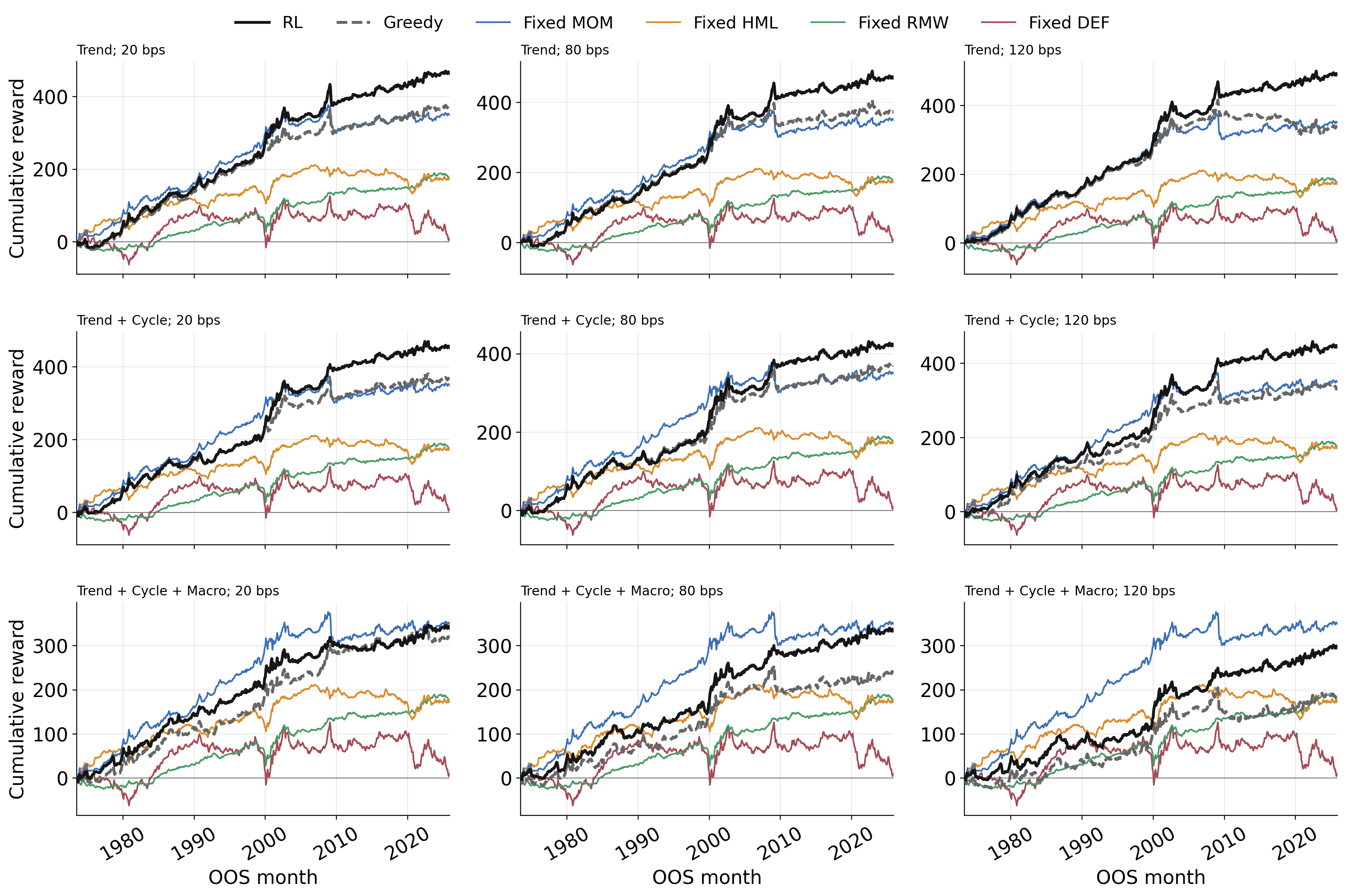} 
\end{figure}

\begin{figure}[htbp]
\caption{
Cumulative Out-of-Sample Rewards. The figure plots the cumulative monthly rewards, in percent, of the RL, Greedy, and four fixed-factor policies. Results are reported when the states are Trend $+$ Cycle, with $\eta=3$ and a switching cost of 20, 80, and 120 bps.}
  \label{Figure_cumulative_rewards_eta3}
  \centering 
        \includegraphics[width=\textwidth]{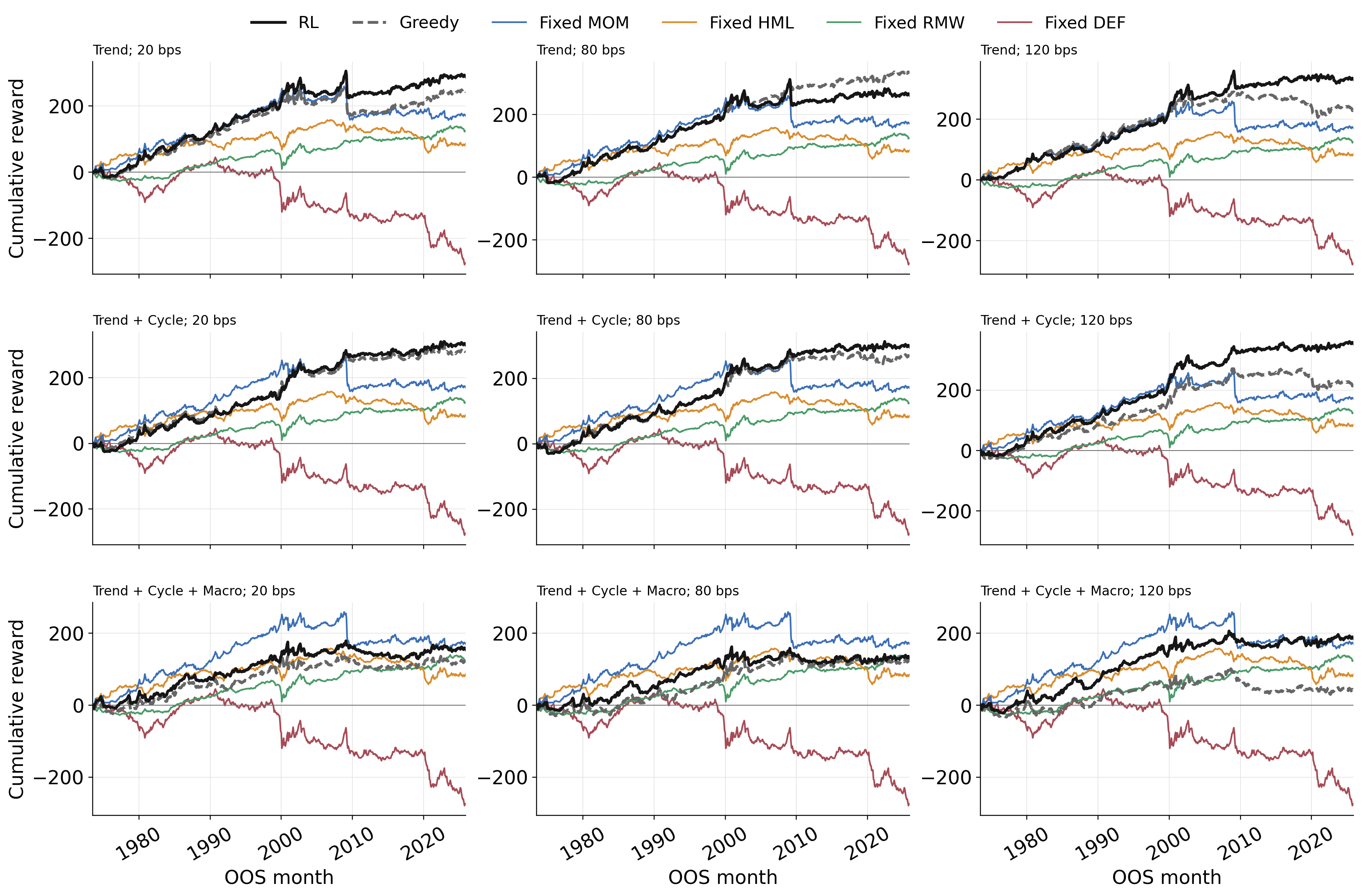} 
\end{figure}

\begin{figure}[htbp]
\caption{
State Dependent Factor Rotations. Results are reported when the states are Trend $+$ Cycle with $\eta=0$. Panels (a) (c) (e) display the percentages of out-of-sample months in which the RL policy selects among the MOM, HML, RMW, and DEF strategies, conditional on the market states classified by the momentum turning points, for costs of 120, 80, and 20 bps, respectively. The corresponding right sub-plots show the destination factor conditional on exiting momentum across different market states. $N$ denotes the number of out-of-sample months in each market state, whereas $n$ denotes the number of months in which the policy exits MOM in each market state. Percentages may not sum to 100 due to rounding. The bottom panel shows the factor selected by the RL policy over the out-of-sample period.}
  \label{Figure_factor_rotation_eta0_bps120}
  \centering
    \begin{subfigure}{\textwidth}
        \centering
        \includegraphics[width=\textwidth]{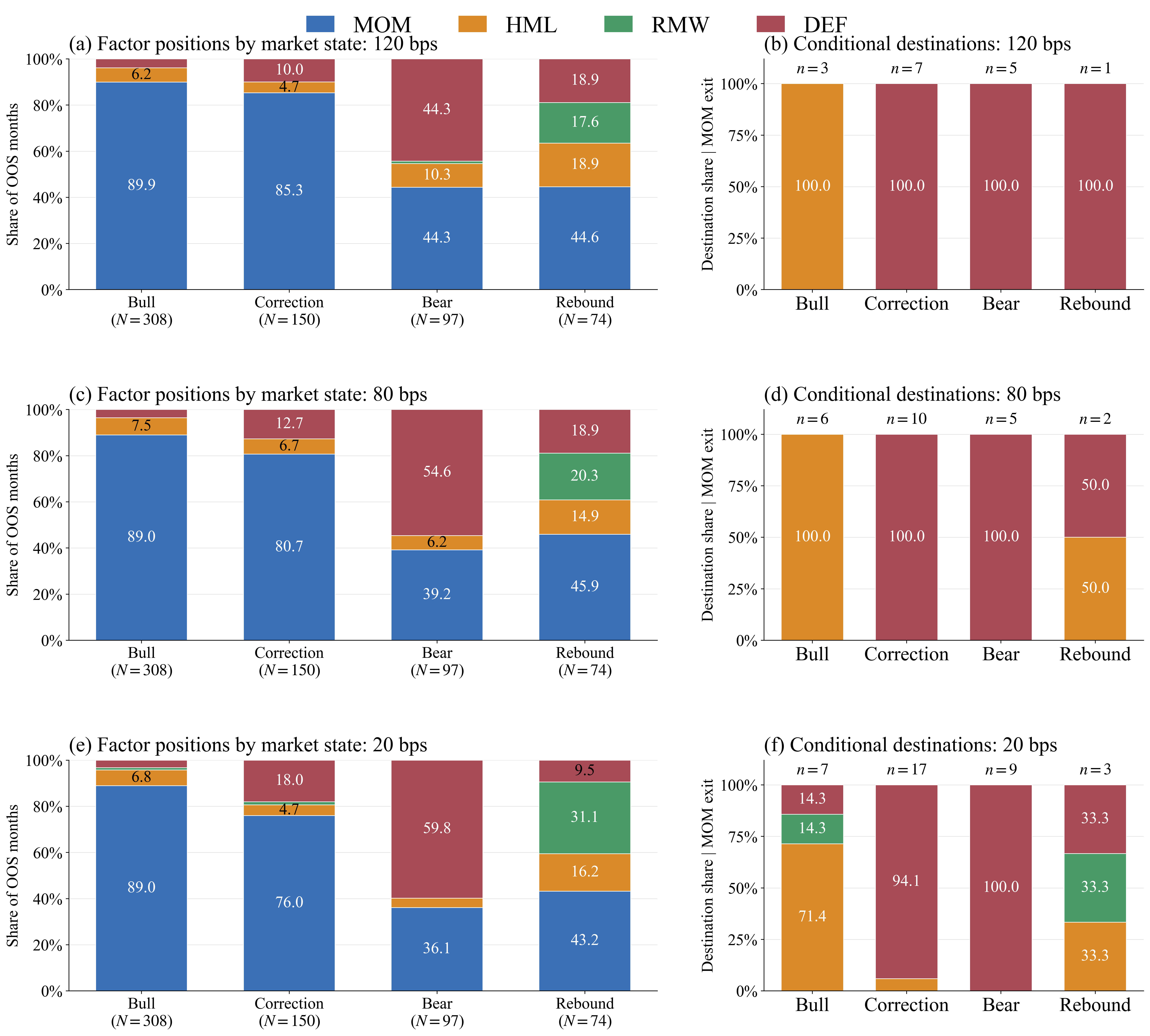} 
        \includegraphics[width=\textwidth]{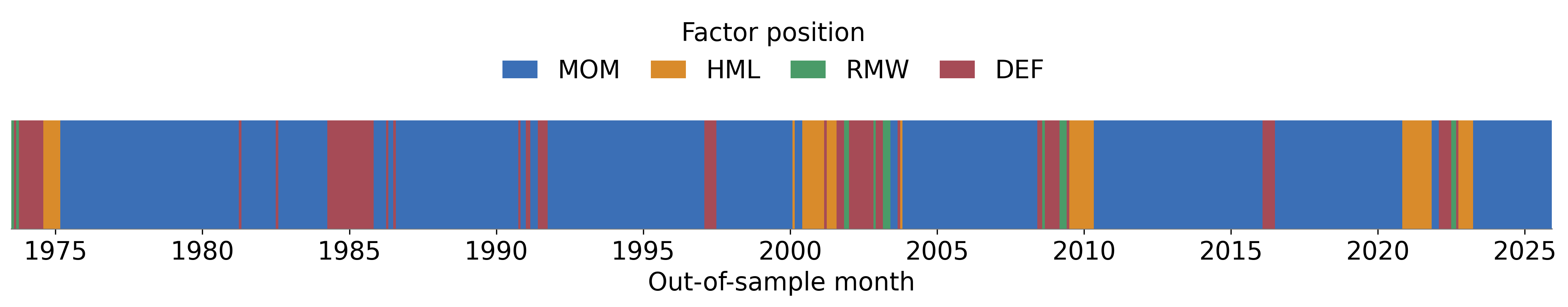} 
    \end{subfigure}
\end{figure}

\begin{figure}[htbp]
\caption{State Dependent Factor Rotations. Results are reported for the Trend $+$ Cycle state specification with $\eta=3$. Panels (a) and (b) correspond to switching costs of 80 and 20 bps, respectively. Each panel reports the percentages of out-of-sample months in which the RL policy selects the MOM, HML, RMW, and DEF strategies, the destination factor conditional on exiting MOM across market states, and the factor selected by the RL policy over the out-of-sample period. For the results with a cost of 120 bps and $\eta=3$, we refer to Figure \ref{Figure_factor_rotation_eta3_120bps} in the main text.
}
  \label{Figure_factor_rotation_eta3_80_20bps} 
  \centering
    \begin{subfigure}{\textwidth}
        \centering
        \includegraphics[width=\textwidth]{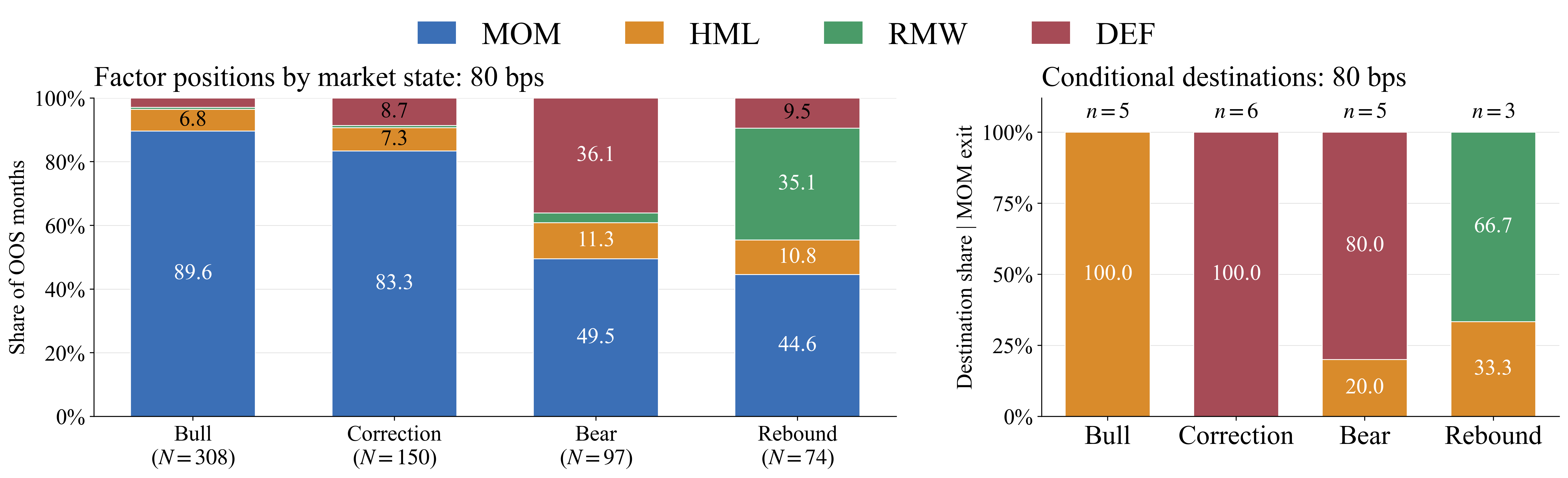} \\
        \includegraphics[width=\textwidth]{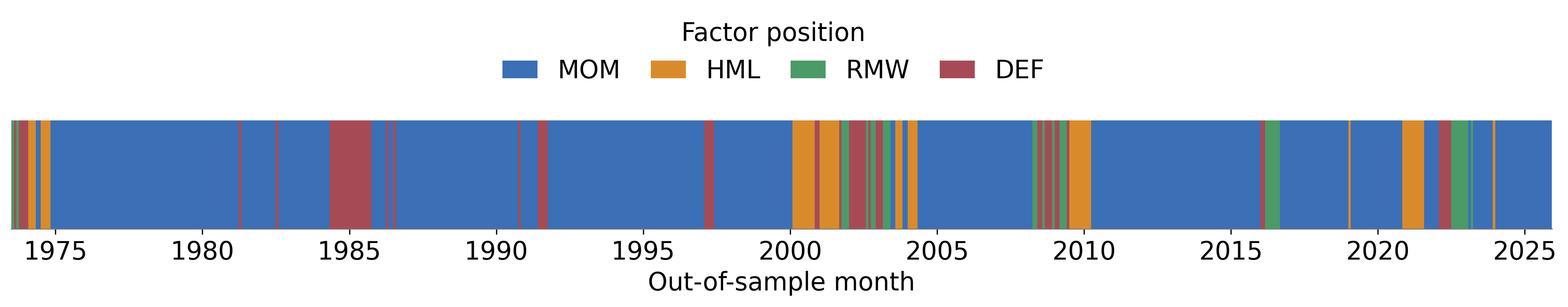}
        \caption{Cost = 80 bps} 
    \end{subfigure}
    \begin{subfigure}{\textwidth}
        \centering
        \includegraphics[width=\textwidth]{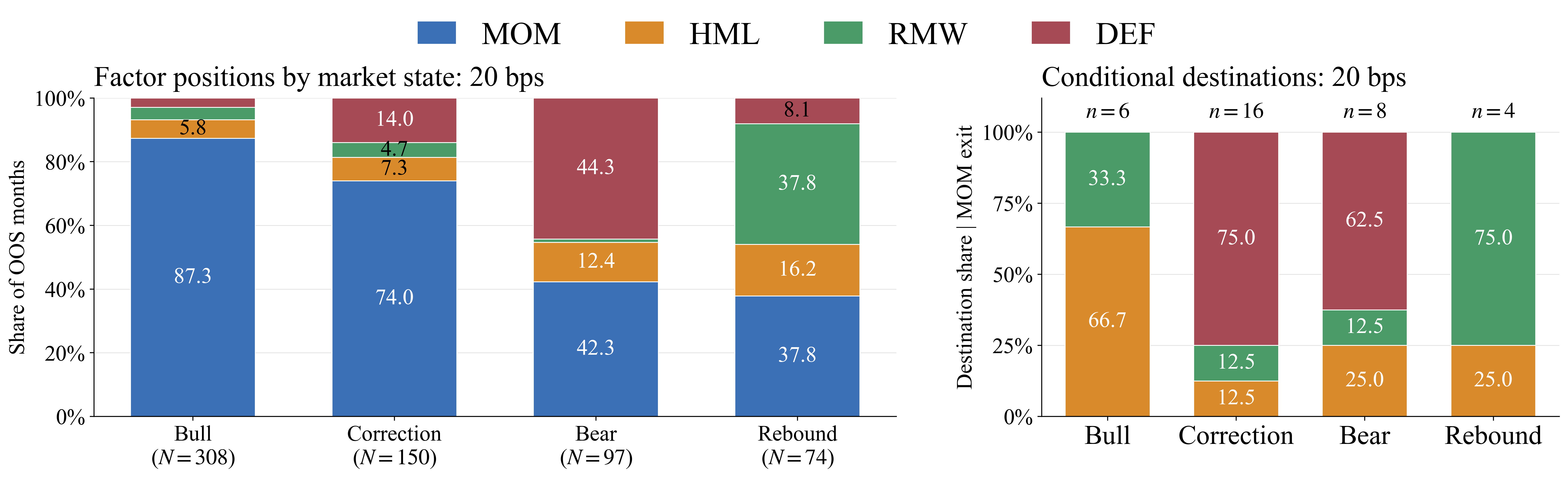}
        \\
        \includegraphics[width=\textwidth]{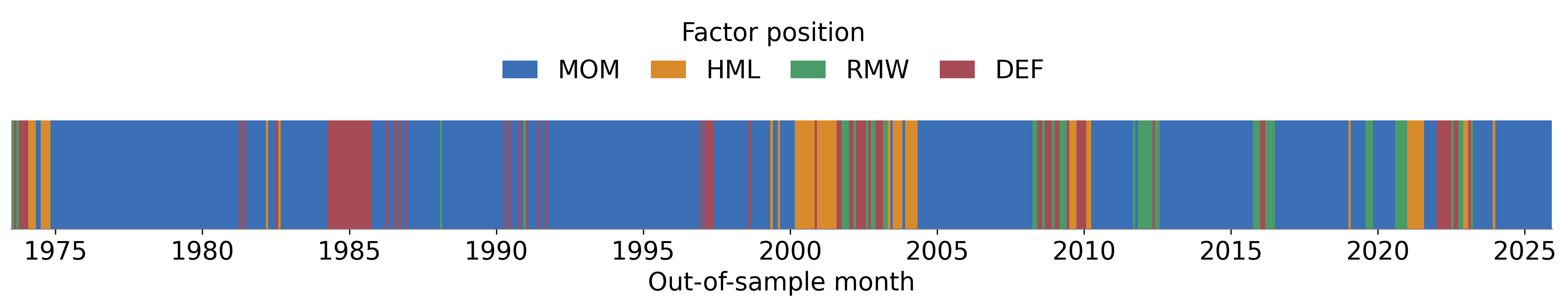}
        \caption{Cost = 20 bps} 
    \end{subfigure}
\end{figure}

In the following, we present the full set of results for the cumulative out-of-sample rewards (Figures \ref{Figure_cumulative_rewards_eta0} and \ref{Figure_cumulative_rewards_eta3}) and the state-dependent factor rotations (Figures \ref{Figure_factor_rotation_eta3_80_20bps} and \ref{Figure_factor_rotation_eta0_bps120}). Figures \ref{Figure_cumulative_rewards_eta0} and \ref{Figure_cumulative_rewards_eta3} present the cumulative out-of-sample rewards for $\eta=0$ and $\eta=3$, respectively. When $\eta=0$, the reward reduces to the factor net return. The performance of these policies is summarised in Table \ref{tab:state_information_performance} of the main text. 
Higher cost leads to stronger regularisation, improving the relative out-of-sample performance of the RL policy. The overall evolution of the reward curves and the relative performance across policies remain broadly similar. When $\eta=3$, the dynamic policies generate smoother reward paths than under $\eta=0$, consistent with the quadratic penalty favouring less volatile factor exposures.

Figure \ref{Figure_factor_rotation_eta0_bps120} presents the distribution of factor rotations across market states when $\eta=0$. Lowering regularisation costs from 120 bps to 20 bps increases the number of rotations and reduces the share of MOM selections. This, in turn, increases the selection of DEF and HML during bear markets and RMW during bull and rebound markets. Compared with the results for $\eta=3$ shown in Figure \ref{Figure_factor_rotation_eta3_120bps} in the main text, the RL policy rotates mainly from MOM to DEF during stressed market conditions. The lower panel shows the evolution of factor rotations at a cost of 120 bps, exhibiting similar clustering patterns in factor selection.

Figure \ref{Figure_factor_rotation_eta3_80_20bps} supplements Figure \ref{Figure_factor_rotation_eta3_120bps} of the main text by showing the factor rotation distribution under $\eta=3$, when costs are 80 and 20 bps, respectively. The main findings remain unchanged, except that at 20 bps, RMW also emerges as a destination following exits from MOM across all market states. 

\subsection{Ridge Penalty and Degrees of Freedom}

The goal of this section is twofold. First, we assess the performance of generalised cross-validation (GCV) relative to a set of reasonable fixed values for the ridge penalty. For GCV, we selected $\rho/n$ from a grid of 31 logarithmically spaced values ranging from \(0.1\) to \(2.5\). 
Second, we highlight the trade-off between estimation and approximation error that arises in our setting. To facilitate interpretation and comparability across specifications, rather than fixing the ridge penalty directly, we fix the effective degrees of freedom, allowing the corresponding ridge penalty to vary with the sample (Hastie et al., 2009). We consider values in $\{8,10,12\}$. A smaller number of effective degrees of freedom corresponds to stronger regularisation. 

The three specifications in our analysis contain 4 (Trend), 7 (Trend+Cycle), and 12 (Trend+Cycle+Macro) raw state variables. The Cycle variables are dummies. As explained in Section \ref{Section_estimationDetailsSupp}, the $Q$-function is modelled using an additive quadratic polynomial in state variables except for dummy variables. In addition, three dummy variables are included to capture the costs (see details in Section \ref{Section_estimationDetailsSupp}). Hence, including an intercept, we need to estimate 12 parameters under the Trend specification (1 intercept+3 dummies for cost+8 parameters for factor trends), 15 parameters under Trend+Cycle (1 intercept+3 dummies for cost+3 dummies for Cycle+8 parameters for factor trends) and 25 parameters under Trend+Cycle+Macro (1 intercept+3 dummies for cost+3 dummies for Cycle+8 for factor trends+10 for macro variables). Thus, for the Trend specification, an effective degrees of freedom of 12 leaves the model essentially unpenalised, whereas regularisation becomes progressively more important for richer specifications.

Table \ref{table-reward_ridge_check} reports the monthly rewards of the dynamic policies under different choices of the ridge penalty for $\eta=0$ and $\eta=3$, respectively. The GCV results have been reported in Table \ref{tab:state_information_performance} of Section \ref{Section_empirical}, which is reproduced here for convenient reference. 
For the RL policy, GCV delivers performance comparable to, and in some cases better than, that obtained by selecting ex post the best of the three target degrees-of-freedom values. For the Greedy policy, GCV is less often the best-performing specification, and the alternative degrees-of-freedom targets yield mixed results. Nonetheless, these differences do not alter our main finding that the RL policy outperforms the Greedy policy when the two approaches are compared under the same fixed ridge specification. We thus conclude that GCV on the estimation sample is a suitable criterion for our problem.

We also observe the standard benefit of regularisation for the larger Trend+Cycle+Macro specification, where the richer approximation comes at the cost of greater estimation error. This pattern, however, is partly confounded by transaction costs. Lower effective degrees of freedom regularise the estimated $Q$-functions. On the other hand, transaction costs tend to smooth differences in $Q$-functions across actions by discouraging switching between strategies. Both mechanisms can therefore stabilise the estimated policy.

\begin{table}[htbp]
\centering
\caption{Monthly Reward across Ridge Specifications. GCV selects the ridge penalty by generalised cross-validation, and the remaining columns fix the effective degrees of freedom. Bold values indicate the best-performing specification under the RL and Greedy policies.
}
\label{table-reward_ridge_check}
\begin{adjustbox}{max width=\linewidth}
\renewcommand{\arraystretch}{1.15}
\begin{tabular}{lccccccccc}
\toprule
& & \multicolumn{4}{c}{RL} & \multicolumn{4}{c}{Greedy} \\
\cmidrule(lr){3-6} \cmidrule(lr){7-10}
\multicolumn{10}{l}{\textit{Panel A: $\eta=0$}} \\
State variables & Cost (bps) & GCV & $df=8$ & $df=10$ & $df=12$ & GCV & $df=8$ & $df=10$ & $df=12$ \\
\midrule
\multirow{3}{*}{Trend} & 20 & \bf{0.738} & 0.693 & 0.675 & 0.607 & 0.588 & \bf{0.657} & 0.644 & 0.572 \\
 & 80 & \bf{0.746} & 0.764 & 0.699 & 0.651 & 0.588 & \bf{0.749} & 0.665 & 0.650 \\
 & 120 & \bf{0.777} & 0.685 & 0.759 & 0.720 & 0.533 & \bf{0.620} & 0.574 & 0.603 \\\cmidrule(lr){2-10}
\addlinespace[2pt]
\multirow{3}{*}{Trend + Cycle} & 20 & \bf{0.721} & 0.660 & 0.638 & 0.634 & 0.580 & 0.621 & \bf{0.637} & 0.606 \\
 & 80 & 0.669 & \bf{0.739} & 0.700 & 0.660 & 0.585 & 0.661 & \bf{0.678} & 0.701 \\
 & 120 &  0.706  & 0.636 & \bf{0.718} & 0.716 & 0.529 & \bf{0.701} & 0.647 & 0.563 \\\cmidrule(lr){2-10}
\addlinespace[2pt]
\multirow{3}{*}{Trend + Cycle + Macro} & 20 & 0.541 & 0.567 & \bf{0.571} & 0.448 & 0.504 & \bf{0.584} & 0.472 & 0.427 \\
 & 80 & 0.531 & 0.534 & 0.498 & \bf{0.568} & 0.377 & 0.452 & 0.562 & \bf{0.563} \\
 & 120 & 0.470 & \bf{0.528} & 0.499 & 0.477 & 0.285 & 0.368 & 0.448 & \bf{0.450} \\
\midrule
\multicolumn{10}{l}{\textit{Panel B: $\eta=3$}} \\
\multirow{3}{*}{Trend} & 20 & 0.460 & \bf{0.475} & 0.461 & 0.399 & 0.384 & 0.482 & \bf{0.484} & 0.385 \\
 & 80 & 0.417 & 0.416 & \bf{0.425} & 0.407 & \bf{0.527} & 0.514 & 0.494 & 0.386 \\
 & 120 & 0.527 & 0.396 & \bf{0.453} & \bf{0.477} & 0.366 & 0.322 & 0.315 & 0.195 \\\cmidrule(lr){2-10}
\addlinespace[2pt]
\multirow{3}{*}{Trend + Cycle} & 20 & 0.477 & \bf{0.502} & 0.458 & 0.423 & 0.441 & \bf{0.470} & 0.447 & 0.437 \\
 & 80 & \bf{0.472} & 0.472 & 0.460 & 0.391 & 0.422 & 0.366 & \bf{0.429} & 0.413 \\
 & 120 & \bf{0.561} & 0.455 & 0.479 & 0.500 & 0.342 & \bf{0.369} & 0.361 & 0.339 \\\cmidrule(lr){2-10}
\addlinespace[2pt]
\multirow{3}{*}{Trend + Cycle + Macro} & 20 & 0.247 & \bf{0.295} & 0.269 & 0.204 & 0.189 & 0.260 & \bf{0.303} & 0.201 \\
 & 80 & 0.210 & \bf{0.237} & 0.234 & 0.255 & 0.191 & 0.230 & 0.268 & \bf{0.301} \\
 & 120 & 0.295 & 0.254 & \bf{0.317} & 0.309 & 0.064 & 0.194 & \bf{0.207} & 0.206 \\
\bottomrule
\end{tabular}
\end{adjustbox}
\end{table}

\clearpage


\section{Finite Sample Analysis via Simulations} \label{Section_simulations}

In this section, we conduct a Monte Carlo simulation to examine the finite-sample behaviour of the proposed algorithm and assess whether its performance is consistent with the asymptotic results. We investigate whether the estimated RL policy approaches the true optimal policy as the training sample size increases and how its finite-sample performance is affected by the inclusion of irrelevant state variables.

\subsection{Simulation Design}\label{Section_SimulationsDGP}

We use a data-generating process (DGP) that is tractable and allows an oracle that knows the DGP to derive the optimal policy.
Let $X_t\in\{1,2,3,4\}$ follow a four-state Markov chain with transition matrix
\begin{equation}
\mathbf{K}=
\begin{pmatrix}
0.85 & 0.05 & 0.10 & 0 \\
0.05 & 0.85 & 0 & 0.10 \\
0.90 & 0 & 0.10 & 0 \\
0 & 0.90 & 0 & 0.10
\end{pmatrix},\label{EQ_transitionMatrixSimulations}
\end{equation}
where $\mathbf{K}_{ij}=\Pr(X_{t+1}=j\mid X_t=i)$. States 1 and 2 are relatively persistent, with probability $0.85$ of remaining in the same state from one period to the next. State 1 can occasionally transition to state 3, while state 2 can occasionally transition to state 4. States 3 and 4 are comparatively short-lived. Conditional on being in state 3 or 4, the chain returns to state 1 or 2, respectively, with probability $0.90$ in the next period. Finally, with probability $0.05$ the chain can move between states 1 and 2.

We assume $a \in{\mathcal A}=\{1,2,3,4\}$ so that we have four strategies. Each strategy $a$ generates returns according to
\begin{equation*}
    \mathrm{Ret}_{t+1}^{a} = \beta 1_{\{X_{t+1}=a\}} + \varepsilon_{t+1}^{a}.
\end{equation*} 
The error terms $\varepsilon_{t+1}^a$ are i.i.d. over time, strictly exogenous, and follow a normal distribution, $N(0,\sigma_\varepsilon^2)$, for each strategy, where $\sigma_\varepsilon=0.04$, corresponding to a monthly volatility of 4\%. The pairwise correlation of the innovations across strategies is set to $0.3$. Thus, strategy $a$ has conditional expected return $\beta$ when $X_{t+1}=a$ and zero otherwise. We set $\beta=0.01$, so that, conditional on $X_{t+1}=a$ strategy $a$ has expected monthly return of 1\%.  

\paragraph{The rewards and optimal oracle policy.}
We consider rewards as in Example \ref{Example_reward} of Section \ref{Section_problemSetup} and discount factor $\gamma=0.9$. To evaluate how closely the estimated policies recover the optimal state-dependent decision rule, we use the oracle policy as a benchmark. The simple DGP is deliberately designed so that the oracle benchmark can be computed from the known transition matrix and return distribution. Unlike the estimated policies, the oracle is not subject to estimation error. At time $t$, it observes the current state $X_t$, but not the next-period state $X_{t+1}$. The oracle policy selects Strategy 1 at time $t$ when $X_t\in \{1,3\}$ and Strategy 2 when $X_t\in \{2,4\}$. Section \ref{Section_simulatonsOptimalPolicy} provides the detailed derivation. For the parameter values considered here, this oracle policy is invariant to risk aversion over a wide range, including $\eta\in[0,164]$. Hence, we set $\eta=0$.

\paragraph{The estimation.}
For the estimation, the raw state at time $t$ is  
\begin{equation}
\tilde{S}_t= \left(1_{\{X_{t}=1\}},1_{\{X_{t}=2\}},1_{\{X_{t}=3\}},1_{\{X_{t}=4\}}\right)\label{EQ_rawStateTrueSimulation} 
\end{equation}
By the Markov chain property, the raw state $\tilde{S}_{t}$ is a signal for the next period return because $X_{t+1}$ depends on it. In the estimation, we omit the first state dummy and include an intercept to avoid perfect multicollinearity. 
We also augment the raw state with noisy variables that follow an AR(1) with autoregressive coefficient $0.5$ and innovations that are $N(0,1-0.5^2)$. This implies that the noisy variables are unconditionally standard normal. These variables are independent of $X_t$ and do not enter the return-generating process. We consider specifications with 2, 4 and 8 such noisy variables and let their innovations have same correlations equal to $0.3$. Including these noise variables allows us to assess the performance of our algorithm in the presence of irrelevant predictors, a practical challenge that researchers may face when implementing the procedure. Moreover, these noisy states are unbounded, allowing us to assess robustness of the procedure.  In the estimation, we use a switching cost of 120 bps, as in the empirical section. The scaled ridge parameter $\rho/n$ is chosen using GCV from a grid of 31 logarithmically spaced values ranging from  $10^{-6}$ to $2.5$. 

The simulation is based on 500 replications. The training sample size is set to 240, 600, or 1,200 observations, and the out-of-sample performance is evaluated over 500 observations.

\subsection{Simulation Results and Discussion\protect\label{Section_simulationResults}}
 
\begin{figure}[htbp]
\caption{Boxplot of Monte Carlo Performance. For each information set, the figure reports the distributions of the RL and Greedy strategies across 500 replications for training-sample sizes of $n_{\mathrm{est}}=240$, 600 and 1200. 
Each replication contains 500 out-of-sample observations. Boxes indicate the interquartile range. The oracle and strategy 1 are presented as benchmarks. Strategy 2 is omitted because its performance is similar to that of Strategy 1, while Strategies 3 and 4 are omitted because of their weaker performance.}
  \label{Figure_boxplot_eta0}
  \centering  
        \includegraphics[width=\textwidth]{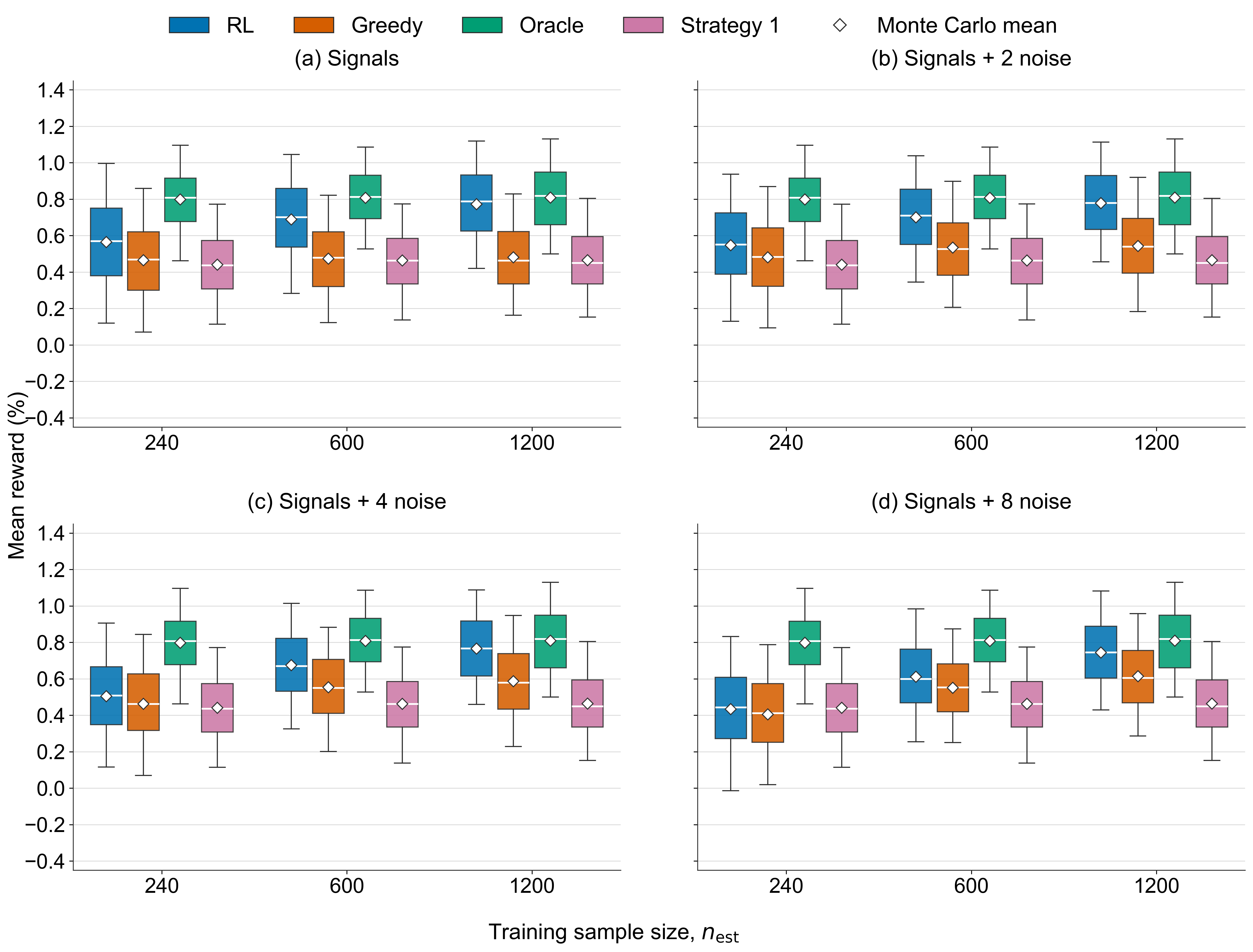}  
\end{figure}

Figure \ref{Figure_boxplot_eta0} summarises our results. 
The results are in line with the asymptotic analysis in Section \ref{Section_asymptotics}. In particular, the reward increases with the sample size. For the largest estimation sample size $n_{\rm est}=1200$, the results are close to the oracle policy, which is the true optimal policy under the known data-generating process. We apply the same oracle policy to each simulated out-of-sample path. Its realised performance therefore varies across replications because states and returns are random. 
The simulation also highlights the benefit of accounting for the continuation value. RL achieves considerably higher rewards than Greedy. It is also interesting to note that choosing strategy 1 would produce suboptimal results, highlighting the importance of using a dynamic approach in strategy selection.
A challenge in dynamic strategy selection is the identification of the relevant state variables. This is reflected in our results. The inclusion of irrelevant state variables deteriorates performance in small samples. This pattern is consistent with the trade-off between information and estimation error discussed in Section \ref{Section_empirical}. As expected from the theory, the deterioration becomes less pronounced as the training sample grows. While not reported here, we also performed the analysis with different fixed ridge penalties throughout the simulations. GCV performed as well as, if not better than, the best fixed ridge penalty selected in hindsight. Monte Carlo standard errors are all below 1 basis point, indicating that the Monte Carlo approximation error in the estimated mean rewards is negligible.

\subsection{Derivation of the Oracle Policy}\label{Section_simulatonsOptimalPolicy}

The oracle policy can be calculated using the operator $\mathbb{K}^{\pi}$ defined in Section S.1. The relevant state consists of $\tilde{S}_{t}\in\left\{ 0,1\right\} ^{4}$ and the previously selected strategy $a\in\mathcal{A}=\{1,2,3,4\}$, i.e. we have four raw binary states and four different strategies. Note that $\tilde{S}_{t}$ is just a binary encoding of $X_{t}$; we use this representation for consistency in the notation used in the simulation. Clearly, $X_{t}=i$ means $\tilde{S}_{t,i}=1$ and $\tilde{S}_{t,j}=0$, $j\neq i$. Hence, the state $S_{t}=\left(\tilde{S}_{t},a\right)$ only takes 16 different values. For any fixed policy $\pi$, the operator $\mathbb{K}^{\pi}$ can therefore be represented by a $16\times16$ transition matrix, with rows and columns indexed by the pairs $(i,a)$ and $(j,a')$ arranged in some fixed order, respectively: $\mathbf{K}^{\pi}=\mathbf{K}_{ij}1_{\left\{ a'=\pi(S_{t}^{a})\right\} }$, where $\mathbf{K}$ is as in (\ref{EQ_transitionMatrixSimulations}).

For any candidate policy $\pi$, the Bellman equation (S.3) is 
\[
V_{\eta}^{\pi}\left(s\right)=\mathbb{K}^{\pi}R_{\eta}\left(s\right)+\gamma\mathbb{K}^{\pi}V_{\eta}^{\pi}\left(s\right),
\]
where $s\in\mathcal{S}$ is the state at which we condition at time $t$, including the previous strategy $a$, and where we make explicit the dependence on the risk aversion parameter $\eta$ from Example 1. Because there are only 16 possible states, let $\mathbf{V}_{\eta}^{\pi}$ denote the vector of the 16 values that $V_{\eta}^{\pi}$ can take and $\mathbf{r}_{\eta}^{\pi}$ the corresponding vector of expected rewards $\mathbb{K}^{\pi}R_{\eta}\left(S_{t}^{a}\right)$. We obtain
\[
\mathbf{V}_{\eta}^{\pi}=\mathbf{r}_{\eta}^{\pi}+\gamma\mathbf{K}^{\pi}\mathbf{V}_{\eta}^{\pi}
\]
using matrix notation. Consequently, $\mathbf{V}_{\eta}^{\pi}=(I-\gamma\mathbf{K}^{\pi})^{-1}\mathbf{r}_{\eta}^{\pi}$ so that if we know $\mathbf{r}_{\eta}^{\pi}$, we can derive $\mathbf{V}_{\eta}^{\pi}$, i.e. the vector that collects the range of values of $V_{\eta}^{\pi}(S_{t}^{a})$. Under Examples (\ref{Example_reward}) and (\ref{Example_costRegularisation}), and the given DGP, the 16 entries in $\mathbf{r}_{\eta}^{\pi}$ can be computed explicitly. By definition, we have that 
\begin{align*}
\mathbb{K}^{a'}R_{\eta}\left(S_{t}^{a}\right)= & \mathbb{E}\left[\left(\beta1_{\left\{ X_{t+1}=a'\right\} }+\varepsilon_{t+1}^{a'}-\mathrm{Cost}_{t}^{\left(a,a'\right)}\right)\right.\\
 & \left.-\frac{\eta}{2}\left(\beta1_{\left\{ X_{t+1}=a'\right\} }+\varepsilon_{t+1}^{a'}-\mathrm{Cost}_{t}^{\left(a,a'\right)}\right)^{2}|S_{t}^{a}\right].
\end{align*}
Let $e_{i}$ be the $4$-dimensional column vector with zero entries except for the $i$th one which is equal to one. Then, $\Pr\left(X_{t+1}=a'|S_{t}^{a}\right)=\tilde{S}_{t}^{\mathrm{T}}\mathbf{K}e_{a'}$. So, using the moment assumptions on the error term, we deduce that
\begin{align*}
\mathbb{K}^{a'}R_{\eta}\left(S_{t}^{a}\right)= & \left[\beta-\frac{\eta}{2}\left(\beta^{2}-2\beta\mathrm{Cost}_{t}^{\left(a,a'\right)}\right)\right]\tilde{S}_{t}^{\mathrm{T}}\mathbf{K}e_{a'}\\
 & -\mathrm{Cost}_{t}^{\left(a,a'\right)}-\frac{\eta}{2}\left(\sigma_{\varepsilon}^{2}+\left|\mathrm{Cost}_{t}^{\left(a,a'\right)}\right|^{2}\right).
\end{align*}
For a given policy $\pi$, to find $\mathbb{K}^{\pi}R_{\eta}\left(S_{t}^{a}\right)$, we set $a'=\pi\left(S_{t}^{a}\right)$ in the above. Then, $\mathbf{r}_{\eta}^{\pi}$ collects the $16$ possible values of $\mathbb{K}^{\pi}R_{\eta}\left(S_{t}^{a}\right)$; $S_{t}^{a}$ only takes 16 different values. 

Consider the candidate policy $\pi(S_{t}^{a})=1$ if $X_{t}\in\{1,3\}$ and $\pi(S_{t}^{a})=2$ if $X_{t}\in\{2,4\}$. To establish optimality, we evaluate this policy using the matrix expression above and verify the Bellman optimality inequalities against all possible actions $a'\in\mathcal{A}$:
\[
V_{\eta}^{\pi}(S_{t}^{a})\geq\mathbb{K}^{a'}R_{\eta}(S_{t}^{a})+\gamma\mathbb{K}^{a'}V_{\eta}^{\pi}(S_{t}^{a})
\]
for the 16 different values of $S_{t}^{a}$. In the calculations, we use the actual numbers $\beta=0.01$, $\sigma_{\varepsilon}=0.04$,  $\mathrm{Cost}_{t}^{(a,a')}=0.012\times1_{\{a\ne a'\}}$ and $\mathbf{K}$ in (\ref{EQ_transitionMatrixSimulations}). By numerical calculation, it can be verified that the policy is optimal for $\eta\geq0$ but smaller than $164$. For a fixed policy, the value function and the Bellman optimality inequalities are affine in $\eta$. Hence, verifying these inequalities at $\eta=0$ and $\eta=164$ establishes optimality throughout the interval.

\pagebreak{}

\end{document}